\documentclass[12pt]{book}

\usepackage[top=1in, bottom=1in, left=1in, right=1in]{geometry}
\usepackage[rflt]{floatflt}
\usepackage{graphicx}
\usepackage{times}

\usepackage[authoryear]{natbib}

\usepackage{enumitem}

\usepackage{url}

\usepackage{graphicx}
\usepackage{subfigure}
\usepackage{algorithm}
\usepackage{algorithmic}
\usepackage{booktabs}

\usepackage{url}
\usepackage{verbatim}
\usepackage{amsmath}
\usepackage{amsfonts}
\usepackage{amssymb}

\DeclareMathOperator*{\argmax}{arg\,\!max}

\newtheorem{definition}{Definition}

\newcommand{\paragraphs}[1]{\paragraph{#1}\hspace{-10pt}}

\newcommand{\fix}[1]{}

\usepackage[T1]{fontenc}

\usepackage{hyperref}

\usepackage{amssymb}
\usepackage{amsfonts}
\usepackage{amsmath}
\providecommand{\joint}[1]{\mathbf{#1}}

\newcommand{\set}[1]{\mathbb{#1}}

\newcommand{\Exp}{\mathbb{E}}

\providecommand{\argsA}[2]{ {#1}_{#2} }             
\providecommand{\argsI}[2]{ {#1}^{#2} }           
\providecommand{\argsT}[2]{ {#1}_{#2} }             

\providecommand{\argsAI}[3]{ {#1}_{#2}^{#3} }
\providecommand{\argsAT}[3]{ {#1}_{#2,#3} }

\providecommand{\argsIT}[3]{ {#1}^{#2}_{#3} }

\providecommand{\agentS}{\set{I}}

\providecommand{\stateS}{\set{S}}
\providecommand{\sI}[1]{\argsI {s}   {#1}}   

\providecommand{\discount}{\gamma}

\providecommand{\hor}{\mathcal{H}}

\providecommand{\nrA}{n} 

\providecommand{\ts}{t}        

\providecommand{\bSymbol}           {{b}}
\providecommand{\bO}               { {\argsT\bSymbol{0}}}               

\providecommand{\pol}{\POL}
\providecommand{\POL}{\pi}

\providecommand{\jpol}      {\joint{\POL}}	 

\providecommand{\polA}[1]   {\argsA{\POL}{#1}}	
\providecommand{\poli}   {\polA i}  

\providecommand{\AOH}{h} 
\providecommand{\AOHS}{\set{H}} 

\providecommand{\JAOH}{\joint{\AOH}{}} 
\providecommand{\JAOHS}{\joint{\AOHS}{}} 
\providecommand{\jh}        {\JAOH}
\providecommand{\jhT}[1]{\argsT {\jh}   {#1}}   
\providecommand{\jhS}       {\JAOHS}

\providecommand{\aoHistA}[1]    {\argsA   {\AOH}{#1}} 
\providecommand{\hi    {\aoHistA i}}  
\providecommand{\hiT}[1]{\argsAT {\AOH} {i} {#1}} 
\providecommand{\hAOHS}[1]{\argsA {\AOHS}  {#1}} 
\providecommand{\aoHistAT}[2]   {\argsAT  {\AOH}{#1}{\,#2}}

\providecommand{\AC}{a}                 
\providecommand{\ACS}{\set{A}}                     

\providecommand{\ja}    {\joint {\AC}}          
\providecommand{\jaS}   {\joint {\ACS}}         
\providecommand{\jaI}[1]{\argsI {\ja}   {#1}}   
\providecommand{\jaT}[1]{\argsT {\ja}   {#1}}   
\providecommand{\jaIT}[2]{\argsIT {\ja}   {#1} {#2}}   

\providecommand{\aA}[1] {\argsA {\AC}   {#1}} 
\providecommand{\ai    {\aA i}} 
\providecommand{\aiT}[1]{\argsAT {\AC} {i} {#1}} 
\providecommand{\aAS}[1] {\argsA {\ACS}  {#1}} 
\providecommand{\aAT}[2]{\argsAT{\AC}   {#1}{#2}}
\providecommand{\aAI}[2]{\argsAI{\AC}   {#1}{#2}} 

\providecommand{\OB}{o}                
\providecommand{\OBS}{\set{O}}

\providecommand{\jo}    {\joint {\OB}}      
\providecommand{\joS}   {\joint {\OBS}}    
\providecommand{\joT}[1]{\argsT {\jo}   {#1}} 
\providecommand{\joIT}[2]{\argsIT {\jo}   {#1} {#2}}   

\providecommand{\oA}[1] {\argsA {\OB}   {#1}} 
\providecommand{\oi    {\oA i}} 
\providecommand{\oiT}[1]{\argsAT {\OB} {i} {#1}} 
\providecommand{\oAS}[1]{\argsA {\OBS}  {#1}}   
\providecommand{\oASi    {\oA i}} 

\providecommand{\jQ}    {\joint {Q}}
\providecommand{\jQpol}    {\joint {Q^{\jpol}}}
\providecommand{\QA}[1] {\argsA {Q}   {#1}} 
\providecommand{\Qi}    {\QA i} 
\providecommand{\jQ}    {\joint {Q}}
\providecommand{\jQmodel}    {\hat{\joint {Q}}}

\providecommand{\jV}    {\joint {V}}  

\providecommand{\VA}[1] {\argsA {V}   {#1}} 
\providecommand{\Vi}    {\VA i} 
\providecommand{\jVmodel}    {\hat{\joint {V}}}
\providecommand{\Vimodel}    {\hat \Vi} 

\providecommand{\jA}    {\joint {A}}  
\providecommand{\Ai}    {\argsA A i} 
\providecommand{\jAmodel}    {\hat{\joint {A}}}

\providecommand{\rT}[1]{\argsT {r}   {#1}} 
\providecommand{\rIT}[2]{\argsIT {r}   {#1} {#2}}   

\providecommand{\nablasub}[1]{\nabla_{\! {#1}}}
\newcommand\nablai{\nabla_{\! \ppi}}
\providecommand{\thetai} {\argsA {\theta}  {i}} 

\providecommand{\valparam}{\theta}
\providecommand{\polparam}{\psi}
\providecommand{\vpi}{\argsA {\valparam} {i}}
\providecommand{\ppi}{\argsA {\polparam} {i}}

\usepackage{xcolor}

\title{An Initial Introduction to \\Cooperative Multi-Agent Reinforcement Learning}
\author{Christopher Amato, Northeastern University}

\begin{document}

\maketitle
\vspace{-20pt}
\tableofcontents

\chapter{Introduction}
\label{sec:intro}

\section{Overview}
\label{sec:overview}
Multi-agent reinforcement learning (MARL) has exploded in popularity in recent years. While numerous approaches have been developed, they can be broadly categorized into three main types: centralized training and execution (CTE), centralized training for decentralized execution (CTDE), and decentralized training and execution (DTE).

CTE methods assume centralization during training and execution (e.g., with fast, free, and perfect communication) and have the most information during execution. That is, the actions of each agent can depend on the information from all agents. As a result, a simple form of CTE can be achieved by using a single-agent RL method with centralized action and observation spaces (maintaining a centralized action-observation history for the partially observable case). CTE methods can potentially outperform the decentralized execution methods (since they allow centralized control) but are less scalable as the (centralized) action and observation spaces scale exponentially with the number of agents. CTE is typically only used in the cooperative MARL case since centralized control implies coordination on what actions will be selected by each agent. 

CTDE methods are the most common, as they leverage centralized information during training while enabling decentralized execution---using only information available to that agent during execution. CTDE is the only paradigm that requires a separate training phase where any available information (e.g., other agent policies, underlying states) can be used. 
As a result, they can be more scalable than CTE methods, and do not require communication during execution.  CTDE fits most naturally with the cooperative case, but may also apply in competitive or mixed settings depending on what information is assumed to be observed. 

Decentralized training and execution methods make the fewest assumptions and are often simple to implement. In fact, any single-agent RL method can be used for DTE by just letting each agent learn separately. Naturally, these approaches have their respective advantages and disadvantages, which will be discussed. It is worth noting that DTE is required if no centralized training phase is available (e.g., through a centralized simulator), requiring all agents to learn during online interactions without prior coordination.   
DTE methods can be applied in cooperative, competitive, or mixed cases. 

MARL methods can be further categorized into value-based and policy gradient methods. Value-based methods (e.g., Q-learning) learn a value function and then choose actions based on those values. Policy gradient methods learn an explicit policy representation and attempt to improve the policy in the direction of the gradient. Both classes of methods are widely used in MARL. 

This text is an introduction to cooperative MARL---MARL in which all agents share a single, joint reward. It is meant to explain the setting, basic concepts, and common methods for the CTE, CTDE, and DTE settings. It does not cover all work in cooperative MARL as the area is quite extensive. I have included work that I believe is important for understanding the main concepts in the area and apologize to those that I have omitted.

I will first give a brief description of the cooperative MARL problem in the form of the Dec-POMDP as well as a short background on the (single-agent) reinforcement learning methods that are relevant to this text. Then, I will discuss the centralized, decentralized, and CTDE settings and algorithms. Note that all the presented methods are model-free (i.e., they do not learn a model of the dynamics or observations) since the vast majority of cooperative MARL methods are model-free. 

For the centralized case, I will discuss the setting as well as ways of modeling the CTE problem, simple applications of single-agent methods, and some more scalable methods that exploit the multi-agent structure. 

For the decentralized case, I will give an overview of the DTE setting and then I will discuss the relevant value-based and policy gradient methods. I'll start with value-based DTE methods such as independent Q-learning and its extensions. Then, I will  discuss the extension to the deep case with DQN, the additional complications this causes, and methods that have been developed to (attempt to) address these issues.  Next, I will discuss policy gradient DTE methods starting with independent REINFORCE \citep{Peshkin00UAI} (i.e., vanilla policy gradient), and then extending to the actor-critic case and deep variants (such as independent PPO \citep{de20,MAPPO}). Finally, I will discuss some general topics related to DTE and future directions. 

Finally, I will present an overview of CTDE and the two main classes of CTDE methods: value function factorization methods and centralized critic actor-critic methods. Value function factorization methods include the well-known VDN \citep{VDN}, QMIX \citep{QMIX}, and QPLEX \citep{QPLEX} approaches, while centralized critic methods include MADDPG \citep{MADDPG}, COMA \citep{COMA}, and MAPPO \citep{MAPPO}. Lastly, I discuss other forms of CTDE such as adding centralized information to decentralized (i.e., independent) learners (such as parameter sharing) and decentralizing centralized solutions.

A detailed treatment of reinforcement learning (in the single-agent setting) is not presented in this text. Anyone interested in RL should read the book by \cite{SuttonBarto18}. Similarly, for a broader overview of MARL, the recent book by Albrecht, Christianos and Sch\"afer is recommended \citep{marl-book}.

\section{The cooperative MARL problem: The Dec-POMDP}
\label{sec:decpomdp}

\begin{figure}[t]
\centering
\includegraphics[height=0.2\linewidth]{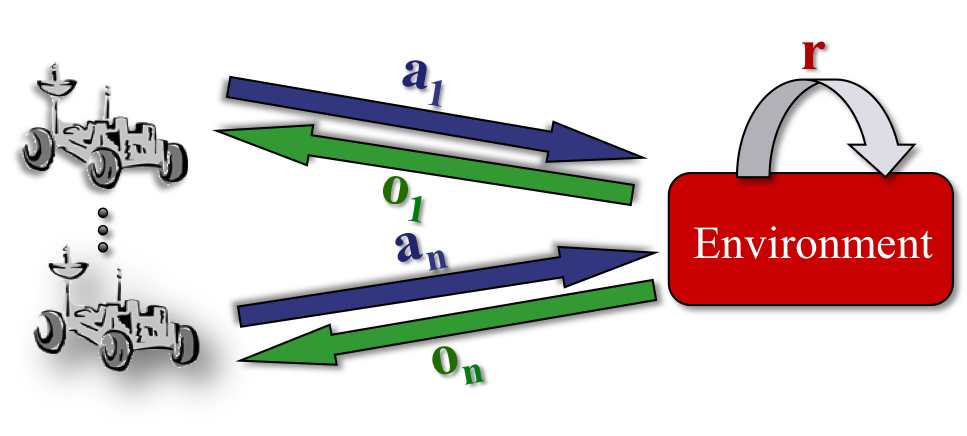}
\caption{A depiction of cooperative MARL---a Dec-POMDP.}
\label{fig:decpomdp}
\end{figure}

The cooperative multi-agent reinforcement learning (MARL) problem can be represented as a decentralized partially observable Markov decision process (Dec-POMDP) \citep{Book16,Bernstein02MOR}. 
Dec-POMDPs generalize POMDPs \citep{Kaelbling98AI} (and MDPs \citep{Puterman94}) to the multi-agent, decentralized setting. They are also a subclass of partially observable stochastic games  (POSGs) where all the agents have the same reward (i.e., a cooperative POSG). 
As depicted in Figure~\ref{fig:decpomdp}, multiple agents  operate under uncertainty based on partial views of the world. Execution unfolds over time. At each step, every agent chooses an action (in parallel) based purely on locally observable information, resulting in each agent obtaining  an observation and the team obtaining a joint reward. The shared reward
function makes the problem cooperative, but their local views mean that execution is decentralized. 

Formally, a  Dec-POMDP is defined by tuple $ \langle  \agentS, \stateS, \{\aAS i\}, T, R, \{\oAS i\}, O, \hor, \discount \rangle $. For simplicity, I define the finite version but it easily extends to the continuous case: 
\begin{itemize}
\item 
$\agentS$ is a finite set of agents of size $|\agentS|=\nrA$;
\item 
$\stateS$ is a finite set of states with designated initial state
distribution $\bO$;\footnote{Some papers refer to other information such as joint observations or histories as `state.' For clarity, I will only use this term to refer to the true underlying state.}
\item  
$\aAS i$ is a finite set of actions for each agent $i$ with $\jaS=\times_i \aAS i$ the set of joint actions; 
\item  
$T$ is a state transition probability function, $T: \stateS \times \jaS \times \stateS \to [0,1]$, that specifies the
probability of transitioning from state $s \in \stateS$ to $s' \in \stateS$ when the actions $\ja \in \jaS$ are taken by the agents (i.e., $T(s,\ja,s') = \Pr(s' | \ja,s)$);
\item 
 $R$ is a reward function: $R: \stateS \times \jaS \to \mathbb{R}$, the immediate reward for being in state $s \in \stateS$ and taking the
actions $\ja \in \jaS$;
\item
 $\oAS i$ is a finite set of observations for each agent, $i$, with $\joS=\times_i \oAS i$ the set of joint observations;
\item
  $O$ is an observation probability function: $O: \joS \times \jaS \times \stateS \to [0,1]$, the
probability of  seeing  observations $\jo \in \joS$  given 
actions $\ja \in \jaS$  were taken and resulting in state $s' \in \stateS$ (i.e.,  $O(\ja,s',\jo) = \Pr(\jo | \ja,s')$);
\item 
$\hor$ is the number of steps until termination, called the horizon;
\item and $\discount \in [0,1]$ is the discount factor. 
\end{itemize}
\vspace{-5pt}
I also include both the horizon and discount in the definition but, typically, only one is used. When $\hor$ is finite, $\gamma$
 can be set to 1 and when $\hor=\infty$,  $\discount \in [0,1)$.\footnote{The episodic case \citep{SuttonBarto18} is typically indefinite-horizon with termination to a set of terminal states with probability 1 and the infinite-horizon case is sometimes called `continuing.'}

A solution to a Dec-POMDP is a \emph{joint policy}, denoted $\jpol$---a set of policies, one for each agent, each of which is denoted $\poli$ and called a \emph{local policy} for agent $i$. Because the state is not directly observed, it is typically beneficial for each agent to remember a history of its observations (and potentially actions). A local policy, $\poli$, for an agent is a mapping from local action-observation histories to actions or probability distributions over actions. 
A \emph{local deterministic policy} for agent $i$, $\poli(\hi)=\ai$, maps $\hAOHS i \to \aAS i$, where $\hAOHS i$ is the set of \emph{local action-observation histories}, $\hi = \{\aiT 0, \oiT 0, \ldots, \aiT {\ts-1}, \oiT {\ts-1} \}$\footnote{Sometimes, $\oiT 0$ is also included as an observation generated by the initial state distribution}, by agent $i$ up to the current time step, $t$. 
Note that histories have implicit time steps due to their length which I do not include in the notation (i.e., I always assume a history starts on the first time step and the last time step is defined by the number of action-observation pairs).  
We can denote the \emph{joint histories} for all agents at a given time step as $\jh=\langle \aoHistA 1, \ldots, \aoHistA \nrA \rangle$. 
A \emph{joint deterministic policy} is denoted $\pi(\jh)=\ja=\langle\polA 1(\aoHistA 1), \dots, \polA \nrA(\aoHistA \nrA)  \rangle=\langle\aA 1, \ldots,  \aA \nrA \rangle$. 
A \emph{stochastic local policy} for agent $i$ is $\poli(\ai| \hi)$, representing the probability of choosing action $\ai$ in history $\hi$.  
A \emph{joint stochastic policy} is denoted  $\jpol(\ja| \jh)=\prod_{i \in \agentS} \poli(\ai| \hi)$. 
Because one policy is generated for each agent and these policies depend only on local observations, they operate in a decentralized manner. 

Many researchers just use observation histories (without including actions), which is sufficient for deterministic policies but may not be for stochastic policies \citep{Book16}. 
Deterministic policies will be used in the value-based methods while stochastic policies will be used in the policy gradient methods.
There always exists an optimal deterministic joint policy in Dec-POMDPs \citep{Book16}, but stochastic (or continuous) policies are needed for policy gradient methods.

The value of a joint policy, $\pi$, at joint history $\jh$ can be defined for the case of discrete states and observations as
\begin{equation}
 \jV^\jpol(\jh)= \sum_{s} P(s|\jh,\bO) \Big[R(s,\ja)+\discount \sum_{s'}P(s'|\ja,s)\sum_{\jo}P(\jo|\ja,s') \jV^\jpol(\jh\ja\jo)\Big]\Big|_{\ja=\jpol(\jh)}
\label{eq:BellmanV}
\end{equation}
where $P(s|\jh,\bO)$ is the probability of state $s$ after observing joint history $\jh$ starting from state distribution $\bO$ and $\ja=\jpol(\jh)$ is the joint action taken at the joint history. Also, $\jh\ja\jo$ represents $\jh'$, the joint history after taking joint action $\ja$ in joint history $\jh$ and observing joint observation $\jo$. 
In the RL context, algorithms do not iterate over states and observations to explicitly calculate this expectation but approximate it through sampling. 
For the finite-horizon case, $\jV^\jpol(\jh)=0$ when the length of $\jh$ equals the horizon $\hor$, showing the value function includes the time step from the history.

We will often want to evaluate policies starting from the beginning---starting at the initial state distribution before any action or observation. Starting from this initial, null history I denote the value of a policy as $\jV^\jpol(\jh_0)$. 
An  \emph{optimal joint policy} beginning at  $\jh_0$ is 
$\argmax_{\jpol} \jV^{\jpol}(\jh_0)$, where the $\argmax$ here denotes enumeration over decentralized policies. 
The optimal joint policy is then the set of local policies for each agent that provides the highest value, which is denoted $\jpol^*$ with value $\jV^*$.

Reinforcement learning methods often use history-action values, $\jQ(\jh,\ja)$, rather than just history values $\jV(\jh)$. $\jQ^{\jpol}(\jh,\ja)$ is the value of choosing joint action $\ja$ at joint history $\jh$ and then continuing with policy $\jpol$, 
\begin{equation}
 \jQ^\jpol(\jh, \ja)= \sum_{s} P(s|\jh,\bO) \Big[R(s,\ja)+\discount \sum_{s'}P(s'|\ja,s)\sum_{\jo}P(\jo|\ja,s') \jQ^\jpol\big(\jh\ja\jo, \ja'\big)|_{\ja'=\jpol(\jh\ja\jo)}\Big],
 \label{eq:BellmanQ}
\end{equation}
while $\jQ^*(\jh,\ja)$ is the value of choosing action $\ja$ at history $\jh$ and then continuing with the optimal policy, $\jpol^*$.

\paragraph{Policies that depend only on a single observation}
\label{sec:singleobs}
It is somewhat popular to define policies that depend only on a single observation rather than the whole observation history. 
That is, $\poli: \oAS i \to \aAS i$, rather than $\hAOHS i \to \aAS i$. This type of policy is often referred to as \emph{reactive} or \emph{Markov} since it just depends on (or reacts from) the past observation. These reactive policies are typically not desirable since they can be arbitrarily worse than policies that consider history information \citep{Murphy00} but can perform well or even be optimal in simpler subclasses of Dec-POMDPs \citep{Goldman04JAIR}. In general, reactive policies reduce the complexity of finding a solution (since many fewer policies need to be considered) but only perform well in limited  cases such as when the problem is not really partially observable or when the observation history isn't helpful for decision-making.

\paragraph{The fully observable case}
\label{sec:fullobs}
It is worth noting that if the cooperative problem is fully observable, it becomes much simpler. 
Specifically, a (decentralized) multi-agent MDP (MMDP) can be defined by the tuple $ \langle  \agentS, \stateS, \{\aAS i\}, T, R, \hor, \discount \rangle $\citep{Boutilier96TARK}. Each agent can observe the full state in this case. In the CTDE context, solving an MMDP can be done by standard MDP methods where the action space is the joint action space, $\jaS=\times_i \aAS i$. The resulting policy, $\stateS \to \jaS$, can be trivially decentralized as $\stateS \to \aAS i$ for each agent $i$. A number of methods have been developed for learning in MMDPs (and other multi-agent models) \citep{Busoniu08IEEE_SMC}. More details about solving MMDPs (for decentralized and centralized control) are given in Section \ref{sec:cte}. 

Finding solutions in Dec-POMDPs is much more challenging. 
This can even be seen by the (finite-horizon) complexity for finite problems---optimally solving an MDP is P-complete (polynomial in the size of the system) \citep{papadimitriou1987complexity,Littman95} and 
optimally solving a POMDP is PSPACE (essentially exponential in the actions and observations) 
\citep{papadimitriou1987complexity}, while optimally solving a Dec-POMDP is NEXP-complete (essentially doubly exponential) \citep{Bernstein02MOR}. 
This can be intuitively understood by considering the simplest possible solution method for each class of problems. 
For an MDP, you could search all policies mapping states to (joint) actions: $|\jaS|^{|\stateS|}$ of them. 
In a POMDP, policies consider histories and for a given horizon $\hor$, there are ${|\joS|^\hor}$ possible observation histories. Up to horizon  $\hor$, there are then  $\sum_{t=0}^{\hor-1} {(|\joS|)^t}=\frac{|\joS|^\hor-1}{|\joS|-1}$ possible observation histories.  Deterministic policies map from these observation histories to actions, giving $|\jaS|^{\frac{|\joS|^\hor-1}{|\joS|-1}}$ possible policies. 
In the Dec-POMDP case, each agent would have $|\aAS i|^{\frac{|\oAS i|^\hor-1}{|\oAS i|-1}}$    
possible policies, so the number of possible joint policies would be $\big(|\aAS i|^{\frac{|\oAS i|^\hor-1}{|\oAS i|-1}}\big)^{|\agentS|}$    
  if all agents have the same number of actions and observations as agent $i$. 
Of course, state-of-the-art methods are more sophisticated than this and the complexity is for the worst case. Most real-world problems are intractable to solve optimally anyway.  
Regardless, these numbers make it clear that searching the joint space of history-based policies is enormous and the resulting partially observable problem is much more challenging than the fully observable case.

\section{Background on (single-agent) reinforcement learning}
\label{sec:back}
While I don't present a full introduction to reinforcement learning (which can be found in other texts \citep{SuttonBarto18}), I will discuss some of the key methods that MARL methods build on. I break the discussion up into value-based methods (e.g., Q-learning) and policy gradient methods, discussing the fully observable case as well as the extension to the partially observable setting. 

\subsection{Value-based methods}
\label{sec:back:value}

Value-based methods learn a value function and then derive the agent's policy from the value function. While Q-learning \citep{Watkins1992} is the most famous such method, the class of value-based methods is much larger. Nevertheless, all the value-based MARL methods that I discuss are based on some form of Q-learning. Therefore, I focus on these Q-learning-based methods below.

\paragraphs{Q-learning} \citep{Watkins1992} is a simple and widely used method for value-based RL. In the single-agent, fully observable case (i.e., an MDP), Q-learning maintains a table of Q-values for each state-action pair. These Q-values are initialized (e.g., randomly) and then the agent takes steps in the environment given the current state as input. That is, given the current state, $s$, the agent chooses an action, $a$, and then observes the next state $s'$ and receives the immediate reward $r$. With this information, the Q-value for $s$ and $a$ is updated with:
 $$Q(s,a) \gets Q(s,a) + \alpha\left[r+ \gamma \max_{a'} Q(s',a')-Q(s,a)\right].$$ The quantity $r+ \gamma \max_{a'} Q(s',a')-Q(s,a)$ is called the TD-error and will be denoted as $\delta$ (resulting in $Q(s,a) \gets Q(s,a) + \alpha\delta$). 
Q-learning has been shown to converge (in single-agent, fully observable environments) with minor assumptions (such as proper exploration and reduction of the learning rate) \citep{Watkins1992,melo2001convergence}. 

Q-learning can be extended to the partially observable case (i.e., a POMDP \citep{Kaelbling98AI}) by using histories rather than states.\footnote{\emph{Beliefs}, or distributions over states, are often used in the model-based or planning cases. However, they require a transition and observation model to calculate so are not typically used in RL.} As noted above, in partially observable environments, the state is not directly observable so agents may need to remember information from the past. It turns out that a POMDP can be transformed into a history-MDP where the `states' of the history-MDP are the histories of the POMDP \citep{historyMDP}. 
Q-learning (and any fully observable RL method) can then be extended to consider histories instead of states. The agent chooses actions, $a$, based on the current action-observation history, $h$, observes the resulting observation, $o$ and receives an immediate reward, $r$. 
The Q-learning update becomes:
 $$Q(h,a) \gets Q(h,a) + \alpha\left[r+ \gamma \max_{a'} Q(h',a')-Q(h,a)\right] $$ 
  where $h'=hao$.

\begin{figure}
\hfill
\begin{centering}
\subfigure[DQN]{
\includegraphics[scale=0.6]{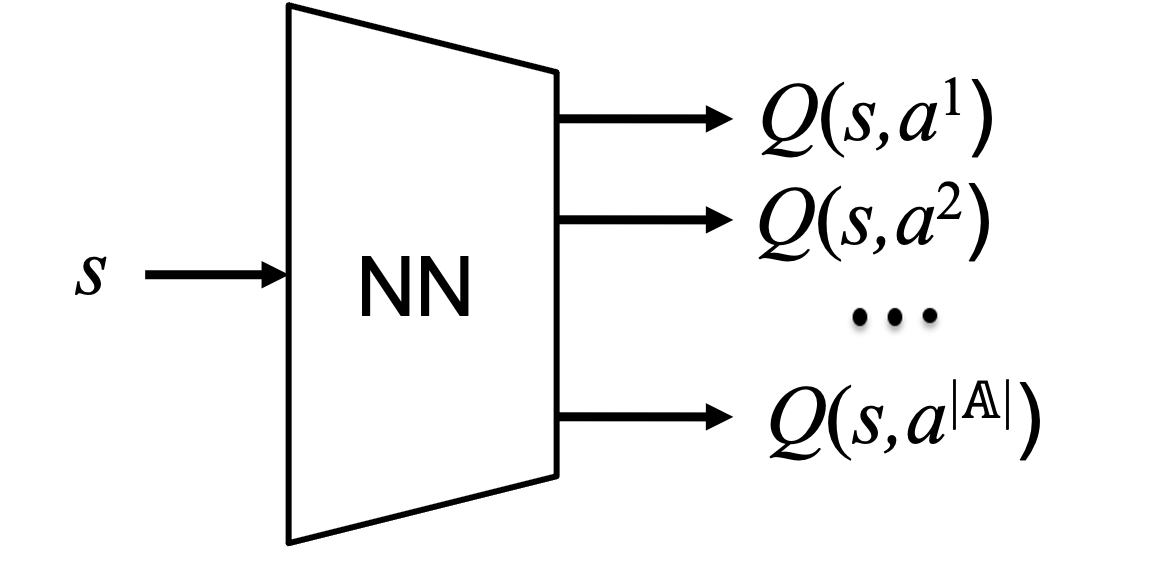} \label{fig:DQN}
}
\subfigure[DRQN]{
\includegraphics[scale=0.55]{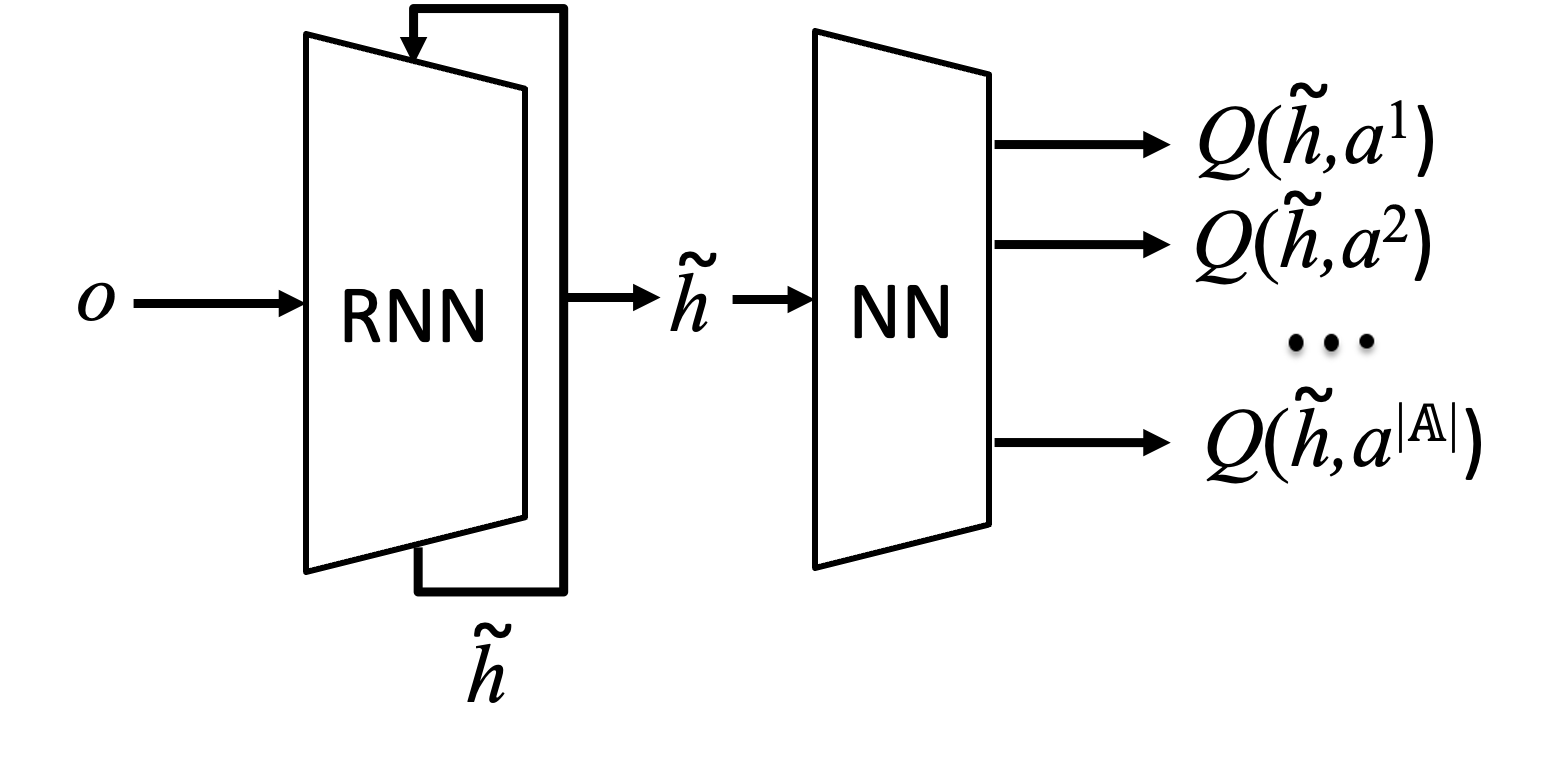}
\label{fig:DRQN}
}
\hfill
\end{centering}
\caption{DQN and DRQN network diagrams.}
\end{figure}

\paragraphs{Deep Q-networks (DQN)}\citep{DQN} is an extension of Q-learning to include a neural net as a function approximator. 
Since DQN was designed for the fully observable (MDP) case, it learns $Q_\theta(s,a)$, parameterized with $\theta$ (i.e., $\theta$ represents the parameters of the neural network), by minimizing the loss:\\[-7pt]
\begin{equation}
    \mathcal{L}(\theta)=\mathbb{E}_{<s, a, r, s'>\sim\mathcal{D}}\Big[\big(y - Q_{\theta}(s, a)\big)^2 \Big] \text{, where\,\, } y=r + \gamma\max_{a'}Q_{\theta^-}(s',a')
    \label{eq:dqn}
\end{equation}
which is just the squared TD error---the difference between the current estimated value, $Q_{\theta}(s, a)$, and the new value gotten from adding the newly seen reward to the previous Q-estimate at the next state, $Q_{\theta^-}(s',a')$. Because learning neural networks can be unstable, a separate target action-value function $Q_{\theta^-}$ and an experience replay buffer $\mathcal{D}$ \citep{Lin92} are implemented to stabilize learning. The target network is an older version of the Q-estimator that is updated periodically with $\langle s,a,r,s'\rangle$ sequences stored in the experience replay buffer and single $\langle s,a,r,s'\rangle$ tuples are i.i.d.~sampled for updates. 
As shown in Figure \ref{fig:DQN}, the neural network (NN) outputs values for all actions ($\AC \in \ACS$) to make maximizing over all actions possible with a single forward pass (rather than iterating through the actions). 

\paragraphs{Deep recurrent Q-networks (DRQN)}\citep{DRQN} extends DQN to handle partial observability, where some number of recurrent layers (e.g., LSTM \citep{LSTM}) are included to maintain an internal hidden state which is an abstraction of the history (as shown as $\tilde h$ in Figure \ref{fig:DRQN}). Because the problem is partially observable, $o,a,r,o'$ sequences are stored in the replay buffer during execution. The update equation is very similar to that of DQN:
\begin{equation}
    \mathcal{L}(\theta)=\mathbb{E}_{<h,a,r,o>\sim\mathcal{D}}\Big[\big(y - Q_{\theta}(h, a)\big)^2 \Big] \text{, where\,\, } y=r + \gamma\max_{a'}Q_{\theta^-}(h',a')
    \label{eq:drqn}
\end{equation}
but since the recurrent neural network (RNN) is used, the internal state of the RNN can be thought of as a history representation. 
RNNs are trained sequentially, updating the internal state at each time step. As a result, the figure shows only $o$ as the input but this assumes the internal state $\tilde h$ has already been updated using the history up to this point $h^{t-1}$. 
Therefore, rather than just sampling single updates ($o,a,r,o'$) i.i.d. like in DQN, histories are sampled from the buffer. The internal state can be updated incrementally and the Q-values learned starting from the first time step and going until the end of the history (i.e., the horizon or episode). 
Technically, a whole history (e.g., episode) should be sampled from the replay buffer to train the recurrent network but it is common to use a fixed history length (e.g., sample $o,a,o',r$ sequences of length 10). 
Also, as mentioned above, many implementations just use observation histories rather than full action-observation histories. 
Note that I will still write $h$ rather than $\tilde h$ in the equations (e.g., $Q_{\theta}(h, a)$ above) to be more general. 
The resulting equations are not restricted to recurrent models (e.g., non-recurrent representation such as the full history or a Transformer-based representation \citep{DTQN,Ni23}). 
With the popularity of DRQN, it has become common to add recurrent layers to standard (fully observable) deep reinforcement learning methods when solving partially observable problems.

Pseudocode for a version of DRQN is shown in Algorithm \ref{alg:DRQN}. The algorithm uses a learning rate, $\alpha$, some form of exploration, which in this case is $\epsilon$-greedy, and $Q$ is estimated using an RNN parameterized by $\theta$ along with using a target network, $\theta^-$, and a replay buffer, $\mathcal{D}$.  Episodes (either a fixed set or until convergence) are generated by interacting with the environment and these are added to the replay buffer. This buffer typically has a fixed size and new data replaces old data when the buffer is full. 
Episodes are then sampled from the replay buffer (a single episode is sampled here but it can also be a minibatch of episodes).
 In order to have the correct internal state, RNNs are trained sequentially.  Training is shown here from the beginning of the episode until the end, calculating the internal state and the loss at each step  and updating $\theta$ using gradient descent. The target network $\theta^-$ is updated periodically (every $C$ episodes in the code) by copying the parameters from $\theta$.

The code is written from the finite-horizon perspective but it can be adjusted to the episodic or indefinite-horizon case by replacing the horizon with terminal states or the infinite-horizon case by removing the loop over episodes. 
The pseudocode, for this approach and throughout the text, is written to be simple and highlight the main algorithmic components. There are many implementation tricks that are not included.

\begin{algorithm}
\begin{algorithmic}[1]
\STATE set $\alpha$, $\epsilon$, and $C$ (learning rate, exploration, and target update frequency)
\STATE Initialize network parameters $\theta$ and $\theta^-$  for $Q_\valparam$ and $Q_\valparam^-$
\STATE $\mathcal{D} \gets \emptyset$
\STATE $e \gets 1$ \hfill \COMMENT{episode index}
\FORALL{episodes}
\STATE $h \gets \emptyset $ \hfill \COMMENT{initial history is empty}
\FOR{$\ts=0$ to $\hor-1$}
\STATE Choose $a$ at $h$ from $Q^\theta(h,\cdot)$ with exploration (e.g., $\epsilon$-greedy)
\STATE See reward $r$,  observation $o$
\STATE append $a,o,r$ to $\mathcal{D}^e $
\STATE $h \gets hao$ \hfill \COMMENT{update RNN state of the network}
\ENDFOR
\STATE sample an episode from $\mathcal{D}$
\STATE $h \gets \emptyset $ 
\FOR{$\ts=0$ to $\hor-1$}
\STATE $a, o, r \gets \mathcal{D}^e(\ts)$
\STATE $h' \gets hao$ 
\STATE $y=r + \gamma\max_{a'}Q^{\theta^-}(h',a')$
\STATE  Perform gradient descent on parameters $\theta$ with learning rate $\alpha$ and loss:  $ \big(y - Q^{\theta}(h, a)\big)^2  $
\STATE $h \gets h'$
\ENDFOR
\IF{$e\mod C = 0$} 
    \STATE $\theta^- \gets \theta$ 
 \ENDIF
 \STATE $e \gets e+1$
\ENDFOR
\RETURN $Q$ 
\end{algorithmic} 
\caption{DRQN (finite-horizon*)}
\label{alg:DRQN}
\end{algorithm}

\subsection{Policy gradient methods}
\label{sec:back:pg}

\begin{figure}
\hfill
\begin{centering}
\subfigure[Network-based policy for an MDP]{
\includegraphics[scale=0.6]{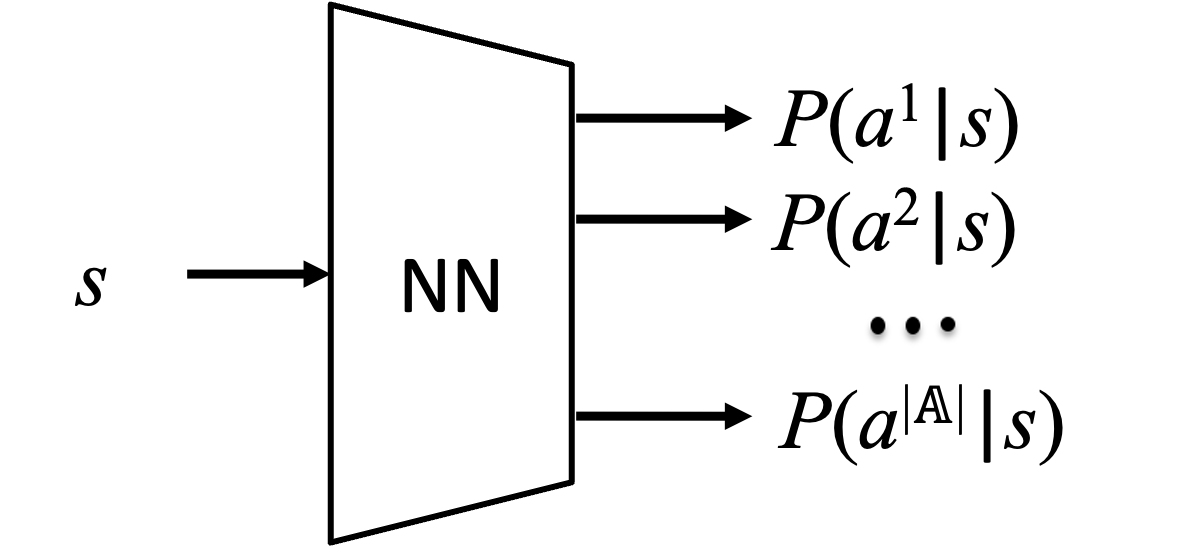} \label{fig:PolNet}
}
\subfigure[RNN-based policy for a POMDP]{
\includegraphics[scale=0.55]{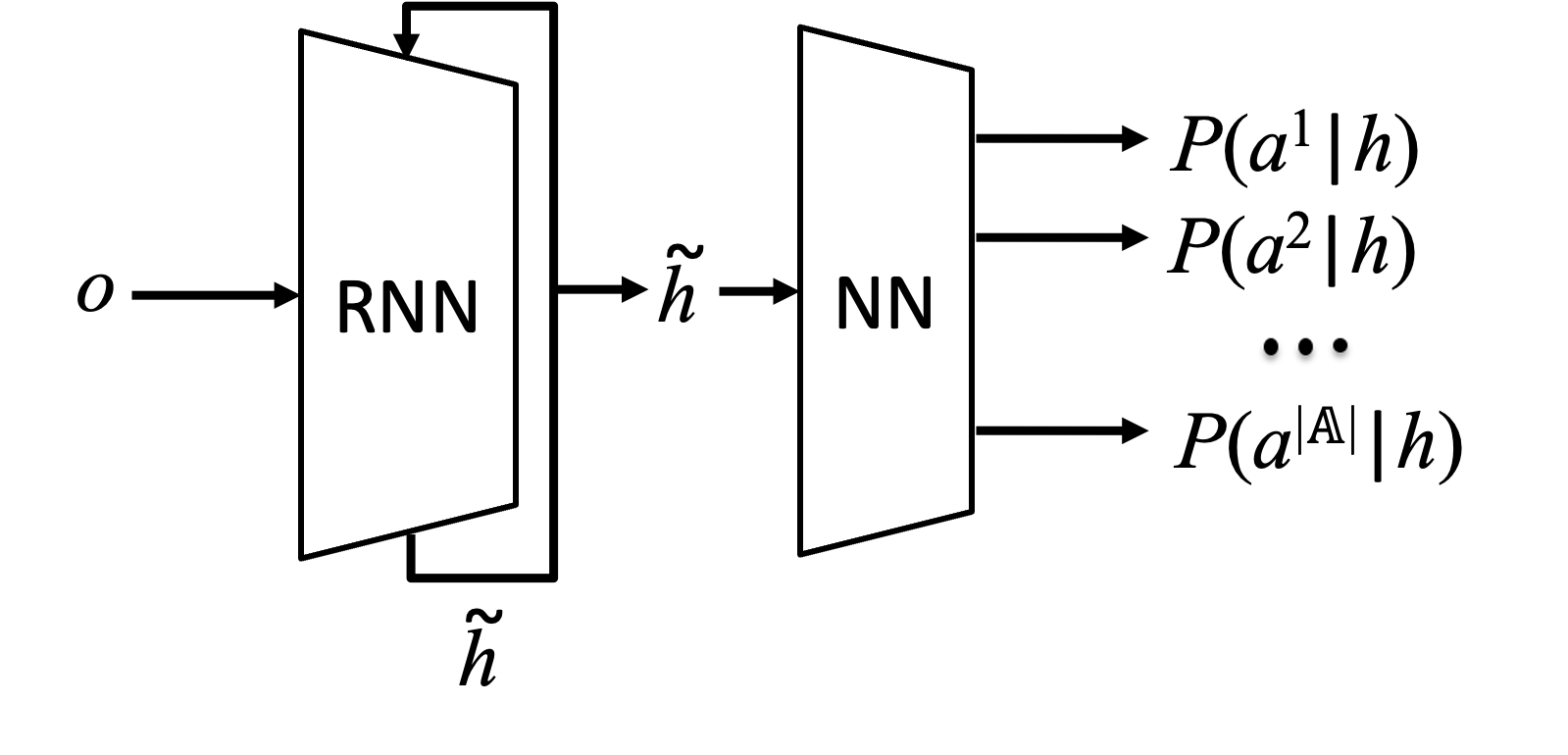}
\label{fig:PolRNN}
}
\hfill
\end{centering}
\caption{State and history-based policies using neural networks.}
\end{figure}

Policy gradient methods explicitly represent the agent's policy and seek to improve it by using a gradient. The policy is parameterized (e.g., using a neural network) and the parameters can be updated using gradient ascent to maximize performance (or gradient descent to minimize loss). For example, a network-based policy for a stochastic policy is given in Figure \ref{fig:PolNet}. Performance is typically denoted as $J$ and I will denote the policy parameters (e.g., weights in the neural network) with  $\polparam$. For instance, performance is often measured for the policy (which I denote as $\pi^\polparam$) from the initial state, so $J(\polparam) = V^{\pi^\polparam} (s_0)$. 

The \emph{policy gradient theorem} provides a way to more easily calculate the gradient of the performance metric with respect to the policy parameters by breaking it up into the discounted state-occupancy using $\pi^\polparam$, $\mu(s)$, the Q-value using $\pi^\polparam$, and the gradient of the policy \citep{SuttonBarto18}:
$$\nabla J(\polparam) = \sum_s \mu(s) \sum_a Q^{\pi^\polparam}(s,a) \nabla \pi^\polparam(a|s)$$.
  Reinforcement learning algorithms approximate this gradient through sampling. 

\paragraphs{REINFORCE} \citep{Williams92ML} is one of the simplest  policy gradient methods. REINFORCE, like many other policy gradient methods, is an on-policy method so it samples states by following the current policy $\pi^\polparam$. As a result, $\mu(s)$ is approximated. REINFORCE uses Monte Carlo estimates for the policy value but these will also equal  $Q^{\pi^\polparam}(s,a) $ in expectation. Assuming a differentiable policy parameterization and sampling an action gives the following update (removing the $\polparam$ superscript from $\pol$ for simplicity of notation): 
$$\polparam \gets \polparam + \alpha \gamma^t G \nabla \log \pol(a| s)$$ 
where $G$ is the Monte Carlo return starting at the current time step (e.g., for the finite-horizon case:  $G=\sum_{k=\ts}^{\hor-1} \gamma^{k-\ts} \rT k $). 

\paragraph{Adding a baseline} 
Because Monte Carlo estimates are often high variance, a baseline is often used to reduce the variance. As long as the baseline doesn't depend on the action, it won't change the gradient. Since the baseline can depend on the state, I will denote it as $b(s)$ but a common baseline is a learned value estimate $\hat V(s)$. REINFORCE with a baseline then uses the following update:
$$\polparam \gets \polparam + \alpha \gamma^t (G-b(s)) \nabla \log \pol(a| s).$$

\paragraphs{Actor-critic methods} combine the TD-learning of value-based methods with policy gradient methods. That is, instead of using Monte Carlo estimate, a TD-based method (e.g., Sarsa or Q-learning) is used to learn a critic, which evaluates the policy.  The policy is referred to as the actor, resulting in the actor-critic name. There are many different actor-critic methods (e.g., PPO \citep{PPO}, soft actor critic \citep{SAC}) but a simple actor-critic update using a value estimate, $\hat V$, is:
$$\polparam \gets \polparam + \alpha \gamma^t (r+\gamma \hat V(s')-\hat V(s)) \nabla \log \pol(a| s).$$ Using the $\delta$ notation from above, this update can also be written as $\polparam \gets \polparam + \alpha \gamma^t \delta \nabla \log \pol(a| s)$.
Actor-critic methods (and policy gradient methods more generally) are widely used and are variants of the ideas above.

\paragraph{Partial observability}
Like the value-based case, we can replace the states of the MDP with histories from the POMDP to extend policy gradient methods to the partially observable case. The policy (and the value function in the actor-critic case) can be represented using neural nets such as in the example of a stochastic RNN-based policy as given in Figure \ref{fig:PolRNN}. Like DRQN \citep{DRQN}, it is common to use recurrent layers (or other methods) to learn history representations. Any of the fully observable methods can be extended but a simple extension of the actor-critic case can be:
$$\polparam \gets \polparam + \alpha \gamma^t (r+\gamma \hat V(h')-\hat V(h)) \nabla \log \pol(a| h)$$
 where $h'=hao$. 
 
 Pseudocode for a simple version of partially observable (advantage) actor-critic is shown in Algorithm \ref{alg:A2C}. This approach is often referred to as advantage actor-critic since it uses the advantage $A(h,a)=Q(h,a)-V(h)$, representing the difference between the Q-value for a given action and the V-value at that history.  In the algorithm, $A(h,a)$ isn't explicitly stored since the TD error is used as an approximation for the advantage---$\hat Q(h,a)$ is approximated with $r_t + \gamma \hat V(h_{t+1})$, which is true in expectation.  $\delta_t$ can be thought of as a sample of  $\hat Q (h_t,a_t)-\hat V(h_t) = \hat A (h_t,a_t)$.  Otherwise, actor and critic models are initialized and then data is generated from episodes. On-policy updates are used for the value function and the actor and critic are updated as discussed above. 
 
 \begin{algorithm}
\caption{Advantage Actor-Critic (A2C) (finite-horizon)}
\begin{algorithmic}[1]
\label{alg:A2C}
\REQUIRE Actor model $\pol(a| h)$, parameterized by $\polparam$
\REQUIRE Critic model $\hat V(h)$, parameterized by $\valparam$
\STATE Initialized $\alpha$, and $\beta$ learning rates for actor and critic)
\FORALL{episodes}
	\STATE  $h_0 \gets \emptyset$ \hfill  \COMMENT{Empty initial history}
	\FOR{$\ts=0$ to $\hor-1$}
		\STATE Choose $a_\ts$ at $h_\ts$ from $\pol(a| h_\ts)$
		\STATE See reward $\rT \ts$ and observation $o_\ts$
		\STATE $h_{\ts+1} \gets h_\ts a_\ts o_\ts$\hfill \COMMENT{Append new action and obs to previous history}
	 \STATE Compute value TD error: $\delta_{t} \gets \rT  \ts + \gamma \hat V(h_{t+1}) - \hat V(h_{t}) $
        \STATE Compute actor gradient estimate: $ \gamma^t \delta_{t} \nabla\log\pol(a_{t}| h_{t})$
        \STATE Update actor parameters $\polparam$ using gradient estimate (e.g., $\polparam \gets \polparam + \alpha \gamma^t \delta_{t} \nabla\log\pol(a_{t}| h_{t})$)
        \STATE Compute critic gradient estimate: $ \delta_{t} \nabla \hat \Vi(h_{t})$
        \STATE Update critic parameters $\valparam$ using gradient estimate (e.g., $\valparam \gets \valparam + \beta \gamma \delta_{t} \nabla \hat V(h_{t}))$
	\ENDFOR
\ENDFOR
\end{algorithmic}
\end{algorithm}

\chapter{Centralized training and execution (CTE)}
\label{sec:cte}

\section{CTE overview}

Centralized execution doesn't fit with the Dec-POMDP definition (since policies need to be decentralized!) but discussing the centralized models is useful and there is some notable work in this area. 
This section begins with assumptions and then discuss the fully observable and the partially observable centralized models, simple centralized solutions, followed by complex approaches leveraging the multi-agent problem structure.

In the CTE case, specifically for the multi-agent POMDP (which is given in detail below), at each step of the problem we assume that:
\begin{itemize}
	\item a centralized controller chooses actions for each agent, $\ja$, based on the current histories for all agents, $\jh$ (which are maintained and updated),
	\item each agent takes the chosen actions $\ja=\langle \aA 1, \ldots, \aA n \rangle$, 
	\item the centralized controller observes the resulting observations $\jo=\langle \oA 1, \ldots, \oA n \rangle$
	\item the (centralized) algorithm observes the current agent histories, $\jh$, (or this could be stored from the previous steps) the actions taken by the agents $\ja$, the resulting observations, $\ja$, and the joint reward $r$. 
\end{itemize}
I have broken the CTE case into the agents, a centralized controller, and the solution algorithm. In this case, the agents and the controller can be the same since the problem can be considered a single-agent problem using the joint actions, observations, and histories (as discussed below).  Nevertheless, I separated them to keep multi-agent execution while emphasizing centralized control. The algorithm is useful to separate from the agent since it has different information (such as rewards) and seeks to optimize the behavior. Agents in any MDP-based model do not condition their policies on rewards so the agents can always operate in environments even without rewards. The rewards are used by the solution algorithm (RL or planning) to evaluate different policies. This separation of algorithm from agent is particularly relevant in the  CTDE case where the algorithm has access to centralized information while the agents do not.

\section{Centralized models}
\label{sec:cte:models}

\paragraphs{The multi-agent MDP (MMDP)} is the fully observable version of the Dec-POMDP, as mentioned in Section \ref{sec:decpomdp}.  Unfortunately, the MMDP model doesn't specify whether the control is done in a centralized or decentralized manner. 
There is a separate model called the Dec-MDP, which assumes the combined observations uniquely determine the state, but agents still only observe their part (e.g., a state feature $s_i$) \citep{Book16}.  Similarly,  the multi-agent POMDP (MPOMDP) model is used for cases when all agents share their observations at each time step (e.g., see the same observations or send their observations to each other) \citep{Book16}. But, again, the MPOMDP model doesn't specify whether control is centralized or decentralized. 
Therefore, I will use the MMDP and MPOMDP names but be explicit about what type of policy is used.\footnote{I propose to call the centralized control versions the Cent-MMDP and Cent-MPOMDP and the decentralized control versions the Dec-MMDP and the Dec-MPOMDP but this is not standard in the literature.}

The MMDP can be defined with the tuple  $ \langle  \agentS, \stateS, \{\aAS i\}, T, R, \hor, \discount \rangle $:
\begin{itemize}
\item 
$\agentS$ is a finite set of agents of size $|\agentS|=\nrA$;
\item 
$\stateS$ is a finite set of states with designated initial state $s_0$ (or state distribution $\bO$);
\item  
$\aAS i$ is a finite set of actions for each agent $i$ with $\jaS=\times_i \aAS i$ the set of joint actions; 
\item  
$T$ is a state transition probability function, $T: \stateS \times \jaS \times \stateS \to [0,1]$, that specifies the
probability of transitioning from state $s \in \stateS$ to $s' \in \stateS$ when the actions $\ja \in \jaS$ are taken by the agents (i.e., $T(s,\ja,s') = \Pr(s' | \ja,s)$);
\item 
 $R$ is a reward function: $R: \stateS \times \jaS \to \mathbb{R}$, the immediate reward for being in state $s \in \stateS$ and taking the
actions $\ja \in \jaS$;
\item 
$\hor$ is the number of steps until termination, called the horizon;
\item and $\discount \in [0,1]$ is the discount factor. 
\end{itemize}
Like in Section \ref{sec:decpomdp}, I am including both the horizon and discount in these definitions but, typically, only one is used. Centralized policies in MMDPs map from states to joint actions, $\jpol_{MMDP}: \stateS \to  \jaS$.

\paragraphs{The multi-agent POMDP (MPOMDP)} can be defined with the tuple $ \langle  \agentS, \stateS, \{\aAS i\}, T, R, \{\oAS i\}, O, \hor, \discount \rangle $:
\begin{itemize}
\item 
$\agentS$ is a finite set of agents of size $|\agentS|=\nrA$;
\item 
$\stateS$ is a finite set of states with designated initial state
distribution $\bO$;
\item  
$\aAS i$ is a finite set of actions for each agent $i$ with $\jaS=\times_i \aAS i$ the set of joint actions; 
\item  
$T$ is a state transition probability function, $T: \stateS \times \jaS \times \stateS \to [0,1]$, that specifies the
probability of transitioning from state $s \in \stateS$ to $s' \in \stateS$ when the actions $\ja \in \jaS$ are taken by the agents (i.e., $T(s,\ja,s') = \Pr(s' | \ja,s)$);
\item 
 $R$ is a reward function: $R: \stateS \times \jaS \to \mathbb{R}$, the immediate reward for being in state $s \in \stateS$ and taking the
actions $\ja \in \jaS$;
\item
 $\oAS i$ is a finite set of observations for each agent, $i$, with $\joS=\times_i \oAS i$ the set of joint observations;
\item
  $O$ is an observation probability function: $O: \joS \times \jaS \times \stateS \to [0,1]$, the
probability of  seeing  observations $\jo \in \joS$  given 
actions $\ja \in \jaS$  were taken and resulting in state $s' \in \stateS$ (i.e.,  $O(\ja,s',\jo) = \Pr(\jo | \ja,s')$);
\item 
$\hor$ is the number of steps until termination, called the horizon;
\item and $\discount \in [0,1]$ is the discount factor. 
\end{itemize}
The MPOMDP definition looks the same as the one for Dec-POMDP but the key difference is the policy used. 
Dec-POMDPs require a joint policy, $\jpol_{Dec-POMDP}$, which is a set of local policies $\pi_{i,Dec-POMDP}$:  
 $\hAOHS i \to \aAS i$, but in a (centralized) MPOMDP the policies map from \emph{joint histories to joint actions}, $\jpol_{MPOMDP}$: $\JAOHS  \to \jaS$.\footnote{Decentralized control in an MPOMDP would result in policies  $\JAOHS \to \aAS i$.} 
  As a result, the (centralized) MPOMDP is really just a factored POMDP with the action sets represented as $\jaS=\times_i \aAS i$ and the observation sets represented as $\joS=\times_i \oAS i$. Similarly, the (centralized) MMDP is really just a factored MDP with action sets $\jaS=\times_i \aAS i$. As a result, standard MDP and POMDP methods can be applied to MMDPs and MPOMDPs\footnote{As mentioned in Section \ref{sec:decpomdp} decentralization is trivial once a centralized solution is found for an MMDP and this is also true for an MPOMDP because in both cases the agents observe the same information.} but such methods may not be scalable due to the potentially large action (and observation) sets. Therefore, scalability has been the focus of CTE MARL methods.

\section{Centralized solutions}

To solve the MMDP, any (fully observable) single-agent RL method can be applied. For instance, the Q-learning update for an MMDP becomes:
 $$\jQ(s,\ja) \gets \jQ(s,\ja) + \alpha\left[r+ \gamma \max_{\ja'} \jQ(s',\ja')-\jQ(s,\ja)\right] $$ 
 where $\ja =\langle \aA 1, \ldots, \aA n \rangle \in \jaS=\times_i \aAS i $. A centralized solution would be $\jpol_{MMDP}: \stateS \to  \jaS$ while a decentralized solution would be $\pi_{i, Decentralized MMDP}: \stateS \to   \aAS i$ for each agent $i$. For the Q-learning case, the solution is determined by $ \jpol_{MMDP}(s)=\argmax_{\ja} \jQ(s,\ja)$. Once the centralized solution is determined, a decentralized solution is obtained using $\pi_{i, Decentralized MMDP}(s)= \aA i$ where $\jpol_{MMDP}(s)=\langle \aA 1, \ldots, \aA n \rangle$.
  For example, one possible  centralized policy for two agents at state $\sI {42}$ could be $\pi_{MMDP}(\sI {42})= \jaI 7=\langle \aAI 1 1,  \aAI 2 7\rangle$ while the corresponding decentralized policy would be $\jpol_{Decentralized MMDP}=\langle \polA 1, \polA 2 \rangle$ where  $\pi_{1, Decentralized MMDP}(\sI {42})=\aAI 1 1$ and $\pi_{2, Decentralized MMDP}(\sI {42})=\aAI 2 7$ (where I would typically drop the agent subscript in the action since it is already in the policy). Because the policy is factored in terms of the agents, there is an action associated with each agent. 
 Other algorithms can be applied by similarly replacing the actions of the MDP with the joint actions of the MMDP. 
 Note that we are currently assuming that we can first solve the MMDP in a centralized manner. If it must be solved in a decentralized manner, additional coordination is needed to ensure the same policy is learned as mentioned in Section \ref{sec:dte} and in other literature \citep{Busoniu08IEEE_SMC,littman2001value}. 

Solving an MPOMDP can be done in a similar manner. We can transform the MPOMDP into a POMDP where the history, $h$, becomes the joint history, $\jh$ and the actions, $a$ become the joint actions, $\ja$. 
For instance, the Q-learning update for an MPOMDP becomes:
 $$\jQ(\jh,\ja) \gets \jQ(\jh,\ja) + \alpha\left[r+ \gamma \max_{\ja'} \jQ(\jh',\ja')-\jQ(\jh,\ja)\right] $$ 
where $\jh=\langle \aoHistA 1, \ldots, \aoHistA n \rangle$, $\ja =\langle \aA 1, \ldots, \aA n \rangle\in \jaS=\times_i \aAS i$, and $\jh'=\jh\ja\jo=\langle \aoHistA 1', \ldots, \aoHistA n' \rangle$ with $\jo=\langle \oA 1, \ldots, \oA n \rangle \in \joS=\times_i \oAS i$. A centralized solution to an MPOMDP would be $\jpol_{MPOMDP}: \JAOHS \to  \jaS$ while a decentralized solution would be $\pi_{i, Decentralized MPOMDP}:  \JAOHS \to   \aAS i$ for each agent $i$. Like the MMDP case, the difference is the output (i.e., the joint or individual agent actions) while the input (i.e., the joint histories) remains the same. 
Other POMDP algorithms can be applied to the MPOMDP case by replacing the histories with joint histories, using the joint observations to update the joint histories (i.e., all agents share their information), and replacing the actions with joint actions. 
Again, additional coordination is needed when solving MPOMDPs in a decentralized manner. 

In the other chapters of the book, I will assume the policies are all for Dec-POMDPs so I will drop the subscript indicating what kind of policy it is. 

\section{Improving scalability}
\label{sec:cte:scale}

While the approaches above appear to be simple, they have one major issue---scalability. As the state and action spaces in the MMDP or the observation and action spaces in the MPOMDP grow, solving these problems becomes computationally intractable. Numerous methods leverage the multi-agent structure to address scalability challenges. I will discuss only a few methods. 

One common solution is to assume additional structure in terms of how the agents can coordinate such as through \emph{coordination graphs} \citep{Guestrin01}. Coordination graphs decompose the Q-value to be the sum of factors that usually depend on a smaller subset of agents. This additional structure allows maximization of the Q-function without enumerating all actions at a state or history (e.g., using variable elimination \citep{Guestrin02ICML}).  
Some other work explores learning more sparse approximations of the factored Q-function \citep{Kok06JMLR} or extends the idea to Monte Carlo tree search and Bayesian RL \citep{Amato15AAAI}. There have been deep extensions of coordination graphs such as assuming payoffs are between pairs of agents \citep{bohmer2020deep} as well as learning and using more general coordination graphs \citep{li2021deep}. 
Sometimes these message passing approaches are called \emph{distributed} since communication between neighbors is allowed during execution. 

Some high-profile methods have also used centralized training and execution such as AlphaStar \citep{AlphaStar} which controls all the units with one centralized agent (unlike the Starcraft Multi-Agent Challenge \citep{SMAC}). The large observation space is dealt with using attention and other methods and the large action space is sampled auto-regressively (sequentially) to exploit the multi-agent structure. 

For empirical analysis, \cite{Gupta17} perform comparisons between various centralized, decentralized (i.e., concurrent), and CTDE through parameter sharing. 

\subsection{Multi-agent transformer (MAT)}
\label{sec:cte:MAT}
The Multi-Agent Transformer (MAT) \citep{MAT} builds on this auto-regressive idea. Specifically, MAT treats the team as a sequence and uses a transformer encoder-decoder to map the agents' observations to the agents' actions at each step. The encoder (critic) attends over all agents' observations to produce a representation and a value estimate for each agent while the decoder (actor) generates actions one agent at a time, conditioning each agent's action on the encoded joint observation as well as the actions already chosen by the agents preceding it. 
MAT doesn't include recurrent layers and conditions on the current joint observation rather than a history, so its policies and value functions are reactive as discussed in Section \ref{sec:singleobs}.\footnote{Also, MAT is described as having a monotonic improvement guarantee like HAPPO, but this is not proven. That guarantee requires updating under the predecessors' updated policies and using the decomposed sequential advantages. MAT doesn't do either.}

MAT's motivation is the same as the one behind coordination graphs. Choosing a joint action by enumeration requires considering $\prod_i |\mathbb{A}_i|$ actions, while choosing them one agent at a time, each conditioned on its predecessors, requires only $\sum_i |\mathbb{A}_i|$ (i.e., additive rather than multiplicative in the number of agents). Coordination graphs obtain this reduction by assuming the value function factors over a graph, while MAT obtains it from the chain rule, appealing to the multi-agent advantage decomposition discussed in Section \ref{sec:sequpdates}. Training is otherwise a standard on-policy actor-critic update using a joint advantage (i.e., a form of centralized PPO). Because each agent's action depends on the other agents' observations (through the encoder) and on their chosen actions (through the decoder), MAT requires centralized execution (even though it is often described as a CTDE method). The authors also evaluate a CTDE variant, MAT-Dec, which keeps the transformer encoder as a centralized critic but replaces the decoder with an independent policy for each agent that conditions only on that agent's own observation. 

Note that there are pros and cons to these approaches (and any approach for that matter). In this case, factoring the value function limits which value functions can be represented, while conditioning each agent's action on its predecessors requires every agent to know what the others have chosen, so the resulting policy needs either centralized execution or communication.

\chapter{Decentralized training and execution (DTE)}
\label{sec:dte}

\section{DTE overview}

Decentralized training and execution (DTE) is the MARL setting that makes the fewest assumptions.  In particular, in the DTE case, at each step of the problem we assume that:
\begin{itemize}
	\item each agent observes (or maintains) its current history, $\hi$, takes an action at that history, $\ai$, and sees the resulting observation, $\oi$,
	\item the (decentralized) algorithm observes the same information ($\hi$,  $\ai$, and $\oi$) as well as the joint reward $r$. 
\end{itemize}
Unlike the CTE case, there is no centralized controller, so the agent must choose actions on its own. In this case, agents only ever observe their own information (actions and observations) and don't observe other agent actions or observations. Similarly, the algorithm can observe the joint reward but (unlike the CTDE case) can only observe the information for that one agent. In DTE, there is strong overlap with the algorithm and agent since the algorithm may specify some aspects of the agent such as exploration but it is still useful to consider them separately to highlight the agent's lack of access to the reward and the optimization role of the algorithm (and to be consistent with other training paradigms). 
It is worth noting that there are some hidden assumptions (such as concurrent learning) that many DTE algorithms make but I will discuss these in more detail below. 

DTE is used in scenarios where centralized information is unavailable or scalability is critical. In particular, due to these limited assumptions, DTE must be used when decentralized agents need to learn without any centralized training phase. That is, if Dec-POMDP agents must learn directly in an environment, agents (and agent optimization algorithms) will not be able to observe centralized information such as other agent actions and observations. As a result, agents will have to learn in a decentralized manner using only the information above. DTE can also be used in settings where the centralization of control, such as in Section \ref{sec:cte}, or algorithms, such as in Section \ref{sec:ctde}, is computationally intractable. This can occur when the joint history (i.e., observation) or joint action spaces become intractably large.

Because the agent and algorithm information are so well aligned, single-agent RL can also be used here. While the CTE case involved transforming the multi-agent problem into a single agent one by combining all the agents' information, the DTE case naturally fits with the single-agent setting by ignoring the fact that other agents exist and learning as if they were the only agent in the problem. I will discuss how such methods can be applied in the value-based and policy gradient case as well as improvements to deal with the multi-agent nature of the problem.

\section{Decentralized, value-based methods}
\label{sec:dte:value}

As discussed in Section \ref{sec:back:value}, value-based methods (such as Q-learning \citep{Watkins1992}) learn a value function and the policy is implicitly defined based on those values (i.e., choosing the action that maximizes the value function). 
Some of the earliest MARL methods are based on applying Q-learning to each agent. 
For instance,
\cite{Claus98AAAI} decompose MARL approaches into \emph{independent learners} (IL) and \emph{joint-action learners} (JAL). ILs learn Q-functions that depend on their own information (ignoring other agents) while JALs learn joint Q-functions, which include the actions of the other agents. These concepts were originally defined for the fully observable, bandit (i.e., stateless) case but can be extended to the partially observable, stateful case. 
For example, we can define an IL Q-function as $\Qi(\hi,\ai)$, which provides a value for agent $i$'s history $\hi$ and action $\ai$. 
A JAL Q-function $\Qi(\jh,\ja)$ for agent $i$ would depend on the joint information $\jh$ and $\ja$. All decentralized value-based MARL methods are ILs since they never have access to the joint information, $\jh$ and $\ja$ and many of the algorithms use the `independent' term in their name. 

I first discuss the tabular (Q-learning based) methods, starting with a simple application of single-agent Q-learning and then discuss improvements to learning for the multi-agent setting. Next, I discuss extensions to the deep case (DQN-based), again in terms of a simple application of DRQN as well as issues and fixes for the multi-agent case. 

\subsection{Independent Q-learning (IQL)}

IQL is one of the simplest MARL methods. It just applies Q-learning to each agent independently, treating other agents as part of the environment. That is, each agent learns its own local Q-function, $Q_i$, using Q-learning on its own data. Pseudocode for a version of IQL for the Dec-POMDP case is provided in Algorithm \ref{alg:IQL}. Just like Q-learning in the single-agent case, that algorithm uses a learning rate, $\alpha$, and some form of exploration, which in this case is $\epsilon$-greedy. Each agent initializes their Q-function (e.g., to 0, optimistically, etc.), and then iterates over episodes (either a fixed set or until convergence). During each episode, the agent chooses an action with exploration, the algorithm sees the joint reward and the agent's own observation and then the algorithm updates the Q-value at the current history, $\hi$, given the current Q-estimates and the reward using the standard Q-learning update (line 9). The history is updated by appending the action taken and observation seen to the current history and the episode continues until the horizon is reached. 

The Q-functions are inherently tied to the history length, which defines the current time step, making the algorithm suitable for finite-horizon cases. Regardless, the algorithm can easily be extended to the episodic/indefinite-horizon case by including terminal states (or histories) instead of a fixed horizon. 

\begin{algorithm}
\begin{algorithmic}[1]
\STATE set $\alpha$ and $\epsilon$ (learning rate, exploration)
\STATE Initialize $\Qi$ for all $\hi \in \hAOHS  i, \ai \in \aAS i$
\FORALL{episodes}
\STATE $\hi \gets \emptyset$ \hfill  \COMMENT{Empty initial history}
\FOR{$\ts=0$ to $\hor-1$}
\STATE Choose $\ai$ at $\hi$ from $\Qi(\hi,\cdot)$ with exploration (e.g., $\epsilon$-greedy)
\STATE See joint reward $r$, local observation $\oi$\hfill \COMMENT{Depends on joint action $\ja$}
\STATE $\hi' \gets \hi\ai\oi$
\STATE $\Qi(\hi,\ai) \gets \Qi(\hi,\ai) + \alpha\left[r+ \gamma \max_{\ai'} \Qi(\hi',\ai')-\Qi(\hi,\ai)\right] $  
\STATE $\hi \gets \hi'$
\ENDFOR
\ENDFOR
\RETURN $\Qi$ 
\end{algorithmic} 
\caption{Independent Q-Learning for agent $i$ (finite-horizon)}
\label{alg:IQL}
\end{algorithm}

\paragraph{Important hidden information}

This algorithm runs in the underlying Dec-POMDP but the details of those dynamics in the environment are not shown (as they are not visible to the agent and algorithm). For example, at the beginning of each episode, the state, $s$, is sampled from the initial state distribution $\bO$, and when agents take joint action $\ja$ the reward is generated from the reward function, $R(s,\ja)$, the next state $s'$ is sampled from the transition dynamics with probability $\Pr(s'|s,\ja)$, and the joint observation is sampled from the observation function  $\Pr(\jo|s',\ja)$. This process continues until the end of the episode.

While independent Q-learning ignores the other agents, it still depends on them in order to generate the joint reward $r$ and the local observation $o_i$ since both of these depend on the actions of all agents. 
That is, from agent $i$'s perspective, it is learning in a (single-agent) POMDP, as Algorithm \ref{alg:IQL} shows (the states of the POMDP would be the states of the Dec-POMDP plus the histories of the other agents).\footnote{ 
The idea of fixing other agent policies and then solving the resulting POMDP (either by planning methods \citep{Nair03IJCAI} or RL methods \citep{Banerjee12}) has had success and results in a best response for agent $i$. Furthermore, iterating over agents, $i$, fixing other agent policies, $\neg i$,  and calculating best responses will lead to a local optimum of the Dec-POMDP \citep{Banerjee12}. Unfortunately, such methods require coordination to communicate which agent's turn it is to learn so they are not appropriate for the DTE case.  }

In particular, as pointed out in previous papers \citep{Lauer00ICML}\footnote{Note the paper considers the fully observable deterministic transition case but I extend the idea to the Dec-POMDP case.}, independent Q-learning is actually learning a Q-function for each agent based on the actions selected by the other agents during training. That is, agent $i$ would use data that depends on the actions taken by the other agents (and observations other agents see) even if it doesn't directly observe those actions (and observations).  In the partially observable case, the following Q-function would be learned:
\begin{equation}
Q_i(\hi,\ai) = \sum_{\ja \in \jaS} \hat P(\ja,\jh|\hi,\ai) \left[ r+\gamma \sum_{\oi} \hat P(\oi|\jh,\ja) \max_{\ai'} Q_i(\hi',\ai')\right]
\label{eq:iqltrue}
\end{equation}
where $\hat P(\ja,\jh|\hi,\ai)$ is the empirical probability that the joint action $\ja$ and joint history $\jh$ occurs when agent $i$ selects its action $\ai$ in $\hi$. This is precisely when independent Q-learning will update $Q_i(\hi,\ai)$ and it will use the joint reward $r$ and the agent's observation function $\hat P(\oi|\jh,\ja)$, which approximates $P(\oi|\ja,s')P(s'|\ja,s)$, where $P(\oi|\ja,s')$ marginalizes out other agent observations from the joint function $P(\jo|\ja,s')$.
$\hat P(\oi|\jh,\ja)$ is the empirical probability based on observing $\oi$ from $\jh$ after taking action $\ja$. 
Therefore, IQL is assuming the other agents are part of the environment and learning a Q-function for the resulting POMDP. If the other agents are not learning and do in fact have fixed policies, the problem would just reduce to this POMDP. Unfortunately, the other agents would also typically be learning, leading to \emph{nonstationarity} in the underlying POMDP and difficulty with coordinated action selection. 
 
 \paragraph{Convergence and solutions}
In the case of IQL, the underlying POMDP that is being learned is nonstationary since the other agents are also learning and thus changing their policies over time.  
 This may cause IQL to not converge \citep{Tan93}. Convergence of Q-learning in the multi-agent case is an active area of research \citep{Claus98AAAI,Tuyls03,Wunder10,Hussain23} and it is still an open question what assumptions will allow convergence and to what (unlike the policy gradient case where convergence to a local optimum is assured under mild assumptions \citep{Peshkin00UAI,JAIR23})\footnote{In fact, while Q-learning can provably converge to a global optimum in the single-agent case \citep{Jaakkola93}, there are no known MARL algorithms that have guaranteed convergence to a global optimum in the partially observable case. This is an interesting area for future research!}. 

It is worth pointing out that even if algorithms can converge to an optimal Q-value (or any Q-value), agents may not be able to select optimal actions if there are multiple optimal policies  \citep{Lauer00ICML}. In this case, agents would need to coordinate on their actions to make sure they choose actions from the same joint policy but this is not possible with only individual Q-values (e.g., if $\QA 1(\aoHistA 1,\aAI 1 1)=\QA 1(\aoHistA 1,\aAI 1 2)$ and $\QA 2(\aoHistA 2,\aAI 2 1)=\QA 2(\aoHistA 2,\aAI 2 2)$ but $\jQ(\aoHistA 1,\aoHistA 2,\aAI 1 1,\aAI 2 2)=\jQ(\aoHistA 1, \aoHistA 2,\aAI 1 2,\aAI 2 1)<\jQ(\aoHistA 1, \aoHistA 2,\aAI 1 2,\aAI 2 2)=\jQ(\aoHistA 1,\aoHistA 2,\aAI 1 1,\aAI 2 1)$---agents must take the same actions to be optimal)

IQL is very simple but can perform well \citep{Tan93,Sen94,Tampuu}. It is also the basis for the other value-based DTE MARL methods.

\subsection{Improving the performance of IQL}

Several extensions of IQL have been developed to improve performance and better fit with the MARL setting. 
While the below algorithms were developed for the fully observable case (the multi-agent MDP described in Section \ref{sec:cte}), I extend them to the partially observable case here to fit better with our Dec-POMDP setting.

\subsubsection{Distributed Q-learning}

In order to promote coordination, Distributed Q-learning \citep{Lauer00ICML} makes strong optimistic assumptions, prioritizing high-reward outcomes from agent actions while discounting suboptimal ones during learning. In particular, extending to the partially observable case, it uses the following Q-update:
\begin{equation}
Q_i(\hi,\ai) \gets  \max \left\{ Q_i(\hi,\ai),  r+\gamma \max_{\ai'} Q_i(\hi',\ai') \right\}
\label{eq:distq}
\end{equation}
where the current Q-value is kept if it is higher than (or equal to) the new estimate $r+\gamma \max_{\ai'} Q_i(\hi',\ai')$. Otherwise, the Q-value is updated to be the new estimate. 
The intuition behind this optimistic update is that other agents will make mistakes during training because they have suboptimal policies or because they are exploring. Agents don't necessarily want to learn from these mistakes. Instead, agents want to learn from the best policies of the other agents such as those that choose actions that generate high-reward values, even if they don't happen very often. 
In the special case of deterministic MMDPs, this approach will learn the same Q-function as centralized Q-learning (which will converge to the optimal Q-values in the limit). 
This is unlike IQL which can get stuck in local optima or never converge at all. 

Because Distributed Q-learning learns individual Q-values, agents may still not be able to select optimal actions due to the coordination issue mentioned above. As a result, Distributed Q-learning stores the current best policy for each agent (i.e., if the Q-value is updated, the action that resulted in that update is stored for that state in the fully observable case or history in the partially observable case). Including this coordination mechanism, the algorithm can converge to optimal Q-functions and optimal policies in the limit for the fully observable deterministic case. 

In the more general stochastic case, distributed Q-learning is not able to distinguish between environmental stochasticity and agent stochasticity. That is, if getting a high-valued reward is unlikely (due to reward or transition stochasticity) distributed Q-learning will assume it can be received deterministically using max rather than taking the expectation (over other agent policies and stochastic transitions) as in Equation \ref{eq:iqltrue}. 
As a result, distributed Q-learning is overly optimistic in the stochastic case and may perform poorly. Learning and incorporating the probabilities in Equation \ref{eq:iqltrue} and then calculating the expectation over the next step Q-value could fix this problem but would assume agents observe the actions of other agents which is not the case for decentralized training. 

\subsubsection{Hysteretic Q-learning}

To maintain the idea of optimism while accounting for stochasticity, hysteretic Q-learning \citep{Matignon07} was developed. The idea is to use two learning rates, $\alpha$ and $\beta$ with $\alpha > \beta$. That is, agents update their Q-values more during a positive experience than a negative one. 

In particular, if we define the TD error as:
$$\delta \gets r + \gamma \max_{a_i'} Q(\hi',a_i')-  Q_i(\hi,\ai),\\$$
then we can generalize distributed Q-learning (using a learning rate $\alpha$) as:
\begin{equation}
  Q_i(\hi,\ai) \gets  \begin{cases}
      Q_i(\hi,\ai) + \alpha\delta  \quad \text{if} \quad \delta > 0\\
     Q_i(\hi,\ai) \quad \text{else}
    \end{cases}
\end{equation}
highlighting the fact that positive experiences (in terms of TD error) cause updates to the Q-function while negative ones do not. 

Hysteretic Q-learning, in contrast, adds the second learning rate $\beta$ that is applied to the negative experiences: 
\begin{equation}
\label{eq:hys}
  Q_i(\hi,\ai) \gets  \begin{cases}
      Q_i(\hi,\ai) + \alpha\delta  \quad \text{if} \quad \delta > 0\\
     Q_i(\hi,\ai) + \beta\delta \quad \text{else}
    \end{cases}
\end{equation}
This ensures learning still occurs in the negative cases but the update is less than in the positive cases because $\alpha > \beta$. Therefore, hysteretic Q-learning is still optimistic (hoping the positive experiences are because of better action choices and not stochasticity) but is less optimistic than distributed Q-learning. 
The implementation is also very simple (although it does require tuning two learning rates). 
Hysteretic Q-learning is very simple but can perform well in a range of domains \citep{Matignon07}. 
Also, note that as $\beta$ approaches $\alpha$, hysteretic Q-learning will become standard Q-learning and as $\beta$ approaches 0 and $\alpha$ approaches 1, it will become distributed Q-learning.

\subsubsection{Lenient Q-learning}

While it is possible to adjust $\beta$ towards $\alpha$ as learning continues, to be more optimistic at the beginning of learning (when a lot of exploration is happening) and still be robust to stochasticity as learning converges, lenient Q-learning \citep{lenient} allows agents to adjust this \emph{leniency} towards exploration on a history-action pair basis. Specifically, lenient Q-learning maintains a temperature, $T(\hi, \ai)$, per history-action pair that adjusts the probability that a negative update at a particular history-action pair will be ignored. These temperature values allow history-action pairs that have not been visited often to not be updated yet, while frequently visited history-action pairs are updated as in IQL. 

The lenient Q-learning update is:
\begin{equation}
\label{eq:lenient}
  \Qi(\hi,\ai) \gets  \begin{cases}
      \Qi(\hi,\ai) + \alpha\delta  \quad \text{if} \quad \delta > 0 \quad \text{or}\quad rand \sim U(0,1) > 1 - e^{-KT(\hi,\ai)}\\
     \Qi(\hi,\ai)  \quad\quad\quad  \text{else}
    \end{cases}
\end{equation}
$T(\hi,\ai)$ is initialized to a maximum temperature, and is decayed after an update with\\ $T(\hi,\ai)~\gets~\lambda T(\hi,\ai)$ for $\lambda \in (0,1)$ and leniency parameter $K$ adjusts the impact of the temperature on the probability distribution. The resulting Q-update will probabilistically update Q based on how often history-action pairs are visited, with more frequently visited pairs being more likely to be updated when the TD error is negative. 

Lenient Q-learning can outperform hysteretic Q-learning by fine-tuning optimism levels on a per-state-action basis (or per-history-action basis in this case). Unfortunately, it also requires maintaining (and updating) temperature values as well as Q-values. Maintaining these values can be more problematic in the partially observable case since the history-action space grows exponentially with the problem horizon (or history length more generally).

\subsection{Deep extensions, issues, and fixes}

In order to scale to larger domains, deep extensions of the above algorithms have been developed. 
These approaches are typically based on deep Q-networks (DQN)~\citep{DQN}. While DQN can scale to larger action and observation spaces, the decentralized multi-agent context can cause challenges such as non-stationarity and coordination difficulties. I discuss a deep version of IQL, independent DRQN \citep{Tampuu}, the coordination issues with the approach, and some proposed fixes. Details about DQN and DRQN are in Section \ref{sec:back:value}. 

\begin{figure}
\centering
\includegraphics[height=0.2\linewidth]{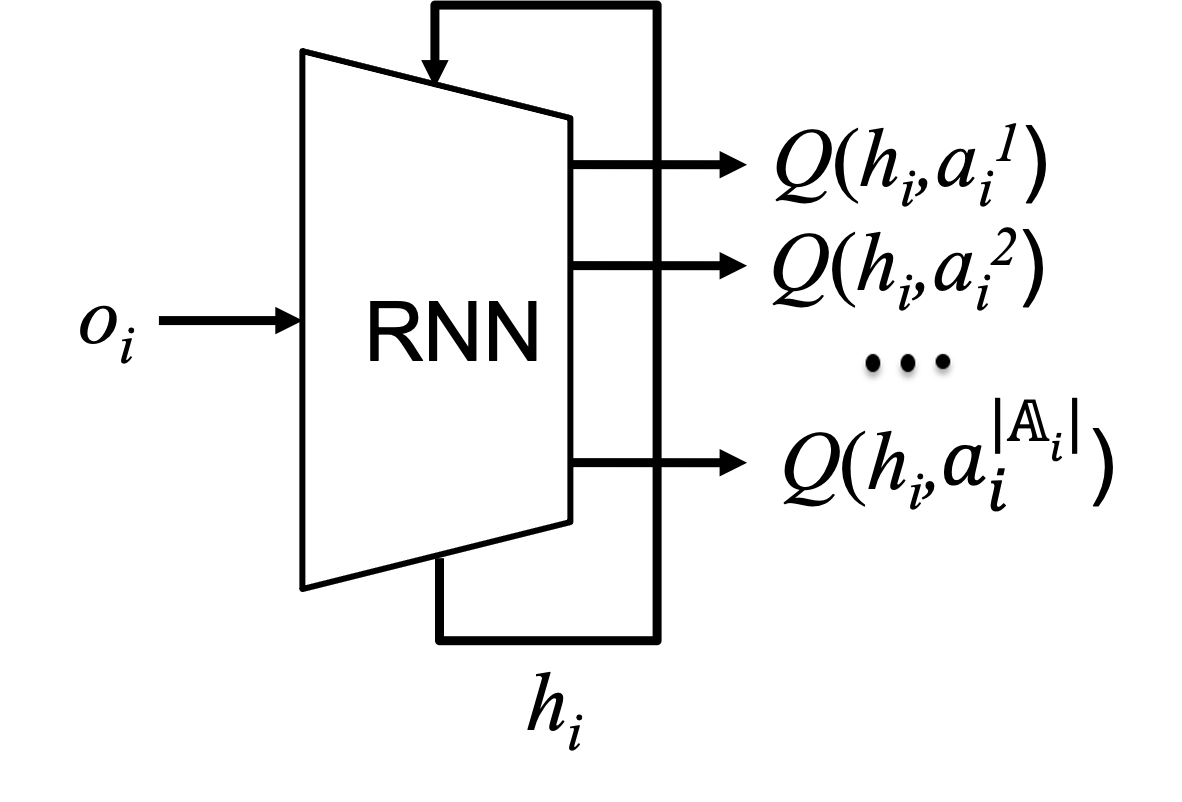}
\caption{Independent DRQN diagram.}
\label{fig:IDRQN}
\end{figure}

\subsubsection{Independent DRQN (IDRQN)}
\label{sec:value:IDRQN}

Independent DRQN (IDRQN) \citep{Tampuu}\footnote{Again, the original algorithm only considers the fully observable case but I consider the extension  to the Dec-POMDP case.} combines IQL and DRQN. As seen in Algorithm \ref{alg:IDRQN}, the basic structure is the same as IQL but $\Qi$ is estimated using an RNN parameterized by $\theta$ (shown in Figure \ref{fig:IDRQN}), along with using a target network, $\theta^-$, and a replay buffer, $\mathcal{D}$.  First, episodes are generated by interacting with the environment (e.g., $\epsilon$-greedy exploration), which are added to the replay buffer. This buffer typically has a fixed size and new data replaces old data when the buffer is full. Episodes can then be sampled from the replay buffer (either single episodes or as a minibatch of episodes). In order to have the correct internal state, RNNs are trained sequentially.  Training is done from the beginning of the episode until the end\footnote{It is common in practice to use a fixed history length and sample windows of that length rather than training with full episodes.}, calculating the internal state and the loss at each step  and updating $\theta$ using gradient descent. The target network $\theta^-$ is updated periodically (every $C$ episodes in the code) by copying the parameters from $\theta$.

\begin{algorithm}
\begin{algorithmic}[1]
\STATE set $\alpha$, $\epsilon$, and $C$ (learning rate, exploration, and target update frequency)
\STATE Initialize network parameters $\theta$ and $\theta^-$  for $\Qi$
\STATE $\mathcal{D}_i \gets \emptyset$
\STATE $e \gets 1$ \hfill \COMMENT{episode index}
\FORALL{episodes}
\STATE $\hi \gets \emptyset $ \hfill \COMMENT{initial history is empty}
\FOR{$\ts=0$ to $\hor-1$}
\STATE Choose $\ai$ at $\hi$ from $\Qi^\theta(\hi,\cdot)$ with exploration (e.g., $\epsilon$-greedy)
\STATE See joint reward $r$, local observation $\oi$\hfill \COMMENT{Depends on joint action $\ja$}
\STATE append $\ai,\oi,r$ to $\mathcal{D}_i^e $
\STATE $\hi \gets \hi\ai\oi$ \hfill \COMMENT{update RNN state of the network}
\ENDFOR
\STATE sample an episode from $\mathcal{D}_i$
\STATE $\hi \gets \emptyset $ 
\FOR{$\ts=0$ to $\hor-1$}
\STATE $\ai, \oi, r \gets \mathcal{D}^e_i(\ts)$
\STATE $\hi' \gets \hi\ai\oi$ 
\STATE $y=r + \gamma\max_{\ai'}\Qi^{\theta^-}(\hi',\ai')$
\STATE  Perform gradient descent on parameters $\theta$ with learning rate $\alpha$ and loss:  $ \big(y - \Qi^{\theta}(\hi, \ai)\big)^2  $
\STATE $\hi \gets \hi'$
\ENDFOR
\IF{$e \mod C = 0$} 
    \STATE $\theta^- \gets \theta$ 
 \ENDIF
 \STATE $e \gets e+1$
\ENDFOR
\RETURN $\Qi$ 
\end{algorithmic} 
\caption{Independent DRQN (IDRQN) for agent $i$ (finite-horizon*)}
\label{alg:IDRQN}
\end{algorithm}

Again, the code can be adjusted to the episodic/indefinite-horizon case by replacing the horizon with terminal states or the infinite-horizon case by removing the loop over episodes. 

\paragraph{Issues with IDRQN}

Independent DRQN can perform well but has a number of issues. 
For example, early methods disabled experience replay \citep{Foerster16}\footnote{Other methods for improving experience replay in MARL exist, but they typically assume the CTDE setting (such as \cite{Foerster17}).}. They found that performance was worse with experience replay than without it. This is because agents are no longer learning concurrently when using the replay buffer---experience is generated from the behavior of all agents (i.e., concurrently) but is then put into the agent's buffer for learning. When learning, different agents will sample from their buffers separately, learning from different experiences. Learning from this different data makes performance noisy and unstable. This makes intuitive sense since agents will update their Q-functions (and resulting policies) based on (potentially very) different experiences.  I discuss a simple fix next. 

\subsubsection{Deep hysteretic-Q and concurrent buffer}
\label{sec:Dec-HDRQN}

Hysteretic Q-learning has been extended to the deep case in the form of decentralized hysteretic deep recurrent Q-networks (Dec-HDRQN) \citep{Omidshafiei17}. Dec-HDRQN makes two main contributions: the use of hysteresis with D(R)QN, and a concurrent replay buffer to improve performance.\footnote{The paper also discusses the multi-task case, but I don't discuss that aspect here.}

The inclusion of hysteresis in Dec-HDRQN just combines hysteresis and IDRQN. That is, since $y - \Qi^{\theta}(\hi, \ai)$ in the IDRQN loss is the TD error, $\delta$, the network parameters are updated (using gradient descent) using learning rate $\alpha$ if $\delta > 0$, and $\beta$ is used otherwise. 

\begin{figure}
\begin{centering}
\includegraphics[scale=0.35]{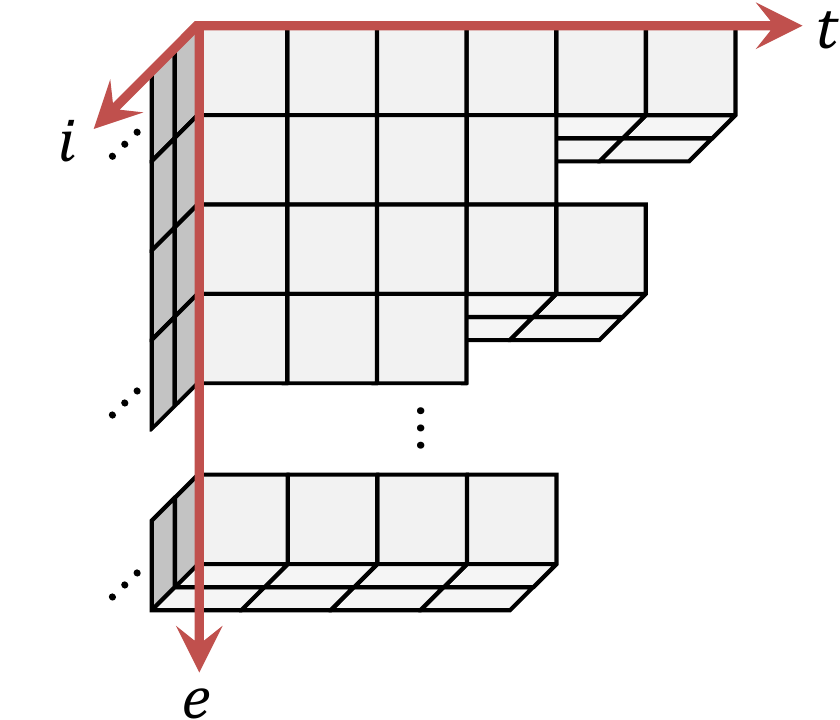}
\par\end{centering}
\caption{Concurrent experience replay trajectories (CERTs) (from \cite{Omidshafiei17})}
\label{fig:CERT}
\end{figure}

The concurrent buffer is also relatively straightforward. 
Concurrent experience replay trajectories (CERTs), are based on the idea of creating a joint experience replay buffer and then sampling from that buffer. 
In particular, we would like to have a joint replay buffer with joint experience tuples, $ \jaIT e 1, \joIT e 1, \rIT e 1,  \ldots, \jaIT e \ts, \joIT e \ts, \rIT e \ts$, where the superscript here represents the episode number, $e$. 
Agents could then sample from this joint buffer to get the local information from the same episode and time step: $\hi, \ai, \oi, r $ (where $\hi$ represents the history at the current time step such as the internal state of agent $i$'s RNN after passing the local history of actions and observations until the time step being sampled). 
Of course, in a decentralized learning context, agents will not have access to such a joint buffer, but agents can store their part of the joint buffer by indexing by episode $e$ and timestep $t$. That is, as seen in Figure \ref{fig:CERT}, agents store their local trajectories along axes for $e$ and $t$. When an $e$ and $t$ are sampled, each agent $i$ will sample data from the same episode and timestep, producing the same data as would be produced with the joint buffer. As long as agents have the same size buffers and coordinate on the seeds for the random number generator, this approach can be used for decentralized learning. 

\subsubsection{Deep lenient Q-learning}

Lenient Q-learning has been extended to the \emph{fully observable} deep case in the form of lenient DQN \citep{Palmer18}. 
In this case, the leniency values are added to the replay buffer of DQN at each step as: $(s_t, a_t,r_t,s_{t+1},l(s_t, a_t) )$
 where $l(s_t, a_t)=1 - e^{-KT(\phi(s_t),a_t)}$. To scale to large state spaces, the approach clusters states using an autoencoder which outputs a hash $\phi(s)$ for a given state $s$. 
The method also introduces a temperature decay schedule after reaching the terminal state and an improved exploration scheme based on average temperature value. 

The approach was shown to perform well in a set of coordinated multi-agent object transportation problems (CMOTPs). It was not extended to the partially observable case but it should be possible by using a method that clusters histories rather than states. 
 
\subsubsection{Likelihood Q-learning}

Distributional Q-learning \citep{bdr2023} has also been used to improve decentralized MARL. Instead of just learning the expected Q-value as in traditional Q-learning, distributional RL learns the distribution of returns that are possible due to stochasticity. Likelihood Q-learning \citep{Lyu20} uses the return distribution to determine the likelihood of the experience (i.e., $\hi, \ai, \oi, r$) given the current estimate, which they call Time Difference Likelihood (TDL). This information can be used to help determine if other agents are exploring (i.e., low likelihood and low value). As a result, the TDL can be used as a multiplier of the learning rate, causing larger updates for more likely experiences---when other agents are not exploring. 

But some low-likelihood experiences may be good to learn from, such as when agents have found a high-quality joint action (i.e., low likelihood but high value).  Distributional information can then be used to estimate risk \citep{morimura2010nonparametric}. In the multi-agent context risk-seeking behavior can mean being optimistic as in hysteretic or lenient learning. 

Each of these ideas can be used in isolation or in combination and they can be used with other decentralized MARL methods to further improve performance. For instance, the combination of Dec-HDRQN (see Section~\ref{sec:Dec-HDRQN}) with TDL was shown to outperform Dec-HDRQN and forms of lenient Q-learning. Incorporating risk can further improve performance.

\section{Decentralized policy gradient methods}
\label{sec:dte:pg}

Single-agent policy gradient and actor-critic methods can similarly be extended to the decentralized training multi-agent case by allowing each agent to learn separately.
These approaches can scale to larger action spaces and have stronger (local) convergence guarantees than the value-based counterparts above. Background on single-agent policy gradient methods is in Section \ref{sec:back:pg}. 

Policy gradient methods can be used with continuous-action or stochastic policies. 
For all approaches, I will assume that we have a stochastic policy for each agent parameterized by $\ppi$, where $\poli^{\ppi}(\ai| \hi)$ represents the probability agent $i$ will choose action $\ai$ given the history $\hi$ and parameters, $\ppi$, $\Pr(\ai| \hi,\ppi)$. This is in contrast to the value-based methods in Section \ref{sec:dte:value}, where deterministic policies were generated based on the learned value functions. 

\subsection{Decentralized REINFORCE}
\label{sec:dte:peshkin}

As mentioned in Section \ref{sec:back:pg}, 
REINFORCE \citep{Williams92ML} is one of the oldest (and simplest) policy gradient methods. The basic idea is to estimate the value of the policy using Monte Carlo rollouts and then update the policy using gradient ascent. 
REINFORCE was extended to the cooperative multi-agent case by \cite{Peshkin00UAI}. 
Importantly, \cite{Peshkin00UAI} also showed that the joint gradient can be decomposed into decentralized gradients. That is, decentralized gradient ascent will be equivalent to joint gradient ascent. 
Convergence to a local optimum is guaranteed when agents are learning concurrently (each agent is learning on the data that is generated by all agents and all agents make updates at the same time) because they will apply the same algorithm to the same data at synchronized time steps. 
This convergence result is in contrast to the value-based methods, which do not have strong convergence guarantees. These convergence guarantees extend to many other policy gradient methods (as discussed below). 

Pseudocode for a version of decentralized REINFORCE is in Algorithm \ref{alg:Peshkin}. Here, a set of stochastic policies, $\poli$, that are parameterized by $\ppi$ (e.g., parameters of a neural network), are learned.  A learning rate of $\alpha$ is used (as usual). The method iterates for some number of episodes, choosing actions from the stochastic policy (which has positive probability for all actions to ensure proper exploration), observing the joint reward and local observation at each time step, and storing the episode. Then,  the return is computed for each step of the episode and the parameters of the policy are updated using the returns and gradient (just like the single-agent version of REINFORCE). Time steps are explicitly represented in the code to make return calculation clear. 
Again, the method can be extended to the episodic/indefinite-horizon case by replacing the horizon with terminal states and it isn't well-defined for the infinite-horizon case (since updating takes place at the end of episodes).

\begin{algorithm}
\begin{algorithmic}[1]
\REQUIRE Individual actor models $\poli(\ai| \hi)$, parameterized by $\ppi$
\STATE set $\alpha$  (learning rate)
\FORALL{episodes}
	\STATE  $\hiT 0 \gets \emptyset$ \hfill  \COMMENT{Empty initial history}
	\STATE $ep \gets \emptyset$ \hfill  \COMMENT{Empty episode}
	\FOR{$\ts=0$ to $\hor-1$}
		\STATE Choose $\aiT \ts$ at $\hiT \ts$ from $\poli(\ai| \hiT \ts)$
		\STATE See joint reward $\rT \ts$, local observation $\oiT \ts$\hfill \COMMENT{Depends on joint action $\ja$}
		\STATE append $\aiT \ts,\oiT \ts,\rT \ts$ to $ep $
		\STATE $\hiT {\ts+1} \gets \hiT \ts \aiT \ts \oiT \ts$\hfill \COMMENT{Append new action and obs to previous history}
	\ENDFOR
	\FOR{$\ts=0$ to $\hor-1$}
        		\STATE Compute return at $\ts$ from $ep$: $G_{\ts} \gets \sum_{k=\ts}^{\hor-1} \gamma^{k-\ts} \rT k $
		 \STATE Update parameters: $\ppi \gets \ppi + \alpha \gamma^t G_{\ts} \nabla \log \poli(\ai| \hiT \ts)$
        \ENDFOR
\ENDFOR
\end{algorithmic}
\caption{Decentralized REINFORCE for agent $i$ (finite-horizon)}
\label{alg:Peshkin}
\end{algorithm}

Decentralized REINFORCE isn't widely used but serves as a basis for the later policy gradient approaches as well as theoretical guarantees.

\subsection{Independent actor critic (IAC)}
\label{sec:dte:iac}

Since REINFORCE is a Monte Carlo method, it must wait until the end of an episode to update. Furthermore, it will typically be less sample efficient than methods that learn a value function. As mentioned in Section \ref{sec:back:pg}, actor-critic methods have been developed to combat these issues by learning a policy representation (the actor) as well as a value function (the critic) \citep{SuttonBarto18}. Actor-critic methods can use the critic to evaluate the actor without waiting until the end of the episode and TD-learning to update the critic more efficiently. 

Independent actor-critic (IAC) was first introduced by \cite{COMA} and it can be thought of as a combination of IQL (or IDRQN) and decentralized REINFORCE.
Each agent learns a policy, $\poli(\ai| \hi)$, as well as a value function such as $\Qi(\hi,\ai)$. 
In the simplest case of IAC (where a Q-value is used for the critic), the gradient associated with the policy parameters $\ppi$ can be derived as
\begin{equation}
   \nablai J =  \mathbb{E}_{\jh, \ja\sim\mathcal{D}} \left[ \Qi(\hi, \ai) \nablai \log\poli^{\ppi}(\ai|\hi) \right] \,, \label{eq:IAC:actor}\footnote{For simplicity and to match common implementations, we ignore discounting but the true gradient would include it \citep{nota2020policy}.}
\end{equation}
assuming a stochastic policy for each agent $i$ that is parameterized by $\ppi$, $\poli^{\ppi}(\ai| \hi)$. Histories and actions are sampled according to the on-policy (discounted) visitation probability 
 (which is a joint visitation probability since concurrent sampling is assumed). This gradient is very similar to the REINFORCE one but uses a Q-value rather than the Monte Carlo return (and is shown here with the expectation rather than just a sample).

In the on-policy, function-approximation (e.g., deep) case, the critic can be updated as: 
\begin{equation}
    \mathcal{L}(\vpi)=\mathbb{E}_{\langle \hi,\ai,r,\oi \rangle \sim\mathcal{D}}\Big[\big(y - \Qi^{\vpi}(\hi, \ai)\big)^2 \Big] \text{, where\,\, } y=r + \gamma \Qi^{\vpi}(\hi',\ai')
    \label{eq:IAC:critic}
\end{equation}
where $\hi'=\hi\ai\oi$ and $\ai'$ is the action sampled at the next time step using $\poli^{\ppi}(\ai'|\hi')$ since it is an on-policy estimate (rather than the max action used in DQN). 
 $\Qi^{\vpi}(\hi,\ai)$ represents the Q-function estimate for agent $i$ using parameters $\vpi$ (just like in Section \ref{sec:dte:value}). 

Pseudocode for IAC is given in Algorithm \ref{alg:IAC}. 
Because the value function is no longer being used to select actions but rather just to evaluate the policy, it is common to instead learn $\Vi(\hi)$. And because a model of $V$ is learned, I denote it $\hat V$. 
Like the REINFORCE case, the algorithm loops for some number of episodes, choosing actions from the stochastic policy and seeing the joint reward and local observation at each step. Now, during the episode, the TD error is calculated and used to update the actor and the critic using the associated gradients described above. 
Since V-values are learned instead of Q-values, the TD error is calculated with  $ \rT  \ts + \gamma \hat \Vi(h_{i,t+1}) - \hat \Vi(h_{i,t})$ rather than $\Qi$. 
As a result, a value-based version of Equation \ref{eq:IAC:actor} is used to update the actors. 
A value-based version of Equation \ref{eq:IAC:critic} is recovered by using $y= r_t + \gamma \hat \Vi(h_{i,t+1}) $ and replacing  $\Qi^{\vpi}(\hi, \ai)$ with $\hat \Vi(h_{i,t})$. 

\begin{algorithm}
\caption{Independent Actor-Critic (IAC) (finite-horizon)}
\begin{algorithmic}[1]
\label{alg:IAC}
\REQUIRE Individual actor models $\poli(\ai| \hi)$, parameterized by $\ppi$
\REQUIRE Individual critic models $\hat \Vi(h)$, parameterized by $\vpi$

\FORALL{episodes}
	\STATE  $\hiT 0 \gets \emptyset$ \hfill  \COMMENT{Empty initial history}
	\FOR{$\ts=0$ to $\hor-1$}
		\STATE Choose $\aiT \ts$ at $\hiT \ts$ from $\poli(\ai| \hiT \ts)$
		\STATE See joint reward $\rT \ts$, local observation $\oiT \ts$\hfill \COMMENT{Depends on joint action $\ja$}
		\STATE $\hiT {\ts+1} \gets \hiT \ts \aiT \ts \oiT \ts$\hfill \COMMENT{Append new action and obs to previous history}
	 \STATE Compute value TD error: $\delta_{i,t} \gets \rT  \ts + \gamma \hat \Vi(\hiT {\ts+1}) - \hat \Vi(\hiT \ts) $
        \STATE Compute actor gradient estimate: $ \gamma^t \delta_{i,t} \nabla\log\poli(a_{i,t}| \hiT \ts)$
        \STATE Update actor parameters $\ppi$ using gradient estimate (e.g., $\ppi \gets \ppi + \alpha \gamma^t \delta_{i,t} \nabla\log\poli(a_{i,t}| \hiT \ts)$)
        \STATE Compute critic gradient estimate: $ \delta_{i,t} \nabla \hat \Vi(\hiT \ts)$
        \STATE Update critic parameters $\vpi$ using gradient estimate (e.g., $\vpi \gets \vpi + \beta  \delta_{i,t} \nabla \hat \Vi(\hiT \ts)$)
	\ENDFOR
\ENDFOR
\end{algorithmic}
\end{algorithm}

Like the other algorithms, IAC can be adapted to the episodic or infinite-horizon case by including terminal states or removing the loop over episodes.   
IAC is a simple baseline that is often used and serves as the basis of many other decentralized MARL actor-critic methods.

\subsection{Other decentralized policy gradient methods}

Any single-agent method can be extended to the decentralized MARL case. Typically, it just entails including recurrent layers to deal with partial observability and then training agents concurrently to generate joint data and synchronize updates. 
One notable example is IPPO \citep{de20,MAPPO}. 
IPPO is the extension of single-agent PPO \citep{PPO} to the multi-agent case.
Forms of IPPO have been shown to outperform many state-of-the-art MARL methods. 
It is worth noting that the standard implementations of IPPO (and the impressive results) use parameter sharing (as discussed below) so they are not technically decentralized but decentralized implementations can still perform well \citep{MAPPO}. 
I discuss IPPO (and MAPPO) in more detail in Section \ref{sec:ctde:pg:mappo} and discuss parameter sharing below.

\section{Other topics}
\label{sec:dte:other}

\paragraph{Concurrent learning}
It is very important to note that all of the methods in this section assume agents are running the same algorithm and perform synchronous updates (i.e., at the same time on the same data). Without these assumptions, methods that have convergence guarantees lose them but may still work well in some domains. As discussed in relation to DQN (see Section \ref{sec:value:IDRQN}), breaking this `concurrent' learning assumption can cause methods to perform poorly.

\paragraph{Parameter sharing}

Parameter (or weight) sharing, where agents share the same network for estimating the value function in value-based methods or policies (and value functions) in policy gradient methods, is common in cooperative MARL. 
The data from each agent can be used to update a single value network for an algorithm such as DRQN. Agents can still perform differently due to observing different histories and agent indices can be added to increase specialization \citep{Gupta17,Foerster16}. 
While parameter sharing is typically used with homogeneous agents (i.e., those with the same action and observation spaces), it can also be used with heterogeneous agents \citep{terry2020revisiting}.
Since parameter sharing requires agents to share networks during training, it isn't strictly DTE as it couldn't be used for online training in a decentralized fashion. Nevertheless, any of the algorithms in this chapter can be extended to use parameter sharing. The resulting algorithms would be a form of CTDE as the shared network would represent a centralized component. Nevertheless, these parameter-sharing implementations may be more scalable than other forms of CTDE, are often simple to implement, and can perform well \citep{Gupta17,COMA,MAPPO}.

\paragraph{Relationship with CTDE}

While CTDE methods are (by far) the most popular form of MARL, they do not always perform the best. In fact, since DTE methods typically make the concurrent learning assumption, they are actually quite close to CTDE methods. 
This is shown explicitly in the policy gradient case as the gradient of the joint update is the same as the decentralized gradient (as described above in Section \ref{sec:dte:peshkin})~\citep{Peshkin00UAI}. This phenomenon has also been studied theoretically and empirically with modern actor-critic methods \citep{AAMAS21Luke,JAIR23}. It turns out that while centralized critic actor-critic methods are often assumed to be better than DTE actor-critic methods, they are theoretically the same (with mild assumptions) and often empirically similar (and sometimes worse). Too much centralized information can sometimes be overwhelming, harming scalability of centralized-critic methods \citep{MAPPO,JAIR23}. 
Furthermore, some CTDE methods, such as using a state-based critic, are fundamentally unsound and can often hurt performance in partially observable domains \citep{AAAI22,JAIR23}. More details about these issues can be found in Sections \ref{sec:ctde:pg:critics} and \ref{sec:ctde:pg:states}. 
Of course, there are many different ways of performing centralized training for decentralized execution, and studying the differences and similarities between DTE and CTDE variants of algorithms seems like a promising direction for understanding current methods and developing improved approaches for both cases.

\paragraph{Other similar types of learning not considered here}

There are a number of papers that focus on `fully' decentralized MARL, such as \cite{pmlr-v80-zhang18n}. These methods assume communication between agents during training and execution. As a result, the assumptions are different than the ones considered here. I would term these `fully' decentralized as distributed or networked MARL to make this point more clear. 
There are also methods that assume additional coordination. For example, some methods assume agents take turns learning while the other agents remain fixed \citep{ma2ql,Banerjee12}. Such methods require coordinating updates but can converge to local (i.e., Nash) equilibria.  More details about this alternating learning case are discussed in Section \ref{sec:ctde:other}.

\chapter{Centralized training for decentralized execution (CTDE)}
\label{sec:ctde}

\section{CTDE overview}

The idea of centralized training for decentralized execution is quite vague. The general idea of CTDE in the Dec-POMDP case originated (with the Dec-POMDP itself) in the planning literature \citep{Bernstein02MOR}, where \emph{planning} would be centralized but execution was decentralized. This idea was also considered much earlier in \emph{team} decision making since it is natural to think of deriving a solution for the team as a whole and then assigning corresponding parts to the team members \citep{Marschak55,Radner62,Ho80}.
This concept was carried over to the reinforcement learning case and has come to mean that there is an `offline'\footnote{Offline does not refer to having a fixed training set \citep{OfflineRL} but rather in a different setting than what is used for execution (which would be considered online).} training phase where some amount of centralization is allowed (such as in \citep{Kraemer16}). Typically, this centralized training phase is assumed to take place using a simulator where centralized information is available while execution takes place in the actual decentralized environment. 
After this centralized training phase is complete, agents must then act in a decentralized manner based on their own histories (as given by the Dec-POMDP definition above). 
Any information that would not be available during decentralized execution can be used in the centralized training phase ranging from neural network parameters, policies of the other agents, a joint value estimate, etc. 
Nevertheless, CTDE has never been formally defined (to the best of my knowledge). I will take a wide view of CTDE and consider it to include any centralized information or coordination during learning---anything that is not available during decentralized execution.  Admittedly, there is a bit of a gray area between DTE and CTDE where methods such as deep hysteretic Q-learning that coordinate on a set of random seeds to use during execution could be considered DTE or CTDE.  

In the CTDE case, at each step of the problem we assume that:
\begin{itemize}
	\item each agent observes (or maintains) its current history, $\hi$, takes an action at that history, $\ai$, and sees the resulting observation, $\oi$,
	\item the (centralized) algorithm observes the \emph{joint} information from all agents ($\jh$,  $\ja$, and $\jo$) as well as the joint reward $r$ (and possibly other information such as the underlying state $s$). 
\end{itemize}

CTDE methods for Dec-POMDPs vary widely. 
Some methods add centralized information to DTE methods. These approaches should be scalable but use limited centralized information. 
Other methods are at the other extreme, where they learn a centralized policy and then attempt to decentralize it so it can be executed without the centralized information. 
Most CTDE methods fall somewhere in between these two ideas. The main class of value-based CTDE methods learns a set of decentralized Q-functions (one for each agent) by factoring a joint value function, as in value function factorization methods.  
The main class of policy gradient CTDE methods learns a centralized critic that approximates the joint value function for all agents and that centralized critic is used to update a set of decentralized actors (one per agent). 
Unlike the DTE and CTE cases, these ideas were developed in the context of deep reinforcement learning rather than first being developed for the tabular case and then extended to the deep case. As a result, I will discuss the methods and implementations from the deep MARL perspective.

\section{Value function factorization methods}
\label{sec:ctde:vff}

As mentioned in Section \ref{sec:cte:scale}, there is a long history of learning factored value functions in multi-agent reinforcement learning, but many previous methods were focused on more efficiently learning a joint value function (e.g., \citep{Kok06JMLR}). Other methods have also been developed that share Q-functions during training to improve coordination \citep{Schneider99}. 
Modern value function factorization methods learn Q-functions for each agent by combining them into a joint Q-function and calculating a loss based on the joint Q-function. In this way, the joint Q-function is factored into local, decentralized Q-values that can be used during decentralized execution. 
This idea is appealing because a joint Q-function can be learned to approximate the one in Equation \ref{eq:BellmanQ} but agents can choose actions based on their decentralized Q-values. Additional details and the most popular methods are below. 

\subsection{VDN}

\begin{figure}
\begin{centering}

\includegraphics[scale=0.7]{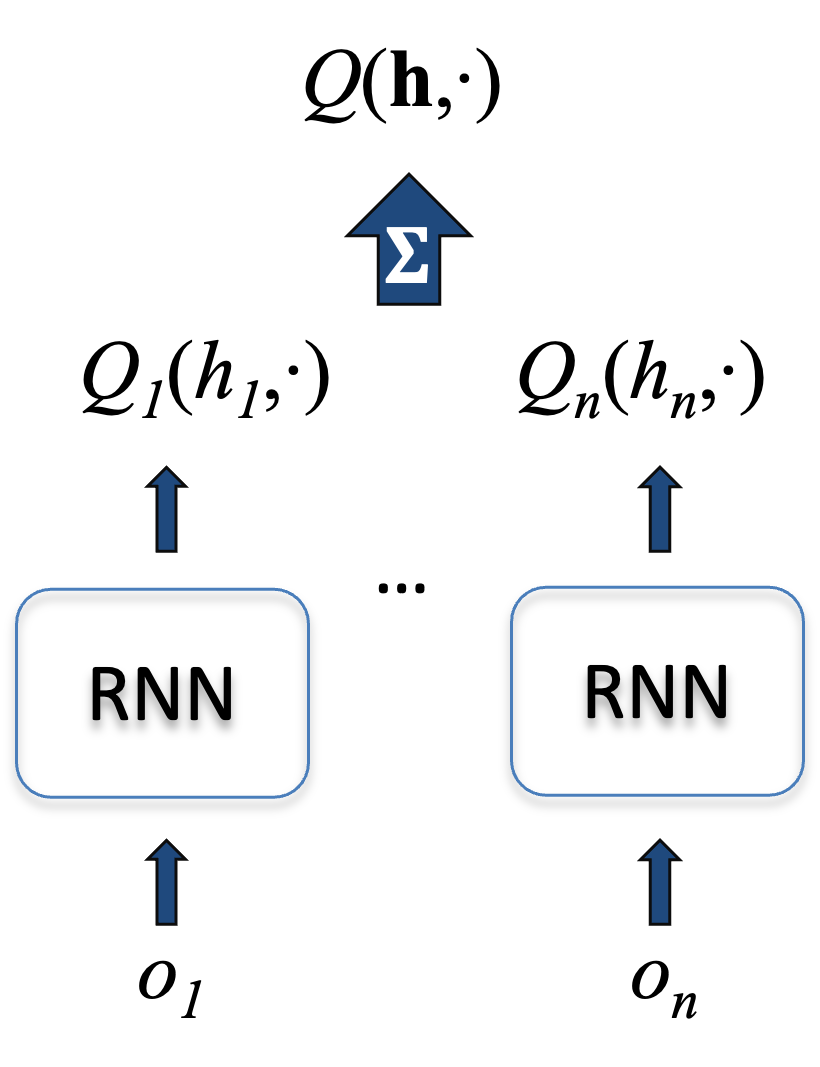}

\end{centering}
\caption{VDN architecture diagram}

\label{fig:VDN}
\end{figure}

Value decomposition networks (VDN) \citep{VDN} began the popular trend in value factorization methods for solving deep multi-agent reinforcement learning. 
The main contribution in VDN is a decomposition of the joint Q-function into additive Q-functions per agent.\footnote{The paper also uses dueling \citep{DuelingDQN} and investigates forms of weight sharing and communication but I view these as orthogonal to the main value decomposition contribution. } This idea is relatively simple but very powerful. It allows training to take place in a centralized setting but agents can still execute in a decentralized manner because they learn individual Q-functions per agent, which can then be maxed over to select an action. 

In particular, it assumes the following approximate factorization of the Q-function:
\begin{equation}
\jQ(\jh, \ja) \approx \sum_{i \in \agentS} \Qi(\hi,\ai) 
\end{equation}
That is, the joint Q-function is approximated by a sum over each agent's individual Q-function. This is a form of factorization---the joint Q-function is factored as a sum of the local $\Qi$ functions \citep{Guestrin01}. 
It is worth noting that each $\Qi$ is technically a utility and not a value function since it is not estimating a particular expected return but can be any arbitrary value in the sum. 

The architecture diagram is given in Figure \ref{fig:VDN}. Each agent learns an individual set of Q-values that depend only on local information. These Q-networks take the current observation as input to a recurrent network, which updates the history representation and then outputs Q-values for that agent. During training, these individual Q-values are summed to generate the joint Q-value which can then be used to calculate the loss like DRQN:
\begin{equation}
    \mathcal{L}(\theta)=\mathbb{E}_{<\jh,\ja,r,\jo>\sim\mathcal{D}}\Big[\big(y - \sum_i^\nrA \Qi^\theta(\hi, \ai)\big)^2 \Big] \text{, where\,\, } y=r + \gamma \sum_i^n \max_{\ai'}\Qi^{\theta^-}(\hi',\ai')
    \label{eq:vdn}
\end{equation}
where $\hi'$ is $\hi\ai\oi$ for $\hi$ taken from $\jh$, $\ai$ from $\ja$, and $\oi$ from $\jo$ which are sampled from dataset $\mathcal{D}$. 

Since the loss uses the sum of the agent's Q-functions, the result is a set of Q-values whose sum approximates the joint Q-function. 
This approximation will be lossless in extreme cases (e.g., full independence) but it may be close or still permit the agents to select the best actions even if the approximation is poor. 
The approach is also scalable since the max used to create the target value $y$ is not over all agent actions as in the centralized case ($\jaS$), but done separately for each agent in Equation \ref{eq:vdn}. Because action selection is over these local Q-values, the policy (and the resulting Q-function) is decentralized. 

A simple version of VDN is given in Algorithm \ref{alg:VDN}. The approach is very similar to standard DRQN \citep{DRQN} as well as multi-agent DTE extensions \citep{Omidshafiei17,Tampuu} which are discussed in Sections \ref{sec:back:value} and \ref{sec:dte:value}. The key difference is the calculation of the loss on Line \ref{alg:vdn:loss}, which approximates the joint Q-value by using the sum of the individual Q-values. 

In more detail, the $\Qi$ are estimated using RNNs for each agent parameterized by $\theta_i$, along with using a target network, $\theta_i^-$, and a replay buffer, $\mathcal{D}$.  First, episodes are generated by interacting with the environment (e.g., $\epsilon$-greedy exploration), which are added to the replay buffer and indexed by episode number and time step ($\mathcal{D}^e(\ts)$). 
This buffer typically has a fixed size and new data replaces old data when the buffer is full. Episodes can then be sampled from the replay buffer (either single episodes or as a minibatch of episodes). In order to have the correct internal state, RNNs are trained sequentially.  Training is done from the beginning of the episode until the end, calculating the internal state and the loss at each step  and updating each $\theta_i$ using gradient descent. The target networks $\theta_i^-$ are updated periodically (every $C$ episodes in the code) by copying the parameters from $\theta$.

\begin{algorithm}
\begin{algorithmic}[1]
\STATE set $\alpha$, $\epsilon$, and $C$ (learning rate, exploration, and target update frequency)
\STATE Initialize network parameters $\thetai$ for each $\Qi$ (denoted $\Qi^\theta$)
\STATE {\bf for all} $i$, $\theta_i^- \gets \theta_i$ 
\STATE $\mathcal{D} \gets \emptyset$
\STATE $e \gets 1$ \hfill \COMMENT{episode index}
\FORALL{episodes}
\STATE  {\bf for all} $i$, $\hi \gets \emptyset $ \hfill \COMMENT{initial history is empty}
\FOR{$\ts=0$ to $\hor-1$}
\STATE {\bf for all} $i$, take $\ai$ at $\hi$ from $\Qi^\theta(\hi,\cdot)$ with exploration (e.g., $\epsilon$-greedy)
\STATE  See joint reward $\rT \ts$, and observations $\joT \ts$\hfill 
\STATE append $\ja,\jo,r$ to $\mathcal{D}^e $
\STATE {\bf for all} $i$,  $\hi \gets \hi\ai\oi$ \hfill \COMMENT{update RNN state of the network}
\ENDFOR
\STATE sample an episode from $\mathcal{D}$
\STATE {\bf for all} $i$, $\hi \gets \emptyset $ 
\FOR{$\ts=0$ to $\hor-1$}
\STATE $\ja, \jo, r \gets \mathcal{D}^e(\ts)$
\STATE {\bf for all} $i$, $\hi' \gets \hi\ai\oi$ 
\STATE $y=r + \gamma\sum_i\max_{\ai'}\Qi^{\theta^-}(\hi',\ai')$
\STATE  {\bf for all} $i$, do gradient descent on $\thetai$ with learning rate $\alpha$ and loss  $ \big(y - \sum_j Q_j^{\theta}(h_j, a_j)\big)^2  $ \label{alg:vdn:loss}
\STATE {\bf for all} $i$, $\hi \gets \hi'$
\ENDFOR
\IF{$e \mod C = 0$} 
   \STATE {\bf for all} $i$, $\theta_i^- \gets \theta_i$ 
 \ENDIF
 \STATE $e \gets e+1$
\ENDFOR
\RETURN all $\Qi$ 
\end{algorithmic} 
\caption{A version of value decomposition networks (VDN) (finite-horizon)}
\label{alg:VDN}
\end{algorithm}

As with the other cases, while the algorithm is presented for the finite-horizon case (for simplicity), it can easily be extended to the episodic and infinite-horizon cases by including terminal states or removing the loop over episodes. 

After training, each agent keeps its own recurrent network, which can output the Q-values for that agent. The agent can then select an action by argmaxing over those Q-values: $\poli(\hi)=\argmax_{\ai}\Qi(\hi,\ai)$. 

VDN is often used as a baseline since using a simple sum to combine Q-functions often results in poor performance compared to more sophisticated methods. 

\subsection{QMIX}

QMIX \citep{QMIX,QMIXJMLR} extends the value factorization idea of VDN to allow more general decompositions. In particular, instead of assuming the joint Q-function, $\jQ(\jh, \ja)$, factors into a sum of local Q-values, $\Qi(\hi,\ai)$, QMIX assumes the joint Q-function is a monotonic function of the individual Q-functions.  This monotonic assumption means increases in local Q-functions lead to increases in the joint Q-function. The sum used in VDN is also a monotonic function but QMIX allows more general (possibly nonlinear) monotonic functions to be learned. 

Specifically, QMIX assumes the following approximate factorization of the Q-function:
\begin{equation}
\jQ(\jh, \ja) \approx f_{mono}(Q_1(\aoHistA 1,\aA 1),\ldots, Q_n(\aoHistA n,\aA n))
\end{equation}

This monotonic assumption means the argmax over the local Q-functions is also an argmax over the joint Q-function. 
This property allows agents to choose actions from their local Q-functions instead of having to choose them from a joint Q-function---ensuring the choice would be the same in both cases.   
Like in VDN, it also makes the training more efficient since calculating the argmax in the update Equation \ref{eq:qmix} is linear in the number of agents rather than exponential.

\cite{QTRAN} formalized this property as the Individual-Global-Max (IGM) principle. IGM states that the argmax over the joint Q-function is the same as argmaxing over each agent's individual Q-function. The definition below is for a particular history but it should ideally hold for all histories (or at least the ones visited by the policy). 
\begin{definition}[Individual-Global-Max (IGM) \citep{QTRAN}] 
For a joint action-value function $\jQ(\jh, \ja)$, where $\jh=\langle \aoHistA 1,\ldots, \aoHistA n \rangle$ is a joint action-observation history, if there exist individual 
functions [$\Qi$], such that:
    \begin{equation}
  \argmax_{\ja} \jQ(\jh, \ja) = \left( \begin{array}{l}
      \argmax_{\aA 1}Q_1(\aoHistA 1,\aA 1)\\
          \hfill \vdots  \hfill\\
      \argmax_{\aA n}Q_n(\aoHistA n,\aA n)
    \end{array}
  \right),
\end{equation}
then,  [$\Qi$] satisfy IGM for $\jQ$ at $\jh$.
  \label{def:IGM}
\end{definition}

\begin{figure}
\begin{centering}

\includegraphics[scale=0.7]{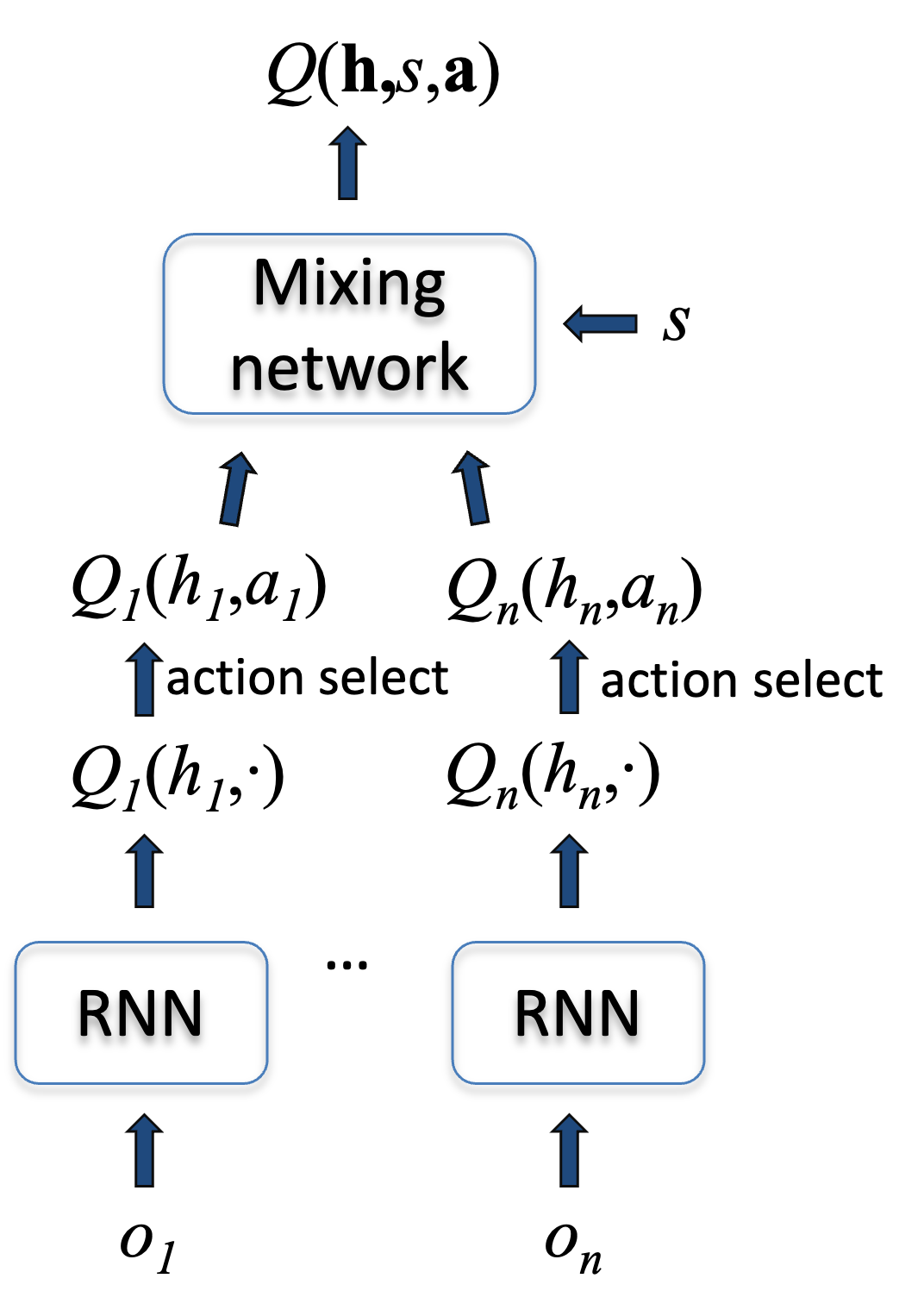}

\end{centering}
\caption{QMIX diagram}

\label{fig:QMIX}
\end{figure}

A simplified version of the QMIX architecture is shown in Figure \ref{fig:QMIX}. Like VDN, each agent has an RNN that takes in the current observation and can output Q-values for the updated history over all actions.\footnote{QMIX typically also inputs the previous action but this is omitted here for clarity and consistency.} A particular action is then chosen (e.g., using $\epsilon$-greedy action selection) using each $\Qi$ and then each $\Qi(\hi,\ai)$ is fed into the mixing network. The mixing network is made to be monotonic by restricting the weights (but not the bias terms) to be non-negative. 
The mixing network can also receive the state to potentially improve performance. In practice, the state is given as input to hypernetworks \citep{hypernetworks} which generate the weights for the layers of the mixing network. 
Because the output depends on the state, it is $\jQ(\jh,s,\ja)$, and is often called $\jQ_{tot}$.

The network is trained end-to-end, like VDN, by calculating a loss between $\jQ(\jh,s,\ja)$ and the joint return. 
The corresponding loss is:
\begin{multline}
    \mathcal{L}(\theta)=\mathbb{E}_{<\jh,s,\ja,r,\jo,s'>\sim\mathcal{D}}\Big[\big(y - \jQ^\theta(\jh,s,\ja)\big)^2 \Big],  \text{where\, } y=r + \gamma  \jQ^{\theta^-}(\jh',s',\tilde \ja'),   \\\text{ and\, }  \tilde \ja'= \langle \argmax_{\aA 1'} Q_1(\aoHistA 1',\aA 1'),\ldots, \argmax_{\aA n'} Q_n(\aoHistA n',\aA n') \rangle
    \label{eq:qmix}
\end{multline}
where the value on the next time step, $\jQ^{\theta^-}(\jh',s',\tilde \ja')$, is gotten by inputting the argmax over each agent's local Q-value using $\jh'=\langle \aoHistA 1', \ldots, \aoHistA \nrA' \rangle$ as well as the corresponding state $s'$ that is sampled from the buffer at  $\jh'$. 

The mixing network is only needed during training. Agents can retain their RNNs (which are just DRQNs) for selecting actions during decentralized execution just like VDN. 

Again, such a factorization may not be possible in all problems so QMIX may not be able to approximate the true joint Q-function in all problems accurately. In particular, as discussed in the paper, this factorization should fail when agents' action choices depend on (at least some) other agents' actions at the same time step. 

\subsection{Weighted QMIX}
Weighted QMIX extends QMIX in an attempt to improve its expressiveness so it can represent more complex Q-functions \citep{WeightedQMIX}.
The main idea is to weigh the actions at different histories differently so the Q-values for some actions can be accurately represented but other action values can be less accurate. 
 For instance, if you knew the optimal policy, you could set the weight highest for the optimal action in each history.  You don't need to accurately represent the other action values but you do need to make sure they are lower than the optimal action (so the policy will choose the optimal action!). 

Specifically, weighted QMIX uses weights $w(\jh,\ja) \in  (0,1]$ to adjust the importance of joint history-action pairs in the loss (as formalized below).\footnote{The paper discusses much of the method and theory in terms of the fully observable case but I consider the partially observable case here. In the partially observable case, the weights can take the history or the history and the state: $w(h,s,a)$. } 
It turns out that there always exists some weighing that will allow the optimal policy to be recovered by independently argmaxing over each agent's $\Qi$. 
An idealized form of the algorithm could also converge in the fully observable case but this proof doesn't extend to the partially observable case or when approximations are used (i.e., the deep RL case).

The method has two main components, a QMIX network where the output is weighted, and a network that approximates the optimal Q-value as $\hat \jQ^*$.
The first network takes the output of QMIX (as seen in Figure \ref{fig:QMIX}) and weighs it as seen in Equation \ref{eq:wqmix}. For easier differentiation with the other Q network, I call the output of the first (QMIX-style) network $\jQ_{tot}$. This network is exactly the same as the one in QMIX but the output is weighted in the loss to prioritize different actions at different histories. 
 The $\hat \jQ^*$ network is similar to the one in QMIX but it does not limit the weights to be non-negative and doesn't use a hypernetwork (just directly inputting the state into the mixing network). These networks don't share parameters and are trained using the losses given in Equation  \ref{eq:wqmix}. The same target ($y$) is used in both cases, which now uses the unconstrained output  $\hat \jQ^*$. The argmax is done the same way as in QMIX, where the actions are chosen using each agent's $\Qi$. As a result, the argmax is tractable but a potentially more accurate value of those actions can be provided from  $\hat \jQ^*$ rather than $\jQ_{tot}$. 
 
 In particular, the losses used are:
\begin{equation}
\begin{split}
 &\mathcal{L}(\theta_{cent})=\mathbb{E}_{<\jh,s,\ja,r,\jo,s'>\sim\mathcal{D}}\Big[\big(y - \hat \jQ^*(\jh,s,\ja)\big)^2 \Big], \hfill \\
   &  \mathcal{L}(\theta_{tot})=\mathbb{E}_{<\jh,s,\ja,r,\jo,s'>\sim\mathcal{D}}\Big[w(\jh,\ja)\big(y - \jQ_{tot}(\jh,s,\ja)\big)^2 \Big],  \\
    & \text{where\, } y=r + \gamma  \hat \jQ^*(\jh',s',\tilde \ja'),  \text{ and\, }  \tilde \ja'= \langle \argmax_{\aA 1'} Q_1(\aoHistA 1',\aA 1'),\ldots, \argmax_{\aA n'} Q_n(\aoHistA n',\aA n') \rangle\\
    \label{eq:wqmix}
\end{split}
\end{equation}

Two different weighting functions are considered. Centrally-Weighted QMIX (CW) uses:
  $$ w(\jh,\ja)= \bigg\{
 \begin{array}{ll}
      1 \quad \text{if \,} y > \hat \jQ^*(\jh,s,\tilde \ja^*) \text{\, or \,} \ja=\tilde \ja^* \\
      \alpha \quad  \text{\, otherwise}\\
    \end{array}$$
which is an approximation of the optimal action $\tilde \ja^*$ using the individual argmax as in Equation \ref{eq:wqmix} and an approximation of the value function, $\hat \jQ^*$, using the unconstrained network. The weight is set to $1$ when the action is already known to be optimal ($\ja=\tilde \ja^*$) or when the action has a higher value than the current best action.   
Optimistically-Weighted QMIX (OW) uses:
        $$ w(\jh,\ja)= \bigg\{
 \begin{array}{ll}
      1 \quad \text{if \,} y>\jQ_{tot}(\jh,s,\ja)  \\
      \alpha \quad  \text{\, otherwise}\\
    \end{array}$$
which just sets the weight to $1$ when the current estimate using $\jQ_{tot}$ is less than the target value, which uses $\hat \jQ^*$. This inequality suggests error in $\jQ_{tot}$ and $\ja$ could (optimistically) be optimal at $\jh$. 
$\alpha$ is a hyperparameter that can be set during training. 

 After training, the $\Qi$ from the constrained network (outputting $\jQ_{tot}$) can be used for decentralized action selection (just like in QMIX). Therefore, $\hat \jQ^*$ is only used to help guide the learning so better (hopefully, optimal) actions will have higher values in each agent's $\Qi$. 
While weighted QMIX can outperform QMIX in some cases, it isn't as widely used as more recent methods (e.g., QPLEX \citep{QPLEX} and MAPPO \citep{MAPPO}). 

\subsection{QTRAN}

QTRAN \citep{QTRAN} provides a different way to factor the joint Q-function into individual Q-functions. The idea generalizes VDN but the intuition is that the sum does not have to equal the true Q-function but rather some \emph{transformed} Q-function $\jQ'$. A separate V term\footnote{While the paper calls this a state value, the value takes histories as input, not states.} can then be used to account for the error between the transformed $\jQ'$ and  the true $\jQ$. The approach cannot represent all IGM-able functions but it is more general than VDN.\footnote{While some have taken the text of the paper to show that QTRAN can represent all IGM-able functions, the proofs only show that the method satisfies IGM (not that any IGM-able function can be represented by the method).} QPLEX (which I talk about below) is more general and the relationship to QMIX and weighted QMIX is unclear. 

Specifically, QTRAN uses Equation \ref{eq:qtran} as the basis for their architecture and losses. 

    \begin{equation}
\sum_{i} \Qi(\hi,\ai) - \jQ(\jh,\ja)+\jV(\jh)= \bigg\{
 \begin{array}{ll}
      0 \quad\quad \ja=\tilde \ja\\
      \ge 0 \quad \ja\ne\tilde \ja\\
    \end{array}
    \label{eq:qtran}
\end{equation}
where $$\jV(\jh)=\max_{\ja}\jQ(\jh,\ja) -\sum_{i} \Qi(\hi,\tilde \ai),$$  
 $$\tilde \ai=  \argmax_{\ai} \Qi(\hi,\ai), \quad \text{and} \quad \tilde \ja= \langle \tilde {\aA 1}, \ldots,  \tilde {\aA n} \rangle$$
Here,  $\jV(\jh)$ has a particular form where it is the difference between the (centralized) max over the joint Q-function, $\jQ$, and the sum of the maxes over the individual Q-functions, $\Qi$. Equation \ref{eq:qtran} is constrained to be $0$ when the actions are the argmax over the individual Q-functions, $\tilde \ja$, and greater than or equal to 0 for other actions. If Equation \ref{eq:qtran} holds, the $\Qi$ satisfy IGM for the $\jQ$ but it isn't clear how general it is. That is, unlike QPLEX, the proof does not show that all functions that satisfy IGM can be represented in this form. 

Instead of enforcing a network structure to get IGM like VDN and QMIX, QTRAN does so by using losses based on Equation \ref{eq:qtran}. The architecture outputs $\jQ'(\jh,\ja)=\sum_i  \Qi(\hi, \ai)$, like VDN, as well as an unconstrained $\jQ(\jh,\ja)$ and $\jV(\jh)$. 
As shown below, there are three losses. There is the standard TD loss,  $\mathcal{L}_{td}$, for learning the true joint Q-function, $\jQ$. There is also a loss for learning the transformed Q-function, $\jQ'$, and value offset,  $\jV$, for the given $\jQ$ when the actions are currently the maximizing ones, $\tilde \ja$, in $\mathcal{L}_{opt}(\theta)$. When the actions are not  maximizing ones  $\mathcal{L}_{nopt}(\theta)$ is used. The notation $\bar \jQ$ is used to note that  $\jQ$  isn't trained from $\mathcal{L}_{opt}(\theta)$ and  $\mathcal{L}_{nopt}(\theta)$. 
\begin{equation}
\begin{split}
 &   \mathcal{L}_{td}(\theta)=\mathbb{E}_{<\jh,\ja,r,\jo>\sim\mathcal{D}}\Big[\big(y -  \jQ(\jh,\ja)\big)^2 \Big], \\
  &   \mathcal{L}_{opt}(\theta)=\mathbb{E}_{<\jh,\ja,r,\jo>\sim\mathcal{D}}\Big[\big(\jQ'(\jh,\tilde \ja) - \bar \jQ(\jh,\tilde \ja)+\jV(\jh)\big)^2 \Big], \\
   &   \mathcal{L}_{nopt}(\theta)=\mathbb{E}_{<\jh,\ja,r,\jo>\sim\mathcal{D}}\Big[\Big(\min\big[\jQ'(\jh, \ja) - \bar \jQ(\jh, \ja)+\jV(\jh),0\big]\Big)^2 \Big], \\
  &  \text{where\, } y=r + \gamma  \jQ^{\theta^-}(\jh',\tilde \ja'),  \text{ and\, }  \tilde \ja'= \langle \argmax_{\aA 1'} Q_1(\aoHistA 1',\aA 1'),\ldots, \argmax_{\aA n'} Q_n(\aoHistA n',\aA n') \rangle\\
    \label{eq:qtranloss}
    \end{split}
\end{equation}

The losses above are used for the standard form of QTRAN called QTRAN-base. An alternate form, called QTRAN-alt, replaces  $\mathcal{L}_{nopt}(\theta)$ with the average counterfactual loss below, forcing the value to be 0 for some action rather than relying on the inequality constraint above.  
$$ \mathcal{L}_{nopt\text{-}min}(\theta)=\mathbb{E}_{<\jh,\ja,r,\jo>\sim\mathcal{D}}\Big[\frac{1}{n}\sum_i^n\Big(\min_{\ai}\big[\jQ'(\jh,\ai, \ja_{-i}) - \bar \jQ(\jh,\ai, \ja_{-i})+\jV(\jh)\big]\Big)^2 \Big]
$$
Using this condition also satisfies IGM. 

The versions of QTRAN (QTRAN-base and QTRAN-alt) can outperform VDN and QMIX on some domains but they are often outperformed by more recent methods.

\subsection{QPLEX}

QPLEX \citep{QPLEX} further extends the value factorization idea to (provably) include more general decentralized policies. In particular, QPLEX can potentially learn any set of Q-functions that satisfy the IGM principle. 

First, QPLEX extends the IGM principle to an advantage-based case. They
define joint values and advantages from the joint Q-function:
$$\jV(\jh)=\max_{\ja} \jQ(\jh,\ja), \text{ and\, } 
\jA(\jh,\ja)=\jQ(\jh,\ja)-\jV(\jh),$$ as well as 
 individual values and advantages from each agent's individual Q-function:
$$\Vi(\hi)=\max_{\ai} \Qi(\hi,\ai), \text{ and\, } 
\Ai(\hi,\ai)=\Qi(\hi,\ai)-\Vi(\hi).$$ 
Note that this notion of advantage is a bit different than the typical notion (such as \citep{DuelingDQN}) since the values and advantages are generated from the Q-values, rather than the other way around. As a result, the advantages will have the property that they will be 0 for optimal actions and negative otherwise. That is, $\Ai(\hi,\ai^*)= 0$ (since $\Qi(\hi,\ai^*)-\max_{\ai}\Qi(\hi,\ai)=0$) and for some non-optimal action $\ai^{\dagger}$, $\Ai(\hi,\ai^{\dagger}) < 0$ (since $\Qi(\hi,\ai^{\dagger})<\max_{\ai}\Qi(\hi,\ai)$). The same holds true for joint advantages. 

Advantage-based IGM extends IGM to the advantage case and states that the argmax over the joint advantage function is the same as argmaxing over each agent's individual advantage function:
\begin{definition}[Advantage-based (IGM) \citep{QPLEX}] 
For a joint action-value function, if there exist individual 
functions [$\Qi$], such that:
    \begin{equation}
  \argmax_{\ja} \jA(\jh, \ja) = \left( \begin{array}{l}
      \argmax_{\aA 1}A_1(\aoHistA 1,\aA 1)\\
           \hfill \vdots  \hfill\\
      \argmax_{\aA n}A_n(\aoHistA n,\aA n)
    \end{array}
  \right),
\end{equation}
where $\jA(\jh, \ja)$ and $\Ai$ are defined above, then [$\Qi$] satisfy advantage-based IGM for $\jQ$ at $\jh$.
\end{definition}
That is, advantage-based IGM is equivalent to the Q-based IGM in Definition \ref{def:IGM}. 

\begin{figure}
\begin{centering}

\includegraphics[scale=0.7]{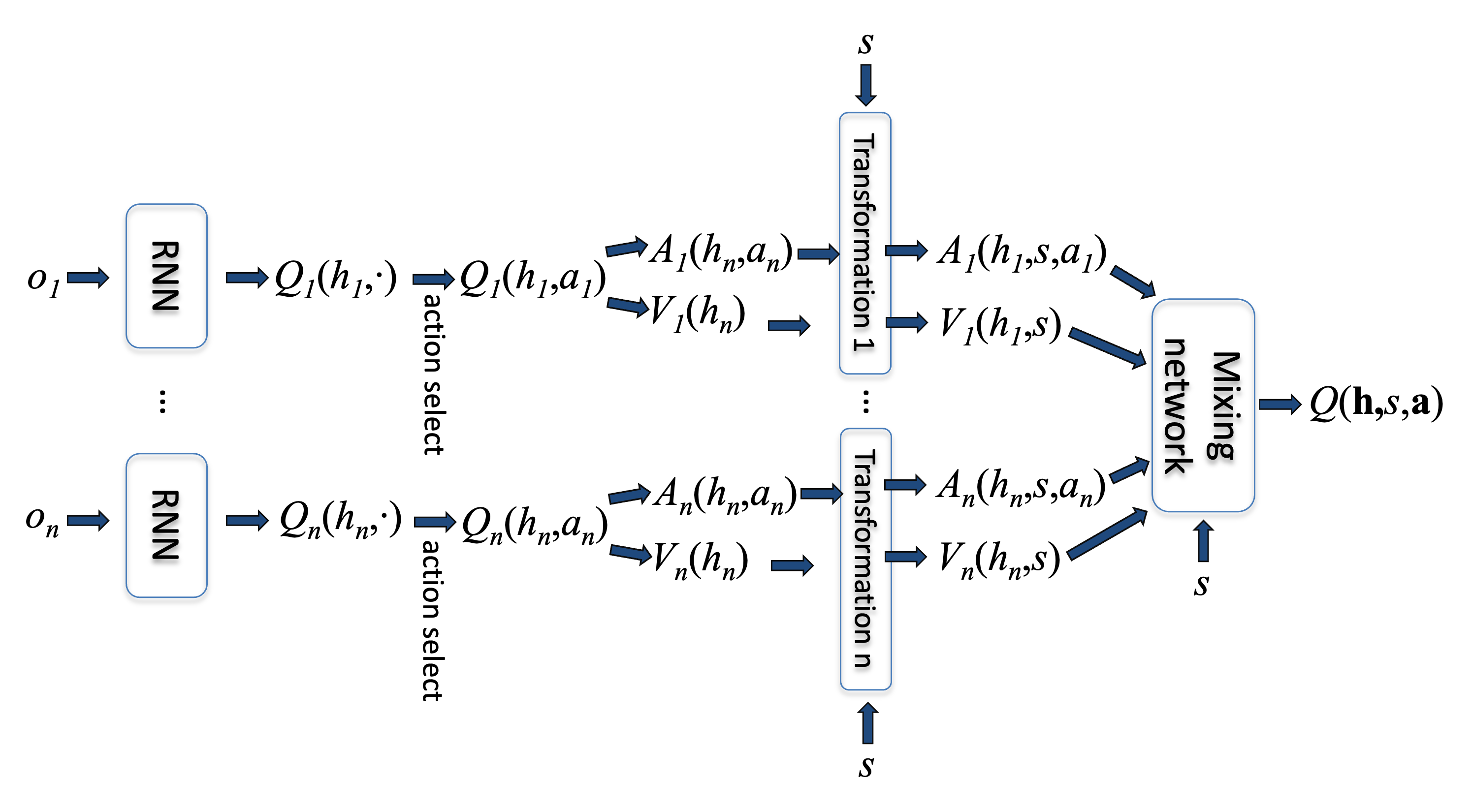}

\end{centering}
\caption{QPLEX diagram}

\label{fig:QPLEX}
\end{figure}

QPLEX uses advantage-based IGM to extend QMIX and QTRAN to represent the full IGM function class. 
The architecture is shown in Figure \ref{fig:QPLEX}. 
Like QMIX (and VDN and QTRAN), QPLEX first takes as input each agent's observation $\oi$ (and previous action $\aiT {t-1}$ but that is not included here for simplicity) and outputs the individual Q-values, $\Qi(\hi,\ai)$, for all $\ai$ using a DRQN-style network. After an action is selected (e.g., using $\epsilon$-greedy exploration), the value and advantage are extracted from the Q-values for each agent as $\Vi(\hi)=\max_{\ai} \Qi(\hi,\ai)$ and $\Ai(\hi,\ai)=\Qi(\hi,\ai) - \Vi(\hi)$. 
Next, the transformation networks generate $\Vi(\hi,s)$ and $\Ai(\hi,s,\ai)$ for each agent with the given $\Vi(\hi)$ and $\Ai(\hi,\ai)$ along with  $s$ as $\Vi(\hi,s)=w_i(s)\Vi(\hi)+b_i(s)$ and $\Ai(\hi,s,\ai)=w_i(s)\Ai(\hi,\ai)$. Finally, the joint Q-value is output based on the output of each agent's transformation network and the state as $\jQ(\jh,s,\ja) = \sum_i \big[\Vi(\hi,s)+\lambda_i(s,\ja) \Ai(\hi,s,\ai)\big]$.\footnote{Attention is used to train the $\lambda$ weights more efficiently.} 
While there aren't constraints on V due to IGM (because it doesn't include the actions), QPLEX uses a sum to combine the local V's. 
All the weights, $w_i$ and $\lambda_i$ (but not necessarily the biases $b_i$) are positive to ensure the advantage is 0 for the max action and negative otherwise in order to satisfy the advantage IGM principle. 
That is, instead of using monotonicity like VDN and QMIX, QPLEX achieves IGM by constraining the max action to have a value of 0 for the joint and individual advantages and the advantages for other actions are negative. This constraint is exactly what IGM says---the argmax of the joint $\jQ$ (or $\jA$) is the argmax of the individual $\Qi$ (or $\Ai$). 
The architecture is trained using  the RL loss, just like QMIX in Equation \ref{eq:qmix}.

As a result of the advantage-based architecture, QPLEX is more general than previous methods with the capacity to represent all Q-functions that satisfy IGM. While this doesn't mean QPLEX is optimal, it does mean that it is at least capable of representing the optimal Q-value (which is something methods like QMIX and VDN may not be able to do). 
QPLEX also performs well, often outperforming other value factorization methods and it is currently one of the best-performing CTDE methods.

\subsection{The use of state in factorization methods}
\label{sec:vff:state}

Many of the current methods use inconsistent notation about the inclusion of state (except QMIX). As a result, the theory is often developed without considering the state input. 
It turns out that, unlike the policy gradient case discussed in Section \ref{sec:ctde:pg:states}, using the state doesn't introduce bias and is thus theoretically sound \citep{Marchesini25}. The intuition is that the state doesn't replace the history, but  augments it (similar to a history-state critic in Section \ref{sec:ctde:pg:states}). Furthermore, actions are not chosen based on the state but only the local, history-dependent Q-values, $\Qi$. 
Nevertheless, in implementations of algorithms such as QPLEX and weighted QMIX, the weights only take the state as input and not the history. This replacement of history with state in these contexts limits the representational power of the weights in partially observable problems. As a result, using the state as input in QPLEX (instead of the joint history) prevents it from being able to represent the full IGM class of functions. It still may be beneficial to use the state instead of the joint history but exploring what additional information to use and in what way is an interesting open question. 

\subsection{Other value function factorization methods}

Qatten~\citep{Qatten} extends QMIX (and weighted QMIX) to add structure to the mixing network, where the structure is inspired by a linear approximation of the Q-function and one set of weights is learned using attention. A wide range of other approaches have been developed but are not included in this introduction to the area. 

\section{Centralized critic methods}
\label{sec:ctde:pg}

Concurrently, Multi-Agent Deep Deterministic Policy Gradient (MADDPG)~\citep{MADDPG} and COunterfactual Multi-Agent policy gradient (COMA)~\citep{COMA} popularized the use of centralized critics in MARL. 
I will first discuss the general class of algorithms that MADDPG and COMA represent---multi-agent actor-critic methods with centralized critics---and then give the details of MADDPG and COMA. I will then describe the very popular PPO-based extension, MAPPO~\citep{MAPPO}. 
The basic motivation is to leverage the centralized information that is available during training and potentially combat nonstationarity that can happen in decentralized training by using a centralized value estimator along with decentralized policies. I will also discuss the (correct and incorrect) use of state in centralized critics as well as the theoretical and practical tradeoffs in the various types of critics. 

\subsection{Preliminaries}

\begin{figure}
\begin{centering}
\hfill
\subfigure[]{
\includegraphics[scale=0.25]{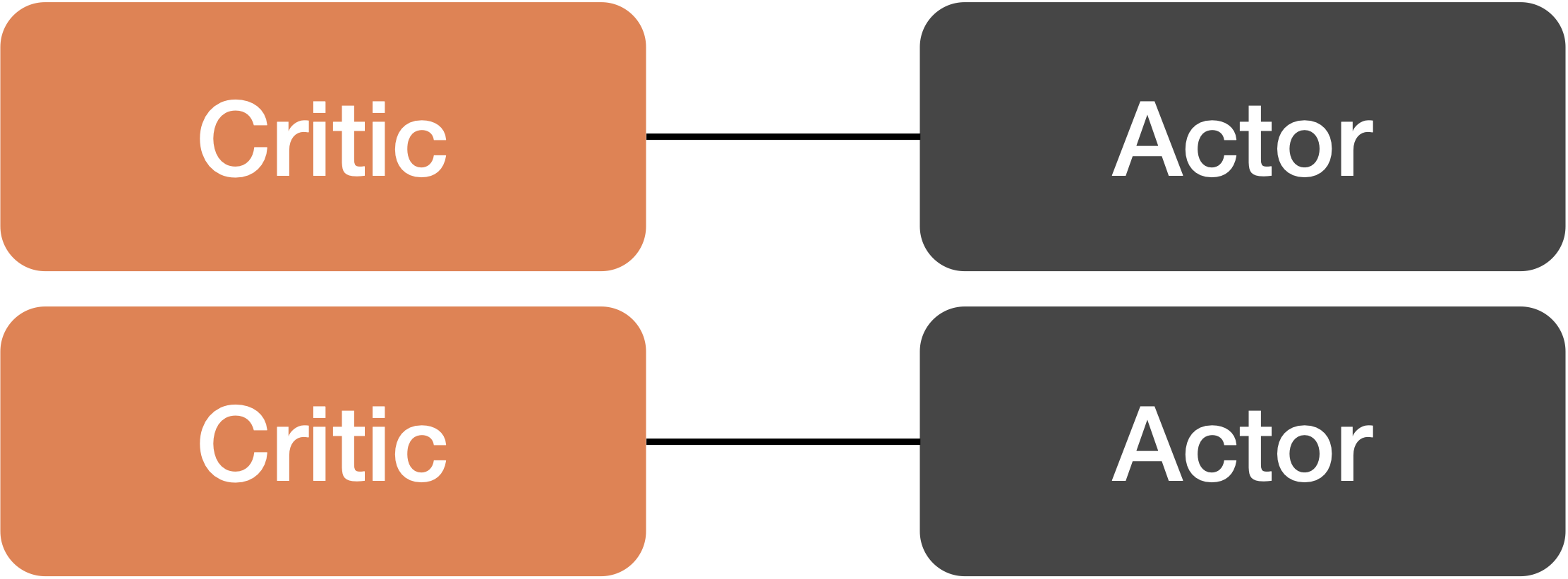} \label{fig:IAC}
}
\hfill
\subfigure[]{
\includegraphics[scale=0.25]{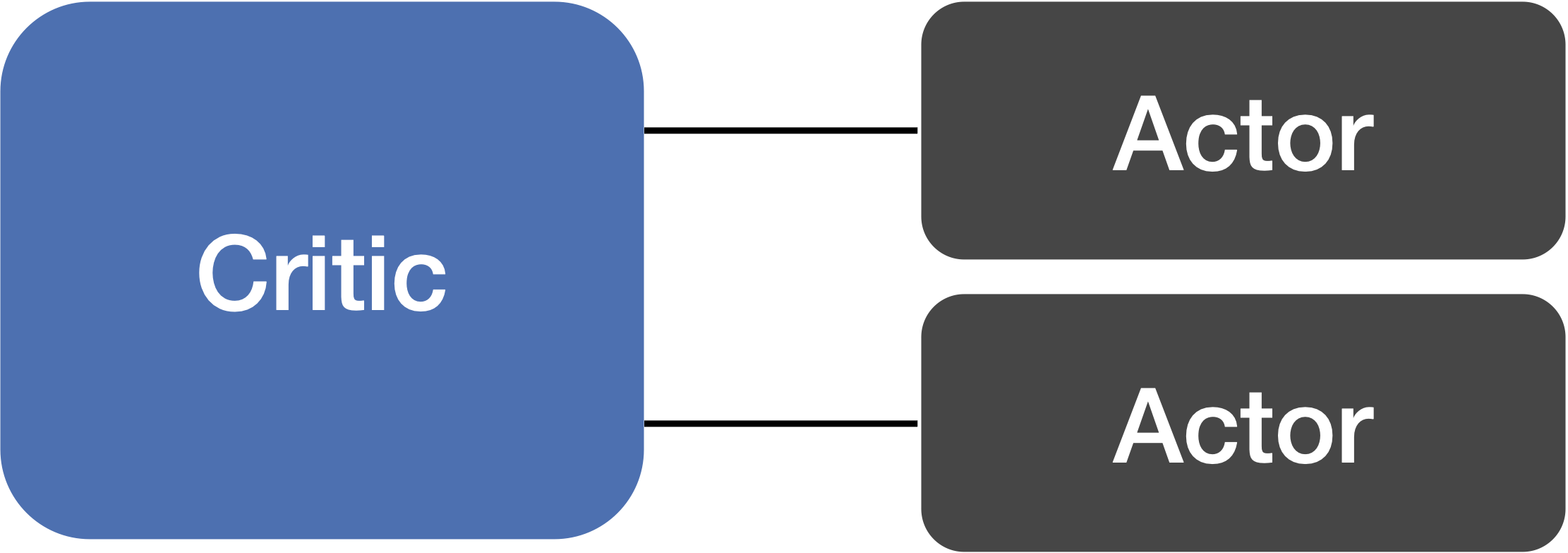} \label{fig:IACC}
}
\hfill
\end{centering}
\caption{Decentralized critics (a) vs.~a centralized critic (b). }
\end{figure}

Single-agent policy gradient and actor-critic methods have also been extended to the CTDE case. In the multi-agent case, there is one actor per agent and either one critic per agent (as shown in Figure \ref{fig:IAC} and discussed in Section \ref{sec:dte:pg}) or a shared, centralized critic (as shown in Figure \ref{fig:IACC} and discussed below). The centralized critic can be used during CTDE training but then discarded during execution. Each agent can act in a decentralized manner by using its actor.  All the policy gradient approaches can generally scale to larger action spaces and have stronger (local) convergence guarantees than the value-based counterparts in Section \ref{sec:ctde:vff}. 

Policy gradient methods use continuous-action or stochastic policies. 
Unless stated otherwise, I will assume a stochastic policy for each agent parameterized by $\ppi$, where $\poli^{\ppi}(\ai| \hi)$ represents the probability agent $i$ will choose action $\ai$ given the history $\hi$ and parameters, $\ppi$, $\Pr(\ai| \hi,\ppi)$. 
Like the value-based methods, the CTDE policy gradient methods also assume algorithms receive agent actions, $\ja$, the resulting observations, $\jo$, and the joint reward, $r$ at each time step.

\subsection{A basic centralized critic approach}

In the simplest form, we can consider a class of centralized critic methods where the joint history critic, which I denote as $\jQmodel(\jh, \ja)$, 
 estimates the value of a policy $\jQpol(\jh, \ja)$ with parameters $\valparam$. $\jQmodel(\jh, \ja)$ is then used to update each decentralized policy $\poli$. 
 
We call this approach independent actor with centralized critic (IACC) and it is described in Algorithm \ref{alg:IACC}. 
Learning is over a set of (on-policy) episodes and the histories are initialized to be empty. Agent $i$'s history  at time step $\ts$ is denoted $\hiT \ts$, while the set of histories for all agents is denoted $\jh_{\ts}$. All agents choose an initial action from their policies $\aiT 0 \sim \poli(\ai| \hiT 0)$ and then the learning takes place over a fixed horizon of $\hor$. At each step, the agents take the corresponding current action $\jaT \ts$ and see observations $\joT \ts$ and the algorithm receives the joint reward $\rT \ts$. The histories for the next step are updated to include the action taken and observation seen, $\hiT {\ts} \aiT \ts \oiT \ts$, and actions are sampled for the next step, $\aiT {\ts+1} \sim \poli(\ai| \hiT {\ts+1})$. Then, the TD error can be computed with the current Q-values, and it is used to update the actors and critic. Each agent's actor is updated using the TD error from the joint Q-function (moving in the direction that improves the joint Q-value). The centralized critic is updated using a standard on-policy gradient update. The action for the next step, $\ts+1$, becomes the action for the current step, $\ts$, and the process continues until the end of the horizon and for each episode. 
The algorithm is written for the finite-horizon case but it  can be adapted to the episodic or infinite-horizon case by including terminal states or removing the loop over episodes. Note that the histories make the value functions based on a particular time step as the history length provides the time step. 

While a sample is used in the algorithm, the full gradient associated with each actor in IACC is represented as:
\begin{equation}
      \nablai J = \mathbb{E}_{<\jh, \ja>\sim\mathcal{D}} \left[ \jQpol(\jh, \ja) \nablai \log \poli(\ai|\hi) \right] \,,
 \label{eq:gradient:qhs}\footnote{Again, for simplicity discounting is removed but the true gradient would include it \citep{nota2020policy}.}
\end{equation}
where the approximation of the joint Q-function, $\jQmodel(\jh, \ja)$, is used to update the policy parameters of agent $i$'s actor, $\poli(\ai| \hi)$. Histories (i.e., observations) and actions are sampled according to the on-policy (discounted) visitation probability. 
The objective, $J$, is to maximize expected (discounted) return starting from the initial state distribution $\bO$ as discussed in Section \ref{sec:decpomdp}. 

The critic is updated using the on-policy loss: 
 $$   \mathcal{L}(\theta)=\mathbb{E}_{<\jh, \ja, r, \jh'>\sim\mathcal{D}}\Big[\big(y - \jQmodel(\jh, \ja)\big)^2 \Big] \text{, where\,\, } y=r + \gamma  \jQmodel(\jh',\ja')$$

\begin{algorithm}
\caption{Independent Actor Centralized Critic (IACC) (finite-horizon)}
\begin{algorithmic}[1]
\label{alg:IACC}
\STATE Initialize individual actor models $\poli(\ai| \hi)$, parameterized by $\ppi$
\STATE Initialize centralized critic model $ \jQmodel(\jh,\ja)$, parameterized by  $\valparam$

\FORALL{episodes}
	\STATE  $\hiT 0 \gets \emptyset$ \hfill  \COMMENT{Empty initial history}
	\STATE Denote $\jh_{0}$ as  $\langle \aoHistAT 1 0,\ldots,\aoHistAT n 0\rangle$ \hfill \COMMENT{Notation for joint variables}
	\STATE {\bf for all} $i$, choose $\aiT 0$ at $\hiT 0$ from $\poli(\ai| \hiT 0)$
	\STATE Store $\ja_0$ as  $\langle \aAT 1 0,\ldots,\aAT n 0\rangle$
	\FOR{$\ts=0$ to $\hor-1$}
		\STATE Take joint action $\ja_{\ts}$, see joint reward $\rT \ts$, and observations $\joT \ts$\hfill 
		\STATE {\bf for all} $i$, $\hiT {\ts+1} \gets \hiT {\ts} \aiT \ts \oiT \ts$\hfill \COMMENT{Append new action and obs to previous history}
		\STATE {\bf for all} $i$, choose $\aiT {\ts+1}$ at $\hiT {\ts+1}$ from $\poli(\ai| \hiT {\ts+1})$
		\STATE Store $\ja_{\ts+1}$ as  $\langle \aAT 1 {\ts+1},\ldots,\aAT n {\ts+1}\rangle$
	 \STATE $\delta_t \gets r_t + \gamma \jQmodel(\jh_{t+1},\ja_{t+1}) - \jQmodel(\jh_t,\ja_t)$ \hfill \COMMENT{Compute centralized TD error}
	 
        \STATE Compute critic gradient estimate: $ \delta_{t} \nablasub \valparam \jQmodel(\jh_t,\ja_t)$ \label{alg:IACC:critic}
        \STATE Update critic parameters $\valparam$ using gradient estimate (e.g., $\valparam \gets \valparam + \beta  \delta_{t} \nablasub \valparam \jQmodel(\jh_t,\ja_t)$ for learning rate $\beta$)
        
            \FOR{each agent $i$} \label{alg:IACC:actor}
       		 \STATE Compute actor gradient estimate: $ \gamma^t \jQmodel(\jh_t,\ja_t) \nablai \log\poli(a_{i,t}| h_{i,t})$ 
                \STATE Update actor parameters $\ppi$ using gradient estimate (e.g., $\ppi \gets \ppi + \alpha \gamma^t \jQmodel(\jh_t,\ja_t) \nablai\log\poli(a_{i,t}| h_{i,t})$ for learning rate $\alpha$)
    \ENDFOR
	\ENDFOR
\ENDFOR
\end{algorithmic}
\end{algorithm}

We can extend this algorithm to the more commonly used advantage actor-critic (A2C) case, as shown in Algorithm \ref{alg:IA2CC}. The advantage is defined as $\jA(\jh,\ja)=\jQ(\jh,\ja)-\jV(\jh)$, representing the difference between the Q-value for a given action and the V-value at that history.\footnote{Note that this advantage definition is the standard one that, unlike QPLEX, allows positive and nonpositive advantages.}
Because $\jV(\jh)$ doesn't depend on the action, it is called a \emph{baseline} and becomes a constant from the perspective of the policy gradient. As a result, using $\jQ(\jh,\ja)-\jV(\jh)$ in place of $\jQ(\jh,\ja)$ doesn't change the convergence properties of the method. 
Nevertheless, using a baseline (e.g., subtracting $\jV(\jh)$) can reduce the variance of the estimate and improve performance in practice. In the algorithm, $\jA(\jh,\ja)$ isn't explicitly stored since the TD error is used as an approximation for the advantage---$\jQmodel(\jh,\ja)$ is approximated with $r_t + \gamma \jVmodel(\jh_{t+1})$, which is true in expectation.  $\delta_t$ can be thought of as a sample of  $\jQmodel(\jh_t,\ja_t)-\jVmodel(\jh_t) = \jAmodel(\jh_t,\ja_t)$.

\begin{algorithm}
\caption{Independent Advantage Actor Centralized Critic (IA2CC) (finite-horizon)}
\begin{algorithmic}[1]
\label{alg:IA2CC}
\STATE Initialize individual actor models $\poli(\ai| \hi)$, parameterized by $\ppi$
\STATE Initialize  centralized critic model $\hat \jV(\jh)$, parameterized by  $\valparam$

\FORALL{episodes}
	\STATE {\bf for all} $i$, $\hiT 0 \gets \emptyset$ \hfill  \COMMENT{Empty initial history}
	\STATE Denote $\jh_0$ as  $\langle \aoHistAT 1 0,\ldots,\aoHistAT n 0\rangle$ \hfill \COMMENT{Notation for joint variables}
	\STATE {\bf for all} $i$, choose $\aiT 0$ at $\hiT 0$ from $\poli(\ai| \hiT 0)$
	\STATE Store $\ja_0$ as  $\langle \aAT 1 0,\ldots,\aAT n 0\rangle$
	\FOR{$\ts=0$ to $\hor-1$}
		\STATE Take joint action $\ja_{\ts}$, see joint reward $\rT \ts$, and observations $\joT \ts$\hfill 
		\STATE {\bf for all} $i$, $\hiT {\ts+1} \gets \hiT {\ts} \aiT \ts \oiT \ts$\hfill \COMMENT{Append new action and obs to previous history}
		\STATE {\bf for all} $i$, choose $\aiT {\ts+1}$ at $\hiT {\ts+1}$ from $\poli(\ai| \hiT {\ts+1})$
		\STATE Store $\ja_{\ts+1}$ as  $\langle \aAT 1 {\ts+1},\ldots,\aAT n {\ts+1}\rangle$
	 \STATE  $\delta_t \gets r_t + \gamma \jVmodel(\jh_{t+1}) - \jVmodel(\jh_t)$ \hfill \COMMENT{Compute centralized advantage estimates (TD error)}
	 
        \STATE Compute critic gradient estimate: $ \delta_{t} \nabla \jVmodel(\jh_t)$\label{alg:IA2CC:critic}
        \STATE Update critic parameters $\valparam$ using gradient estimate (e.g., $\valparam \gets \valparam + \beta \gamma \delta_{t} \nabla \jVmodel(\jh_t))$
        
           \FOR{each agent $i$} \label{alg:IA2CC:actor}
        \STATE Compute actor gradient estimate: $ \gamma^t \delta_t \nabla\log\poli(a_{i,t}| h_{i,t})$ \label{alg:IA2CC:actorgrad}
        \STATE Update actor parameters $\ppi$ using gradient estimate (e.g., $\ppi \gets \ppi + \alpha \gamma^t \delta_{t} \nabla\log\poli(a_{i,t}| h_{i,t})$)
    \ENDFOR
	\ENDFOR
\ENDFOR
\end{algorithmic}
\end{algorithm}

While the critic is called \emph{centralized} because it uses centralized information, it estimates the \emph{joint} value function of the decentralized policies. That is, the centralized critic estimates the value of the current set of decentralized policies (not a centralized policy) \citep{JAIR23}. This is the correct thing to do since the critic's job is to evaluate the policies, which are decentralized in this case. 

With appropriate assumptions (e.g., on exploration, learning rate, and function approximation), IACC (and IA2CC) will converge to a local optimum \citep{Peshkin00UAI,JAIR23,Lyu24}. These results put policy gradient methods on solid theoretical ground and I discuss details of the various methods as well as some theoretical shortcomings. 

This idea of learning a centralized critic to update decentralized actors is very general. It can be (and has been) used with different types of critics, actors, and updates. I present some of the most popular methods below.

\subsection{MADDPG} 
\label{sec:ctde:pg:maddpg}
Multi-Agent DDPG (MADDPG) \citep{MADDPG} is a centralized critic method that considers the more general case of possibly competitive agents as well as continuous actions. To deal with the different reward functions of different agents in the competitive case, separate centralized critics are learned for each agent. 
As I am only concerned with the cooperative case in this text, I assume a single shared critic among agents, do not consider learning policy models of the other agents (since these are assumed to be accessible in a cooperative CTDE setting), and do not consider ensembles of other agent policies to improve robustness. MADDPG is also an off-policy method, unlike the previous algorithms discussed, so it makes use of replay buffer similar to DQN-based approaches. Nevertheless, MADDPG for the cooperative case is very similar to Algorithm \ref{alg:IACC} with the main changes noted below.

To deal with continuous actions, MADDPG extends the continuous action single-agent deep actor-critic method DDPG \citep{DDPG} to the multi-agent case. 
I denote the deterministic continuous-action policies for each agent as $\mu_i$. 
In the cooperative case, MADDPG uses the following policy gradient:
\begin{equation}
      \nablai J = \Exp_{x, \ja\sim \mathcal{D}} \left[ \nablai \mu_i(\oi) \nablasub \ai \jQpol(x, \ja) \mid_{a_i=\mu_i(\oi)} \right] \,,
\end{equation}
where $x$ and $\ja$ are sampled from the buffer with $x$ discussed below and $\ja=\langle \aA 1 ,\ldots,\aA  n \rangle$. 
Since we can no longer sum over actions and weigh their value by their probability as is done in the stochastic policy case, we 
have to consider the change in the Q-value function evaluated at the chosen action $a_i$ for the agent \citep{DPG}. 

Note that MADDPG considers reactive policies that only map from the last observation to an action (as seen in $\mu_i$). As noted in Section \ref{sec:singleobs}, the policy, $\mu_i$, should choose actions based on histories, $\hi$, rather than single observations, $\oi$.
The Q-value is from a centralized critic so the paper states $x$ ``could consist of the observations of all agents [...] however, we could also include additional state information if available." 
More generally, $x$ should be history representations for each agent and I discuss the use of states in more detail in \ref{sec:ctde:pg:states}. 
These issues are easily fixable and result in the following policy gradient (we can incorporate the state information later):
\begin{equation}
      \nablai J = \Exp_{x, \ja\sim \mathcal{D}} \left[ \nablai \mu_i( \hi) \nablasub \ai \jQpol(\jh, \ja) \mid_{a_i=\mu_i( \hi)} \right] \,.
 \label{eq:maddpg:actor}
\end{equation}

Because MADDPG is off-policy, it maintains a replay buffer like DQN-based approaches along with `target' copies of both the actor and critic networks, $\mu_{\ppi^-}$ and $\jQ_{\valparam^-}$. The Q-update (for the history-based case) is then:
\begin{equation}
    \mathcal{L}(\theta)=\Exp_{<\jh, \ja, r, \jh'>\sim\mathcal{D}}\Big[\big(y - \jQ_{\valparam}(\jh, \ja)\big)^2 \Big] \text{, where\,\, } y=r + \gamma \jQ_{\valparam^-}(\jh',\ja')\mid_{\ai=\mu^-(\hi)\ \forall i \in \agentS} 
    \label{eq:maddpg:critic}
\end{equation}
MADDPG executes a stochastic exploration policy (e.g., by adding Gaussian noise to the current deterministic policy) while estimating (and optimizing) the value of the deterministic policy. Specifically, the action, $\ja$, is sampled from the (behavior policy) dataset, while the next action, $\ja'$ is selected by the target policy, $\mu^-$. 

Algorithm \ref{alg:IACC} can be updated to incorporate these changes.  Since the continuous-action policy is deterministic, (Gaussian) noise is added when selecting actions to aid in exploration. The approach is off-policy, so the experiences are first stored in a replay buffer (like DQN and other value-based methods in Section \ref{sec:ctde:vff}). Similarly, episodes are sampled from the replay buffer and target networks used (again, like DRQN-based approaches) for the actor and the critic. 

MADDPG is no longer widely used since newer methods (such as MAPPO) often have better performance but the ideas (such as the centralized critic) have been adopted extensively. 

\subsection{COMA}
\label{sec:ctde:pg:coma}
The main contributions of Counterfactual Multi-Agent Policy Gradients (COMA) were the introduction of the centralized critic along with a counterfactual baseline \citep{COMA}. The idea of comparing an agent's contribution against a counterfactual has been studied as \emph{difference rewards}, where each agent is rewarded by the difference between the team reward and the reward the team would have received without that agent \citep{Wolpert01,Agogino04}. COMA can be seen as computing this difference from a learned critic rather than from a model of the environment.
In the learning case, baselines are common in policy gradient methods since they are high variance. The baseline value is subtracted from the Q-value and it can be anything that is not dependent on the agent's action. 

In the case of COMA, the motivation for the baseline is not only variance reduction but also better credit assignment by subtracting off the perceived contribution to the Q-value from the other agents. Specifically, it marginalizes out the agent's action from the Q-function to get an estimate of what the (counterfactual) Q-function would be while holding the other agent actions fixed. The result is an agent-specific advantage function that subtracts the baseline for agent $i$ from the joint Q-function:\footnote{The COMA paper uses a state-based critic and advantage, $\jQ(s,\ja)$ and  $\Ai(s,\ja)$, but this is incorrect as I'll discuss in Section \ref{sec:ctde:pg:states}. }
$$\Ai(\jh,\ja)=\jQ(\jh,\ja)-\sum_{\ai'} \poli(\ai'| \hi) \jQ(\jh,\ai',\ja_{-i})$$
This baseline does not depend on agent $i$'s action, so it will not bias the gradient.

Rather than have separate baseline networks for each agent, there is a single centralized critic that takes in the other agent actions, $\ja_{-i}$, the joint observation, $\jo$, an agent id, $i$, the proposed action $\ai$, and policy probabilities $\poli(\ai'| \hi)$,
and outputs the advantage for that agent $\Ai(\jh,\ja)$ using the equation above. 
This network can also output the joint Q-values for all agent $i$'s actions, $\jQ(\jh,\cdot,\ja_{-i})$, given the other agent actions, $\ja_{-i}$, the joint observation, $\jo$, and an agent id, $i$. Both of these values are needed for the algorithm.

The COMA algorithm can also be an extension of Algorithm \ref{alg:IACC}.  The critic update can be calculated in the same way (using the Q-values from the network described above) but the agent-specific advantage is incorporated in the actor update. That is, during the actor update (starting on line \ref{alg:IACC:actor}), the agent-advantage $\Ai(\jh,\ja)$ is calculated for the given agent and $\Ai$ is used in the actor gradient estimate instead of $\jQmodel$: $ \gamma^t \Ai(\jh_t,\ja_t) \nablai \log\poli(a_{i,t}| h_{i,t})$. 

While COMA has been very influential, it isn't widely used since newer methods tend to outperform it. 

\subsection{MAPPO}
\label{sec:ctde:pg:mappo}
Multi-Agent PPO (MAPPO) extends PPO~\citep{PPO} to the centralized critic MARL case~\citep{MAPPO}.
The motivation behind PPO is to adjust the magnitude of the policy update to learn quickly without becoming unstable. 
PPO does this using a simple clipped loss that approximates the more complex trust region update \citep{TRPO}.

Like IA2CC (Algorithm \ref{alg:IA2CC}), MAPPO uses an advantage-based update (but not the agent-specific counterfactual one used in COMA). In particular, instead of the loss being $\gamma^t \delta_t \log\poli(a_{i,t}| h_{i,t})$ as reflected on line \ref{alg:IA2CC:actorgrad}, which is an approximation of $\gamma^t \jA_t \log\poli(a_{i,t}| h_{i,t})$,  the loss becomes:\footnote{Here, I consider the stochastic (not continuous) action case and for simplicity do not include GAE, mini-batching, or a policy entropy term (which is common in policy gradient methods to improve exploration). Also, the paper uses a local advantage loss, $\Ai$, but this value would not be different than $\jA$ in a general Dec-POMDP if it is updated using  $\jA$ and $r$ since these quantities would be the same for all agents. Finally, the paper lists policies over single observations rather than histories, which are typically not sufficient for partially observable problems (general Dec-POMDPs) and state-based critics are used, which are incorrect as discussed below.}
 \begin{equation}
\mathcal{L}^{MAPPO}_{clip}(\ppi)= \min\Big(r_{\ppi,i}\jA,\text{clip}(r_{\ppi,i},1-\epsilon,1+\epsilon)\jA\Big),
\text{\ where\ } r_{\ppi,i} = \frac{\pi_{\ppi}(\ai|\hi)}{\pi_{\polparam_{i,old}}(\ai|\hi)}.
 \label{eq:mappo:actor}
 \end{equation}
 Maximizing $\frac{\pi_{\ppi}(\ai|\hi)}{\pi_{\polparam_{i,old}}(\ai|\hi)}\jA$ seeks to maximally improve the new policy, $\pi_{\ppi}$, compared to the old policy, $\pi_{\polparam_{i,old}}$, by reweighing the advantage using an importance sampling ratio (considering the advantage was calculated using the old policy). 

 It can be difficult to estimate this value from a small number of samples and it may result in too large of an update (resulting in parameters that perform worse). As a result, PPO also considered a term using a clipped ratio, limiting $r_{\ppi,i}$ so it can't be too far above or below 1 (which would be the value when the new policy is equal to the old one). By minimizing over the unclipped and clipped values, PPO limits changes in the policy. Note the ratio is always positive but the advantage could be positive or negative. When the advantage is positive, the loss will be clipped if the ratio is too large, limiting how much more likely the action can be in the policy. Similarly, when the advantage is negative, the loss will be clipped when the ratio is near 0, limiting how much less likely the action can be in the policy. 
Like in IA2CC, $\jA$ can still be approximated as $r_t + \gamma \jVmodel(\jh_{t+1}) - \jVmodel(\jh_t)$.

The critic loss for MAPPO also includes clipping (with separate clipping parameter $\epsilon_v$) and is given by:
\begin{equation}
    \mathcal{L}^{MAPPO}(\valparam)=\max \left[ (\jV(\jh_t)-\hat R_t)^2,\left(\text{clip}(\jV(\jh_t),\jV_{old}(\jh_t)-\epsilon_v ,\jV_{old}(\jh_t)+\epsilon_v)-\hat R_t\right)^2 \right]
    \label{eq:mappo:critic}
\end{equation}
where $\hat R$ are Monte Carlo returns starting from the current history (i.e., $\hat R_t=\sum_{j=t}^{\hor-1} \gamma^{j-t} r_j$) and $\jV_{old}$ represents the value function from the previous iteration but TD can also be used and value clipping does not have to be used.

All agents use the same centralized critic and parameter sharing can also be used so all agents share the same actor network.

\paragraphs{Independent PPO (IPPO)} is a version of MAPPO where each agent uses its own local advantage rather than the joint advantage. As a result, the IPPO actor loss becomes:
 \begin{equation}
\mathcal{L}^{IPPO}_{clip}(\ppi)= \min\Big(r_{\ppi,i}\Ai,\text{clip}(r_{\ppi,i},1-\epsilon,1+\epsilon)\Ai\Big),
 \label{eq:oppo:actor}
 \end{equation}
 with $r_{\ppi,i} = \frac{\pi_{\ppi}(\ai|\hi)}{\pi_{\polparam_{i,old}}(\ai|\hi)}$ as before and $\hat \Ai=r_t + \gamma \Vimodel(\hiT {t+1}) - \Vimodel(\hiT t)$. 
The IPPO critic loss becomes: 
\begin{equation}
    \mathcal{L}^{IPPO}(\valparam)=\max \left[ (\Vi(\hiT t)-\hat R_t)^2,\left(\text{clip}(\Vi(\hiT t),V_{i,old}(\hiT t)-\epsilon_v ,V_{i,old}(\hiT t)+\epsilon_v)-\hat R_t\right)^2 \right]
    \label{eq:ippo:critic}
\end{equation}
where $\hat R$ are Monte Carlo returns starting from the current \emph{local} history, $\hiT \ts$. Parameter sharing is used with IPPO so all agents share the actor and critic network, making it a CTDE approach. As a result, there is only one actor and one critic, but it is updated using the data from all agents. \\

MAPPO and IPPO perform very well on the standard benchmark domains. Both algorithms typically perform similarly, but a version of MAPPO where the information in the critic was hand designed to only include features that are relevant to the task could outperform the standard versions of MAPPO and IPPO. This makes sense as the agents did not have to learn what information was necessary, which is difficult in high-dimensional settings. 

\subsection{State-based critics}
\label{sec:ctde:pg:states}

Critics that use state (and not history) are unsound in partially observable environments (i.e., general Dec-POMDPs). 
That is, they are incorrect and could be arbitrarily bad. 
They can work well in domains that are nearly fully observable (such as SMAC \citep{SMAC}) and domains in which remembering history information is not helpful (observations are not very informative or the task is too hard to solve). 
This result has been shown theoretically and empirically \citep{AAAI22,JAIR23}. 

In particular, state-based critics estimate $\jQ(s,\ja)$ or $\jV(s)$. 
MADDPG \citep{MADDPG} and COMA \citep{COMA} popularized the idea of using state-based critics using the intuition that this ground truth information that is available during training can help address partial observability. 
In truth, it is typically a bad idea to eliminate partial observability in the critic. A Dec-POMDP typically \emph{is} partially observable and the actor is using partially observable information (i.e., histories). The state-based critic can be biased (i.e., incorrect) because there is a mismatch between the actor and the critic. The actor needs to use history information to choose actions. The state-based critic uses state values instead. As a result, the state values are averaged over the histories that are visited and can't possibly have the correct history values unless there is a one-to-one mapping from states to histories (which would happen in the fully observable case). That is, the critic loses the information about partial observability, which is necessary for the actor to make good choices. 

For example, in the classic Dec-Tiger problem \citep{Nair03IJCAI,Book16}, there are only 2 states (the tiger being on the left or on the right) but noisy observations are received when an agent listens (an agent can also choose to open one of the two doors). A state-based critic would only have two values $\jV(\sI 1)$ and $\jV(\sI 2)$ but there are an exponential number of histories that depends on the horizon. As more information is gathered (i.e., the same observation about the tiger's location is heard multiple times) the value of the history, $\jV(\jh)$, should increase (so  $\jV(\jh_{tiger-left-10-times-in-a-row}) > \jV(\jhT 0)$, the initial history). This can't be reflected in the state-based critic. Furthermore, information-gathering actions (i.e., listening in this tiger problem) won't improve the value according to the critic so they shouldn't be taken.  As a result, the state-based critic values will be incorrect in the Dec-Tiger problem and the agents would learn a policy to open one of the doors and hope for a good outcome (assuming the tiger is randomly initialized). 

The updates to Algorithm \ref{alg:IA2CC} for the state-based critic case are straightforward---$\jVmodel(s)$ is used in place of $\jVmodel(\jh)$. Then, the TD error calculation becomes $\delta_t \gets r_t + \gamma \jVmodel(s_{t+1}) - \jVmodel(s_t)$, which is used in the actor and critic updates.  

It turns out that history-state critics, those that take both the state and history as input ($\jQ(\jh,s,\ja)$ or $\jV(\jh,s)$) are unbiased and often perform the best \citep{AAAI22,JAIR23,MAPPO}. This result has been theoretically and empirically shown for the general single and multi-agent cases \citep{AAMAS22Baisero,AAAI22,JAIR23} as well as empirically with variants of MAPPO \citep{MAPPO}. 
The intuition for correctness of the history-state critic is that no history information is lost (as it is in the simpler state-based critic). That is, the correct history value function can be (and is) recovered from the history-state value:
$\jQpol(\jh,\ja) = \mathbb{E}_{s\mid \jh} \left[ \jQpol(\jh, s, \ja) \right]$. 
The MAPPO paper also showed that if you can handcraft the features of the history-state critic to retain only the relevant information, the performance can be the highest since the agent doesn't need to learn which information is relevant and which is not. Unfortunately, determining which information is relevant and properly separating it can be difficult to do. 
The algorithmic updates for the history-state critic case are also straightforward---$\jVmodel(\jh,s)$ is used in place of $\jVmodel(\jh)$. The TD error calculation becomes $\delta_t \gets r_t + \gamma \jVmodel(\jh_{t+1},s_{t+1}) - \jVmodel(\jh_t,s_t)$, which is used in the actor and critic updates.

\subsection{Choosing different types of decentralized and centralized critics}
\label{sec:ctde:pg:critics}

As noted in Section \ref{sec:dte:other}, while CTDE methods are popular, they don't always perform the best and are often quite close to DTE methods. This is because DTE methods typically make the concurrent learning assumption where all the agents learn using the same algorithm, making updates at the same time on the joint data. 
As mentioned, while centralized critic actor-critic methods are often assumed to be better than DTE actor-critic methods, they are theoretically the same (with mild assumptions) and often empirically similar (and sometimes worse) \citep{Peshkin00UAI,MAPPO,JAIR23}.

The theory assumes learned critics (or sufficient Monte Carlo estimates) but different types of critics may be easier to learn in different cases. 
As a result, it is not straightforward to determine when to use each critic in practice. 
Centralized critics often cause higher variance actor updates since information from the other agents needs to be marginalized out (e.g., through sampling) while information from the other agents is already removed from the decentralized critics. 
Centralized critics can also be more difficult to learn in problems with large numbers of agents or with large action or observation sets, making decentralized critics more scalable. 
Decentralized critics may be difficult to learn due to agents changing their policies (i.e., nonstationarity), making critic estimates stale. 
State-based critics are often the easiest to learn (because no history representation needs to be learned) but are biased in partially observable settings and will have high variance (again, because of the mismatch between the actor and the critic). 
In general, learning a history representation is difficult. Agents need to figure out what history information is relevant and what is not from a noisy and often sparse RL signal. Using a recurrent network (or even a transformer) is usually not sufficient for this task. 
History-state critics further increase the variance in the policy update (because the state information needs to also be marginalized out) but can often be easier to learn than centralized critics with only history information.\footnote{The exact reason why this is the case is unclear and a great topic for future research!} 
As a result, the choice of critic is often a bias-variance tradeoff since, in practice, a centralized critic should have a lower bias than decentralized critics with more stable values that are more easily updated when policies change but higher variance because the policy updates need to be averaged over other agents.

\subsection{Methods that combine policy gradient and value factorization}

Several methods combine using a centralized critic with value factorization. 
One notable method is
FACtored Multi-Agent Centralized policy gradients (FACMAC) \citep{facmac} which extends MADDPG to include a QMIX-style factored centralized critic, but since the local Q-values don't need to be used for action selection (the actor does that), the factorization can be nonmonotonic.  Also, rather than sampling the actions of other agents from the off-policy dataset, as MADDPG does, actions for other agents are sampled from the current policies during the actor updates. 
Decomposed Off-Policy policy gradient (DOP) \citep{DOP} uses a factored centralized critic that is a weighted sum of the local Q-values 
and a multi-agent extension of tree backup \citep{TreeBackup} to more efficiently calculate off-policy updates. While a centralized critic is learned, each agent's local Q-value estimate is used to update its actor and with mild assumptions the method is shown to converge to a local optimum even with the simple critic. 
Learning Implicit Credit Assignment (LICA) \citep{LICA} also uses an unconstrained factored centralized critic, but implements it as a hypernetwork that maps the state to the weights which mix the agents' action representations into the joint value.
Credit assignment is implicit in the sense that each policy is updated by backpropagating the joint action-value gradient rather than by comparing against a baseline as in COMA. The critic conditions on the state while the policies condition on observations, so it is an (incorrect) state-based critic of the type
discussed in Section \ref{sec:ctde:pg:states}, and the stated purpose of the hypernetwork is to increase how much state information reaches the policy gradients.

\subsection{Other centralized critic methods}

Many other methods use centralized critics. For instance, 
\cite{MAAC} use attention for determining which other agent information to include in the centralized critic. Other approaches are related to the ones presented here but some form of MAPPO is still the most widely used centralized critic method. 
Learning Individual Intrinsic Reward (LIIR) learns an intrinsic reward for each agent  \citep{LIIR}. Specifically, the approach learns two critics, the standard (extrinsic) one and a proxy critic that sums the intrinsic and extrinsic rewards. Each policy is updated from its proxy critic, while the intrinsic reward parameters are updated through that update to maximize the extrinsic return, so the intrinsic rewards change how credit is distributed without changing what is optimized. While LIIR can perform well on a number of domains it is often outperformed by newer methods.

\section{Other forms of CTDE}
\label{sec:ctde:other}

While the methods above are the ones that are most widely used, there are several other forms of CTDE. I discuss common approaches and include a note on the many other topics that are not in this text in order to focus on core issues of CTDE. 

\subsection{Adding centralized information to decentralized methods}

Decentralized training and execution (DTE) methods can be augmented with centralized information to potentially improve performance. Such methods are only possible when a centralized training phase is available for sharing information but the resulting methods seek to balance scalability and performance. These extensions are discussed below. 

\paragraph{Parameter sharing}

Many cooperative MARL methods use parameter sharing. As noted in Section \ref{sec:ctde:pg:mappo}, the idea behind parameter (or weight) sharing is instead of each agent using a different network to estimate the value function or policy, all agents share the same networks. 
More details about parameter sharing are given in Section \ref{sec:dte:other}. 
Since parameter sharing requires agents to share networks during training, I consider it a form of CTDE. 

\subsection{Sequential agent updates}
\label{sec:sequpdates}
A number of methods allow agents to take turns learning. For instance, DTE methods can be used but all agents are fixed (not learning) except for one. The learning agent learns until convergence, generating a best-response to the other agent policies. If this process continues until no agents can further improve their polices, the result is a Nash equilibrium \citep{Banerjee12,ma2ql}. 
In limited settings with additional strong assumptions (e.g., deterministic environments, full observability, additional coordination mechanisms to ensure coordinated policies), such methods can potentially converge to optimal solutions \citep{Lauer00ICML,jiang2023best,jiang2022i2q}. 
While these methods are sometimes called `decentralized' since coordination and communication are needed during learning that is not available during execution, I consider them to be CTDE.

\cite{kuba2022trust} use a form of sequential agent updates in methods including Heterogeneous-Agent Proximal Policy Optimisation (HAPPO). 
Specifically, they formalize the joint advantage of an ordered subset of agents as the sum of advantages of each agent given the actions already taken by all preceding agents in the sequence (sequential advantages \citep{Kuba24}). 
 That is, they first define a multi-agent Q-function for an ordered subset of $k$ agents $i_{1:k}$  as the expected return when agents $i_{1:k}$ take actions $\ja_{i_{1:k}}$, and all remaining agents' actions are marginalized out under $\pi$:
\[
\jQ_{i_{1:k}}^\pi(s, \ja_{i_{1:k}}) = \Exp_{\ja_{-i_{1:k}} \sim \pi_{-i_{1:k}}} \left[ \jQ^\pi(s, \ja_{i_{1:k}}, \ja_{-i_{1:k}}) \right]
\]
When $k=0$, no agents have chosen actions so this reduces to $\jV^\pi(s)$ and when $k=\nrA$, this is just $\jQ^\pi(s,\ja)$.
Then, the \textit{sequential advantage} of agent $i_j$, given that agents $i_{1:j-1}$ have already taken actions $\ja_{i_{1:j-1}}$, is the marginal gain (difference) for agent $i_j$ taking action $a_{i_j}$:
\[
\jA_{i_j}^\pi(s, \ja_{i_{1:j-1}}, a_{i_j}) = \jQ_{i_{1:j}}^\pi(s, \ja_{i_{1:j}}) - \jQ_{i_{1:j-1}}^\pi(s, \ja_{i_{1:j-1}})
\]
This can be viewed as fixing the other agents' policies (either integrating over their actions under those policies or conditioning on sampled actions) to compute a single-agent-style advantage for the current agent.
The \emph{multi-agent advantage decomposition lemma} states that 
given a joint policy $\jpol$, state $s$, and ordered agent subset $i_{1:m}$ the advantage for the subset is the sum of the sequential advantages:
\[
\jA_{i_{1:m}}^\pi(s, \ja_{i_{1:m}}) = \sum_{j=1}^{m} \jA_{i_j}^\pi(s, \ja_{i_{1:j-1}}, a_{i_j}),
\]
allowing advantages to be calculated sequentially given fixed previous agent actions. While this lemma holds by the earlier sequential advantage and marginal Q-value definitions, it allows a sequential update scheme with monotonic improvement guarantees (in the fully observable case). The sequential update orders the agents and then uses the advantage decomposition to update the agents in that order (using the advantage definition above). Importance sampling is used to avoid collecting new data after each agent update. This approach is general as it can be applied to policy gradient and action-value methods \citep{Kuba24}. Policy gradient methods are a more natural fit with versions that build off of PPO \citep{PPO} and TD3 \citep{TD3} performing the best. Since they also remove parameter sharing, they call the methods heterogeneous agent PPO (HAPPO) and heterogeneous agent TD3 (HATD3). 
Note that the theoretical guarantees only hold in the fully observable case and only guarantee convergence to some Nash equilibrium (which could be good or bad). 
Also note that all of these sequential update methods only assume they know the other agents' policies during training but they still use decentralized policies so agents don't (explicitly) condition on other agent actions or require them as input during execution. 

A closely related method is the Multi-Agent Transformer (MAT) \citep{MAT}, discussed in Section \ref{sec:cte:MAT}, which builds on the same advantage decomposition but uses the sequential structure differently. HAPPO updates the agents' policies in sequence while executing them in parallel, while MAT updates in parallel and selects actions in sequence, which requires centralized execution. MAT-Dec removes the sequential structure from both, updating the agents in parallel and having them act in parallel from their own observations, which leaves a centralized critic with independent actors like those in Section \ref{sec:ctde:pg}. 

\paragraph{Addressing nonstationarity}

Decentralized learning methods face a nonstationary problem due to changing policies of other agents. Some methods try to address this challenge by directly modeling these changes. For example, \cite{Foerster17} propose using importance sampling to correct for the probability differences and decay old data or simpler `fingerprints' (e.g., episode number) to mark the data's age. 
Other methods try to model the other agent learning updates as well \citep{LOLA,COLA}.

\subsection{Decentralizing centralized solutions}

Another potential CTDE approach is to learn a centralized solution during training and then (attempt to) decentralize it before execution. This is much harder than it seems. 
A centralized policy can map $\jhS \to \jaS$ without the constraint (in the stochastic case) that $\jpol(\ja| \jh)=\prod_{i \in \agentS} \poli(\ai| \hi)$. 
That is, each agent's policy can depend on \emph{other agent histories}: $\jpol(\ja| \jh)=\prod_{i \in \agentS} \poli(\ai| \jh)$. The centralized policy class is much larger and richer than the decentralized policy class. 

For example, just considering deterministic policies of horizon $\hor$, there would be 
$\left(|\aAS i|^{|\oAS i|^{\hor}}\right)^{|\agentS|}$ possible decentralized policies (assuming all agents have the same size action and observation sets) vs.~$|\jaS|^{|\joS|^\hor}$ possible centralized policies. 
Plugging in 4 actions per agent, 5 observations, a horizon (history-length) of 10 and 4 agents gives 
$\left(4^{5^{10}}\right)^{4} \approx 2.6\times 10^{23{,}517{,}968}$
vs.~$\left(4^{4}\right)^{\left(5^{4}\right)^{10}} \approx 5.23\times 10^{21{,}902{,}814{,}890{,}489{,}476{,}617{,}493{,}993{,}742}$. This is a massive difference. 

As a result, most centralized policies will not be directly decentralizable in the sense that they will be equivalent.\footnote{Note that this is not true in the fully observable case as discussed in Section \ref{sec:fullobs}.}  Policies can only be decentralized when they don't depend on other agent information but centralized agents that make joint action choices based on the information of all agents can often perform much better. In fact, this centralized partially observable case is called a multi-agent POMDP and can be solved using single-agent methods as discussed in Chapter \ref{sec:cte}. 

As a result of these issues, approaches that decentralize centralized solutions are not common. Nonetheless, there are some instances. For example, 
\cite{liexplicit} attempt to reconstruct the global information using a reconstruction loss, other methods attempt to mimic a centralized controller \citep{lin2022decentralized}, and others allow communication during training but reduce or remove it during execution \citep{Foerster16,Wang2020}.
FOP (for Factorizes the Optimal joint Policy) \citep{FOP} makes the assumption that centralized solutions are decentralizable and then seeks to learn solutions using a max entropy formulation along with a value factorization scheme that is similar to  methods in Section \ref{sec:ctde:vff}. While FOP (and other methods) could be applied in general Dec-POMDP settings, they are likely to perform poorly when the centralized solution is sufficiently different from the decentralized solution.  

\chapter{Other topics}

\section{Communication}
\label{sec:other:com}
As mentioned in \cite{Book16}, communication has been widely studied in planning and learning for Dec-POMDPs. While the Dec-POMDP model allows for implicit communication through the actions and observations (i.e., actions can provide information that other agents see through observations), \emph{explicit communication} methods have separate communication actions (in addition to standard domain actions). For instance, the Dec-POMDP-Com model extends the Dec-POMDP to include a set of communication messages along with corresponding costs \citep{Goldman04JAIR}. Of course, such models can also be represented by a standard Dec-POMDP (by adding the communication messages to the action and observation sets and adding the costs into the reward) but it may be more intuitive or scalable to consider communication separately. 
It is also worth noting that with instantaneous, free, and noiseless communication, a Dec-POMDP (or Dec-POMDP-Com) can be transformed into an MPOMDP (see Section \ref{sec:cte:models}) by each agent sharing their observations with the other agents at each step and such an information sharing policy is optimal (i.e., from a theoretical perspective, if agents can share information at each step, they should always just share their observations)~\citep{Pynadath02JAIR}. 
Since sharing all information is typically intractable, communication methods typically differ in what the agents communicate, when they communicate, and which other agents they communicate with.

\paragraph{Learning what to communicate}
In the modern deep MARL era, perhaps the first to explore communication for the partially observable setting was \cite{Foerster16}. While the exact problem being solved isn't formalized, it fits with the Dec-POMDP setting. Specifically, two algorithms are proposed---Reinforced Inter-Agent Learning (RIAL) and  Differentiable Inter-Agent Learning (DIAL). These methods are based on DRQN \citep{DRQN} (see Section \ref{sec:back:value}) and in both cases, each agent learns a Q-function that includes the current observation, the message on the previous time step, the previous history representation, and the current action: $\Qi(o_{i,t},m_{i,t-1},h_{i,t-1},a_{i,t})$. RIAL is a form of independent DRQN (see Section \ref{sec:value:IDRQN}) that doesn't use a replay buffer (because, as noted, it can destabilize learning) and breaks up the maximization to first maximize over $\aAS i$ to select the action and then maximizes over $\set{M}_i$ to select a communication message. A parameter sharing (CTDE) version of RIAL is also introduced where all agents use the same Q-network and an agent id is used to allow agents to be more heterogeneous. DIAL is similar but allows for continuous communication messages during training. Specifically, DIAL uses a Q-network with two types of outputs, the standard IDRQN outputs for actions as well as a real-valued output that could be used a continuous communication. The continuous communication is optimized using backpropagation across agents but discretized so agents can use discrete communication messages during execution. DIAL also has a parameter shared version and a non-parameter shared version. DIAL (with parameter sharing) performed the best in their experiments and these methods were very influential in later communication methods for MARL. 
Other methods learn
what to share more efficiently \citep{CommNet,BiCNet}.
 For instance, CommNet uses multiple rounds of communication through a feedforward network to learn relevant information to share with a REINFORCE-style algorithm \citep{CommNet}.
 
 \paragraph{Learning when to communicate}
 When communication has a cost, an agent also has to decide whether a message is worth sending.
  IC3Net extends CommNet to learn a binary gate to decide to send a message or not \citep{IC3Net}. 
  Variance Based Control (VBC) builds on VDN or QMIX so that a message received from another agent is converted into an adjustment to the recipient's local action values \citep{VBC}. Since the adjustment is added to those values, a message can only change a decision if it affects the agent's
actions differently. An agent then asks for information only when its own best two actions are close in value and the other agents reply only when their message would change that comparison.
  
 \paragraph{Learning who to communicate with}
  Learning Attentional Communication for
Multi-Agent Cooperation (ATOC) uses attention to choose subsets of agents to communicate, forming communication groups that share their observation and action information and learn to act using an independent form of DDPG over groups of agents (instead of agents) \citep{ATOC}.  
Targeted Multi-Agent Communication (TarMAC) uses attention to learn communication messages as well as message 'signatures' to help direct messages to the most relevant agents \citep{TarMAC}. Agents also learn queries to decide how to weigh signatures (with a dot product) to produce the attention weights for a message. 
Unlike ATOC, communication is learned directly from the RL loss rather than through an additional loss. 

Many other communication methods have been developed over the years. Additional information can be found in surveys on communication \citep{zhu2024survey} as well as other thoughtful papers about communication \citep{Ilowepitfalls}. 

\section{Exploration}
\label{sec:other:explore}
Exploration is an important topic in reinforcement learning. If agents never experience good behavior, how can they learn about it? As a result, exploration has been widely studied \citep{RLexploration,RLexploration2}. 
In the multi-agent context, there is the added question of how coordinated the exploration should be. That is, agents should not only explore new actions in new histories but also consider the exploration of other agents. In some cases, we may want agents to all have very similar policies but in other cases, agents must have different behavior. Balancing the agent's individual exploration with the exploration of the team as a whole while maintaining decentralized policies is a challenge. 

\paragraph{Committed exploration}
One of the most well-known MARL exploration methods is Multi-Agent Variational ExploratioN (MAVEN) \citep{MAVEN}.  MAVEN learns a distribution for a latent variable $z$, which is used to help with exploration by introducing a shared source of randomness.  Specifically, MAVEN builds on top of QMIX \citep{QMIX} (but could potentially use other MARL methods as a base) and the local Q-functions in QMIX are now conditioned on $z$. A single $z$ is sampled at the beginning of the episode and used throughout that episode, leading to committed exploration since the agents will use those Q-functions to choose actions for the whole episode. The approach is trained using both an RL loss and a (variational) mutual information loss maximizing the mutual information between $z$ and the joint trajectory. The intuition behind the mutual information loss is to learn different Q-functions (and thus different policies) for different $z$'s. MAVEN is a very influential method but is complex and can be outperformed by more recent methods. 

\paragraph{Intrinsic rewards}
Many MARL exploration methods use intrinsic reward to direct the agents toward under-explored areas. For example, 
\cite{Wang20} propose Exploration via Information-Theoretic Influence (EITI) and Exploration via Decision-Theoretic Influence (EDTI). EITI and EDTI  optimize an objective which is the sum of an extrinsic value function, an intrinsic value function and an interaction value for each agent $i$: $V^{ext,\jpol}(s_0)+ V_i^{int,\jpol}(s_0)+ \beta I_{-i|i}^{\jpol} $, where  $\beta$ is a weighting term and $V_i^{int,\jpol}$ is curiosity-based \citep{curiosity} (but other intrinsic rewards could be used). For simplicity, they assume the state is decomposed into a factor per agent and each agent observes that factor (i.e., an agent-wise factored Dec-MDP \citep{Book16}) and there are only two agents. With that assumption, EITI uses an interaction value that is based on maximizing the mutual information between the other agent's state and the current agent's state and action. The intuition is that this interaction value will encourage agents to visit states where they can influence the other agent's dynamics. Because some of these states may not be necessary to complete the task, EDTI attempts to calculate the Value of Interaction (VoI) which also incorporates value by calculating the expected difference in the Q-value when marginalizing out the other agent. The intuition in this case is that agents should visit states where interaction can cause the value to change. 
Both of these ideas are interesting but scaling EITI and EDTI  to more agents and (true) partial observability remains a challenge.

Other methods that use intrinsic rewards include CDS and EMC \citep{CDS,EMC}.
Celebrating Diversity in Shared MARL (CDS) learns an intrinsic reward using an information theoretic loss that seeks to maximize the mutual information between the agent id and the agent's trajectory \citep{CDS}. This intrinsic reward will encourage diverse behavior among the agents. The approach uses QPLEX as a base method along with partially shared individual Q-functions but the intrinsic reward could be used with any MARL method. 
Episodic Multi-agent reinforcement learning with Curiosity-driven exploration (EMC) \citep{EMC} has two main contributions, the use of curiosity and episodic memory. Curiosity is a way to direct agents to parts of the state space where there is uncertainty. In this case, EMC uses a value decomposition approach (they show results using VDN, QMIX, and QPLEX) and trains $\tilde \Qi$ predictor networks to predict the $\Qi$'s from the value decomposition approach. The error between the predicted and actual $\Qi$'s is used for an intrinsic reward that is added to the extrinsic reward when training $\jQ_{tot}$.  
The episodic memory is used to remember high-valued returns and use them to help train the $\jQ_{tot}$. Specifically, states are encoded and the highest-valued return is retained within a threshold of the encoded state. There is another loss added when training $\jQ_{tot}$ which regresses it towards the episodic memory return value at the corresponding state. 
Both concepts were developed in the single-agent context \citep{curiosity,episodiccontrol} but are being extended to the MARL case here. 
EMC can perform well on difficult SMACv1 tasks but the episodic memory may bias learning in stochastic domains. 

\paragraph{Exploring toward goals}
Some methods also use coordinated exploration towards a sampled goal \citep{CMAE,MASER}.
For example, cooperative multi-agent exploration (CMAE) \citep{CMAE} breaks up the state space using a `space tree' and samples states using count-based exploration within promising regions of the space to be the current goal. The space tree is used to expand the parts of the state space that are searched over time and new goals are sampled from these larger spaces. Separate exploration and target policies are learned to (cooperatively) reach the goal and optimize the return, respectively. CMAE can perform well but there are concerns about scalability and it requires domain knowledge to construct and grow the space tree. 
Multi-Agent reinforcement learning with Subgoals generated from Experience Replay (MASER) \citep{MASER} is a QMIX-like value function decomposition method that learns a joint $\jQ_{tot}$ as well as $\Qi$'s for each agent. 
A joint observation is chosen as a goal from the replay buffer based on a weighted combination of $\jQ_{tot}$ and the $\Qi$'s and intrinsic rewards are generated to reach the goal based on a distance metric from the current observation. The $\jQ_{tot}$ is trained based on joint rewards that are the extrinsic reward plus a sum of the intrinsic rewards and the  $\Qi$'s are trained based on local rewards that are based on an estimation of the agent's contribution to the extrinsic reward along with the agent's intrinsic reward. New goals are sampled every M episodes. 
MASER can perform well on sparse-reward tasks but goal selection and the intrinsic rewards can sometimes be poor. 

\paragraph{Choosing an exploration method}
There are many other exploration approaches in MARL, including new methods that are not presented here. The state-of-the-art is somewhat unclear since methods will often perform well in some domains and less well in others. Methods such as intrinsic rewards and goals can also bias learning, causing poorer performance than the base method when the particular approach doesn't fit well with the domain. As a result, it often makes sense to try a standard MARL method first and then choose the appropriate exploration method based on domain characteristics if the standard methods perform poorly. 

\section{Hierarchy}
\label{sec:other:hier}

Due to scalability concerns and potential decomposability in MARL, hierarchical methods have been studied for many years. 

\paragraph{Early work}
One of the earliest treatments of hierarchy in multi-agent learning is the layered learning framework of \citet{stone2000layered}, where lower-level skills are trained first and subsequent learning takes place over those skills. The top-layer method, team-partitioned opaque-transition reinforcement learning (TPOT-RL) \citep{stone1999tpot}, is decentralized, with each agent estimating values over coarse action-dependent features in which each candidate action is classified by a skill learned at a lower layer. The setting is not formalized, but while the task is cooperative, agents optimize hand-designed local reward functions rather than a shared team reward.

Another early work is by  \citet{makar2001hierarchical} who extend the single-agent hierarchical method MAXQ \citep{dietterich2000maxq} to the cooperative case. There, agents share a task hierarchy and learn with decentralized semi-MDP Q-learning, but use joint
actions only at designated cooperative subtasks near the top. This is the IL/JAL distinction of Section~\ref{sec:dte:value} applied where the joint space is small enough to be learnable. The model is not formalized, but it seems to be a factored Dec-MDP \citep{Book16} where each agent observes the shared features and its own component, and subtasks terminate on events rather than states. Communication is also used with each agent broadcasting their current subtask as a proxy for its unobserved state. \citet{ghavamzadeh2004communicate} also add communication costs to the hierarchy, so agents learn when to communicate. Additional details are in \citet{Ghavamzadeh06JAAMAS}. 

\paragraph{Macro-actions}
Both of these lines of work show that coordination is cheaper at an abstract level and \citet{stone2000layered} further argue that agents cannot assume synchronized decision points, but neither provides a formal model for temporally extended actions under partial observability and decentralized execution. This motivates the work on macro-actions (i.e., temporally extended actions which may require different amounts of time). 
Macro-actions enable the decision-making to take place at a higher level---at the level of deciding which macro-actions to execute---and are executed to completion. 
Macro-actions also more naturally represent real-world behavior which may be complex and require multiple time-steps (e.g., navigation to a waypoint, lifting an object or waiting for another agent) and can be defined by a (human) expert, making them more explainable than other sequences of primitive actions. 
Macro-actions are built from options \citep{Sutton99} but extended to the multi-agent case. Specifically, extensions of Dec-POMDPs to include macro-actions have been developed in the form of  Macro-action Dec-POMDP (MacDec-POMDP)~\citep{JAIR19} and decentralized partially observable \emph{semi-}Markov decision process (Dec-POSMDP)~\citep{IJRR17DecPOSMDP}. 

While macro-actions chosen by the agents no longer terminate at the same time, Dec-POMDP algorithms can be extended to generate asynchronous policies with macro-actions. The resulting algorithms need to reason about how far along other agents are in their macro-actions (implicitly or explicitly).
DTE and CTE value-based methods have been developed \citep{CoRL19}, along with CTDE value-based approaches \citep{ICRA20}, centralized and decentralized critic actor-critic methods \citep{NeurIPS22Xiao}, value function decomposition methods \citep{IROS26} as well as methods for improving credit assignment \citep{liang2025asynchronous} and better use of temporal information \citep{jung2025agentcentric}. 
By extending MARL algorithms to the macro-action case, the horizon and action space often become much smaller, allowing very large problems to be solved, including for multiple mobile  robots in warehouses \citep{IJRR25}, logistics \citep{IJRR17MacDec}, and aerial delivery \citep{IJRR17DecPOSMDP}. 

\paragraph{Agent roles}
Learning agent roles can also be considered a form of hierarchy---agents choose a role and then the agent's policy or local Q-value can be parameterized by the role.  
ROMA \citep{ROMA} builds on QMIX \citep{QMIX} learning a role embedding from each agent's observation which is then used by a hypernetwork to generate the parameters of that agent's local Q-network. Auxiliary losses encourage roles to be identifiable (maximizing the mutual information between an agent's role and its trajectory, given the observation) and specialized (limiting the number of dissimilar agent pairs so that roles are compact but distinct).
RODE \citep{RODE} also builds on QMIX, but uses two levels: a role selector assigns each agent a role every $c$ steps, and each role restricts the agent to a subset of the actions, so the lower-level policy only chooses among the actions available to its current role. Roles are discovered by clustering actions according to their effects on the environment and the other agents. As a result, a role is a set of actions with similar effects rather than an assigned subtask. 
\citet{yang2020hierarchical} use unsupervised skill discovery where a skill can be seen as a role. Each agent selects a latent skill and then chooses actions conditioned on it, with skill selection trained by QMIX and the low-level policies trained by independent Q-learning. The low-level reward combines the team reward with an intrinsic reward based on the accuracy of a decoder that predicts the skill from the resulting trajectory. This is related to the mutual information objective that ROMA uses as a regularizer, but here it is used as a reward. 
HAVEN \citep{HAVEN} also uses QMIX at both levels, where a high-level policy selects one of a fixed number of latent variables every $k$ steps and the low-level Q-function is conditioned on it. The high-level advantage is used as an intrinsic reward for the low level, but it is computed from a state-based value function and is therefore biased in partially observable settings (Section~\ref{sec:ctde:pg:states}). These latent variables are called macro actions, but they have no termination condition and all agents reselect at the same time steps, so they are more similar to roles. 
It is worth noting that unlike macro-actions, roles do not change the timing of decisions. Agents still act at every step and the policy is conditioned on the role so these methods gain specialization and reduced exploration but not necessarily scalability in the horizon.

\paragraph{Managers and workers}
Another class of methods has an upper-level agent (e.g., manager) directing the others (e.g., workers) rather than each agent selecting its own role.
For example, COPA \citep{COPA} adds a coach that observes the full state and sends each agent a message every $T$ steps, which plays a similar part to a role in the methods above but is now chosen by a separate agent rather than by each agent itself. The goal of this work is generalization to different team compositions rather than temporal abstraction.
Feudal Multi-Agent Hierarchies (FMH) \citep{ahilan2019feudal} uses a similar structure to COPA, with a separate agent at the upper level, but makes the hierarchy more explicit. A manager agent selects subgoals from a hand-designed set and sends them to the worker agents, which receive intrinsic rewards for achieving the subgoal they were given. This extends feudal RL \citep{dayan1993feudal, vezhnevets2017feudal} to the multi-agent case. Workers are first pretrained on subgoals that the manager samples uniformly. The approach is built on DDPG \citep{DDPG} with parameter sharing. 
ALMA \citep{ALMA} instead has the upper-level agent assign agents to subtasks, selecting a new allocation every $N_t$ steps after which agents act in a decentralized manner within their assigned subtask. Because the number of allocations is exponential in the number of agents, the allocation is sampled from a learned proposal distribution that assigns one agent at a time.
In all cases, the upper-level agent requires information about the other agents during execution, so these methods are not decentralized, and FMH also gives the manager and the workers different reward functions, so that problem is no longer a Dec-POMDP.

\section{LLMs as cooperative agents}
\label{sec:other:llms}

Many recent approaches have applied large language models  to multi-agent settings. Much of the best-known work, such as Generative Agents \citep{park2023generative}, AutoGen \citep{AutoGen}, MetaGPT \citep{MetaGPT}, CAMEL \citep{CAMEL}, and ChatDev \citep{ChatDev}, builds systems of LLM agents for software engineering, social simulation, or role-play. 
These systems use multiple agents to manage the scale and complexity of a single task, similar to the centralized methods of Section \ref{sec:cte:scale}. The agents share all information (through communication or direct observation), they don't optimize a return or learn and the interaction protocol is fixed by hand. Nevertheless, there are a number of methods that are more similar to those in this text which are discussed below. 

\paragraph{Prompting pretrained LLMs} The most common approach uses pretrained LLMs as policies, without any learning. 
These use a hand-written function to convert the state into a text description, a prompt template gives the task rules and the currently legal actions, the LLM is prompted to choose a high-level action, and that choice is translated into actions that the environment accepts (i.e., domain dependent).
For example, CoELA \citep{zhang2024building} formalizes the problem as a Dec-POMDP-Com (Section \ref{sec:other:com}) but the planning module enumerates the high-level plans that are currently available and prompts an LLM to choose one of them (without consideration of rewards). 
The chosen plan is carried out by hand-designed low-level controllers, so the high-level plans play the role of the macro-actions of Section \ref{sec:other:hier}, and the agent's memory of the map, the task progress, and the other agents serves as a hand-designed
history representation. Sending a message is one of the plans available to the planner, so the agent decides when communicating is worth it. 
\citet{chen2024scalable} assume that every agent has full knowledge of the environment and compare a centralized LLM vs.~round robin communication to find a solution as well as hybrid models. 
In RoCo \citep{mandi2024roco} the agents speak in turn and each one sees what the others have said, and a centralized motion planner executes the agreed plan. ProAgent \citep{zhang2024proagent} uses a single LLM agent that never communicates and instead infers its (fixed) partner's intentions from their behavior (like a single-agent POMDP). \citet{li2023theory} has agents with private observations and no means of sharing information other than messages, but messages are broadcast to the whole team.

\paragraph{What the evaluations show} Several benchmarks assess how well LLM agents coordinate without any training (i.e., so MARL isn't involved). 
For instance, \citet{agashe2025llm} pair frozen LLM agents either with a copy of themselves or with a held-out partner, as is common in self-play and the zero-shot coordination problem~\cite{pmlr-v119-hu20a} rather than training the team together.  Also, the LLM agents are given the rules of the game in natural language along with hand-built scaffolding that the RL baselines do not receive. The results show that the LLM agents are competitive with PPO and population-based training on Overcooked, which is fully observable, but reach about half the score of the strongest methods on Hanabi, which has significant partial observability. 
The LLM agents don't seem to degrade when the partner changes, potentially showing the generalization abilities of LLMs. 
\citet{sun2025collab} note that most benchmarks for LLM agents do not require collaboration at all, since a single agent can complete the tasks alone, and introduce a version of Overcooked in which each agent can reach only part of the kitchen and only one of them knows the recipe. On this benchmark, the agents respond to a request for help more reliably than they make one, which may be due to the fact that recognizing that a teammate has something you need requires reasoning about what that teammate knows and can do. 
\citet{cemri2025why} analyze 1642 execution traces from seven multi-agent LLM frameworks and report failure rates between 41\% and 86.7\%, with the largest share coming from how the systems are organized rather than from the models being used. In multi-agent organizations such as hand-coded pipelines and manager-worker hierarchies, they identify failures including agents that repeat work already done, do not recognize when the task has ended, lose their interaction history, and never check whether the task was completed. 

\paragraph{LLMs as components of MARL algorithms} 
Another group of methods keeps a standard MARL algorithm and uses a frozen LLM to augment it in some way.
instructRL \citep{hu2023language} addresses equilibrium selection for coordination with a human, since several conventions are equally good in self-play and a human partner may not know which one the agent adopted. In this work, a human states the desired convention in natural language, the LLM is asked whether each action is consistent with the instruction, and the answers become a prior policy that is added
to the Q-values during both action selection and bootstrapping. 
LAMARL \citep{zhu2025lamarl} uses an LLM to replace the hand-designed rewards and controllers that multi-robot tasks usually require in a setting with partial observability and individual rewards. Given a description of the task and a set of primitive functions, the LLM is asked what constraints the task imposes, what skills the robots need, and what sub-goals the reward should contain, and it then writes a reward function and a prior policy. 
The final policy is trained with an independent version of DDPG using the generated reward function and a behavior cloning loss toward the actions of the prior policy, with the weight of that loss annealed during training. The resulting method significantly improves sample
efficiency on their shape assembly task.
LLM-MCA \citep{nagpal2025leveraging} uses an LLM to solve the multi-agent credit assignment problem. A frozen LLM is given a description of the
environment together with the sequence of observations, actions, and global rewards from an episode, and it returns a numerical credit for each agent at each time step. The agents are then trained on those individual credits (using double DQN) rather than on the global reward. On level-based foraging, robotic warehouse, and a new benchmark, this outperforms QMIX, MAPPO, and LICA. Since the LLM is used only during training, the agents still execute in a decentralized manner.

\paragraph{Training LLMs with cooperative MARL} The final group of methods treats the LLMs as the policies and optimizes them with cooperative MARL. \citet{AAAI26} and \citet{ICML26} formalize LLM collaboration as a Dec-POMDP in which a local observation is the prompt given to an agent, a local action is the complete response that the agent generates, the horizon is the turn limit of the dialog, and all of the agents share a joint reward. Treating a whole response as a single action rather than each token as an action can be seen as a form of macro-action like those in Section \ref{sec:other:hier}.  Also, the dialog history is kept in the key-value cache of each agent, which serves as a history representation in place of recurrent layers and it stays private so that inference remains decentralized. MAGRPO \citep{AAAI26} updates each agent using Monte Carlo estimates of the joint return with a baseline computed from a group of sampled responses, which is the decentralized policy gradient of \citet{Peshkin00UAI} discussed in Section \ref{sec:dte:pg}. \citet{ICML26} show that these estimates are unbiased but high variance (as is standard in Monte Carlo methods). To overcome this issue, they compare a centralized critic conditioned on the joint history with decentralized critics conditioned on local histories, and find that all three methods (MAGRPO and the two critic-based approaches) perform similarly on short-horizon tasks with dense rewards while the centralized critic is considerably better on long-horizon and sparse-reward tasks, which is consistent with the discussion in Section \ref{sec:ctde:pg:critics}.

Large language models and other forms of generative and agentic AI are only getting more popular. This work suggests that the difficulties described throughout this book do not disappear when the agents are language models. Future work will have to further bridge the gap to improve performance using tools from MARL and LLMs. 

\section{Safe methods}
\label{sec:other:safe}
In many applications, agents must not only maximize return but also avoid certain outcomes entirely, such as colliding with each other or entering a restricted region. Safe MARL methods add constraints of this kind and try to satisfy them while still learning a good policy. This use of the term is somewhat narrower than the one used in discussions of AI safety, where the concern is with risks from many AI systems interacting, such as agents failing to coordinate, working at cross purposes, or cooperating in ways that are harmful to others \citep{Hammond25}. Those questions are largely outside the scope of this text (since they haven't been formulated in terms of MARL to the best of my knowledge), but coordination failure among agents that share a goal is a core problem that MARL is solving (from the MARL perspective, of course). MARL methods have been considered that incorporate constraints that the system must satisfy. Some approaches bound the expected cost of a policy (constrained methods), so individual violations remain possible, while others prevent unsafe actions at every step (shielding and barrier functions), which matters when the agents are learning on real hardware. All of these methods use some form of constraints but constrained formulations can be thought of as using soft constraints while shielding and 
control barrier functions use hard constraints (over discrete or continuous models, respectively). 

\paragraph{Constrained formulations}
One class of methods extends constrained MDPs \citep{Altman99} to the multi-agent case. \citet{Gu23} formulate the problem as a constrained Markov game, where the agents share a reward but each agent also has its own cost function and threshold. The goal is to maximize return while keeping the expected discounted cost of each agent below its threshold.  They give a policy iteration approach that improves the return monotonically while satisfying the constraints at every iteration, using the sequential update scheme and advantage decomposition of Section \ref{sec:sequpdates}. They derive two practical algorithms. The first enforces the constraint directly, recovering a feasible policy through a backtracking line search when an update would violate it, while the second replaces the constraint with Lagrange multipliers, which is simpler to implement but only approaches feasibility over a number of gradient steps. They also develop safety-aware versions of the multi-agent MuJoCo \citep{facmac} and Robosuite \citep{Robosuite} benchmarks.
\citet{E2C} note that this formulation gives each agent its own constraints, while safety in a cooperative problem is usually a property of the team, and extend their improvement result to constraints on the joint behavior. They also observe that constraining behavior suppresses the exploration needed to discover coordinated behavior, and add a bonus for visiting new observations to recover it.
CMIX \citep{CMIX} instead targets a Dec-POMDP with two kinds of constraints, one bounding the cost at each individual time step and one bounding the expected discounted cost. It converts the step-based constraint into a penalty on the rewards at that step, which allows a single factored action-value function to be learned per constraint (though the step-based constraints are then only enforced in expectation).

\paragraph{Shielding}
A different approach prevents unsafe actions more directly. A shield is synthesized from a safety specification written in temporal logic, and it either
restricts an agent to a set of safe actions before it chooses or corrects the action it has chosen \citep{Alshiekh18}. 
\citet{AAMAS21Ingy} extend this idea to the multi-agent case with a centralized shield over the joint action, along with factored shields that each monitor a subset of the agents, since the cost of synthesis grows exponentially with the number of agents.
Because a centralized shield can't be used during decentralized execution, \citet{NeurIPS22Melcer} decompose a centralized shield into one shield per agent, so each agent can choose independently while the joint action remains safe. These methods assume the fully observable case (for the safety-relevant information at least). \citet{RLC24} remove that assumption, handling agents that receive different observations and cannot communicate. Decentralized shield synthesis is undecidable in general, so their algorithm may report failure, but it applies in many cases where the earlier methods do not.

\paragraph{Control barrier functions}
Control theory offers a third approach. A control barrier function defines a set of safe states together with a condition on the action, where satisfying that condition at all times guarantees that the system never leaves the safe set. \citet{Qin21} learn a barrier function for each agent along with its control policy, where an agent's barrier function depends only on its own state and the states of the agents it can see, so the whole system is safe as
long as every agent satisfies its own condition. Safety then holds along every trajectory rather than in expectation (although only where the barrier function has been learned).  Since the dynamics are assumed to be known and differentiable, they train the policy by differentiating through the model rather than by estimating a gradient from sampled returns (as in policy gradient). 
\citet{CaiCBF} instead combine barrier functions with a standard MARL algorithm in a Markov game, giving each agent its own barrier functions that correct the action its policy proposes in the same way that a shield does. 
Both methods assume a known model with continuous time and a continuous control input, and the barrier condition is evaluated on the state, so neither applies directly to a Dec-POMDP.

More details on safe reinforcement learning (primarily for the single-agent case) can be found in the survey by \citet{Gu22survey}.

\section{Model-based methods}
\label{sec:other:model}
All of the methods discussed so far are model-free (learning values or policies directly from experience). Model-based methods learn a model of the environment and use it to generate additional experience, which can considerably reduce the number of environment interactions needed to perform well. Note that, for this section, the model is learned rather than given as in the planning literature for Dec-POMDPs \citep{Book16}. 

A model of a Dec-POMDP describes the joint dynamics so it is naturally centralized. As a result, where the model is used matters as much as how it is learned. The simplest choice is to use it during training only. MAMBPO \citep{MAMBPO} learns a centralized model that predicts the reward and the next joint observation from the current joint observation and joint action. As noted elsewhere, an observation model may be a poor approximation of the state-based model (which could also be used here). The approach generates imagined transitions from the model to train the agents (using a form of multi-agent SAC \cite{SAC}), and then executes the resulting policies without the model. 
Most of the other methods build on DreamerV2 \citep{DreamerV2}, a single-agent method that learns a latent-variable model of the environment from observations and then trains a policy entirely on trajectories imagined in that latent space.
MABL \citep{MABL} learns two latent states for each agent. The agent latent state is inferred from the agent's own observation (as in the single-agent case), while a second, global latent state is inferred from the true state. A term in the loss forces the agent latent state to be similar to one predicted using the global latent state. Policies are trained on the imagined trajectories with a form of MAPPO.

The alternative is to model each agent's local dynamics rather than the joint dynamics. This avoids the joint action space but leaves each agent's dynamics non-stationary (since they depend on the other agents). MARIE \citep{MARIE} and MAMBA \citep{MAMBA} both address this by adding a module that aggregates across agents and gives each agent's model a summary of what the others are doing, but they use it differently. MARIE discards the model after training, with the policies conditioning only on their own observation histories, while MAMBA keeps the models running at execution, where the aggregation becomes communication between the agents (Section \ref{sec:other:com}).

\section{Offline methods}
\label{sec:other:offline}

Offline methods learn from a fixed dataset of experience without further interaction with the environment, which matters when interaction is expensive or unsafe. The main difficulty in the single-agent case is extrapolation error, where the value function is evaluated at inputs that do not appear in the dataset. These estimates can't be corrected by interacting with the environment, so the error accumulates through bootstrapping. 
 \citet{ICQ} show that the rate at which error propagates from unseen inputs to seen ones depends on the size of the action space relative to the dataset, and the joint action space grows exponentially in the number of agents. 
They propose replacing the maximization in the target with an expectation over the actions recorded in the dataset, reweighted according to their estimated values so that it still improves on the behavior policy. 
Another approach for combating extrapolation error is to learn a conservative value function, one that deliberately underestimates the value of actions that are not in the dataset. \citet{OMAR} find that methods of this kind lose performance as agents are added, and attribute it to the shape this produces, which is non-concave in the action, so an actor following the gradient can stop at a poor local maximum. The effect vanishes when the task doesn't need coordination, so they add a sampling-based search over actions and a term pulling the actor toward the best one found.

A second group of methods avoids value learning altogether and treats the problem as sequence modeling.  
The multi-agent decision transformer (MADT) \citep{meng2021offline} pre-trains on an offline dataset and then fine tunes online. Two networks are pre-trained: an actor, which predicts the action taken in the dataset from the agent's own observation history using a cross-entropy loss, and a critic, which predicts the return from the state history. Learning the policy in this phase is behavior cloning with parameter sharing, and  the transformer serves as a history representation in the same way as the recurrent layers in Section \ref{sec:back:value} (unlike MAT, the transformer here runs over time for each agent separately rather than over agents). In the online phase, the pre-trained networks are used as the actor and critic in MAPPO (see Section \ref{sec:ctde:pg:mappo}). The critic is trained in the same way in both phases, but the actor's objective changes  so only the network initialization carries over for the policy. The paper shows fine-tuning with MAPPO requires considerably fewer environment
interactions than training MAPPO from scratch. Note that  since the critic conditions on states while the actor conditions on observation histories, it is
a state-based critic of the type discussed in Section \ref{sec:ctde:pg:states}, and using a sequence of states rather than a single state does not remove the bias.
MADiff \citep{MADiff} is a purely offline method where a diffusion model is trained on the dataset to generate observation sequences for all agents at once. It is conditioned on the return, so the reward selects which behavior to imitate rather than being used in a value estimate. Each agent then uses a learned inverse dynamics model to infer which action would take it from its current observation to the next one that is generated. At execution, an agent conditions the model on its own observation, predicting what the whole team will do and choosing its part. In general, this may perform poorly since the observation transition also depends on what the other agents did and saw.  The authors also report that the method loses to a value-based one in stochastic environments and that it scales poorly  since every agent has to predict the future of every other agent.  

These methods differ in how they handle partial observability. ICQ conditions each agent's value function and policy on that agent's own action-observation history, and MADT does the same through its transformer. OMAR conditions on the current observation instead, so its policies and value functions are reactive as discussed in Section \ref{sec:singleobs}, and MADiff can condition on a window of past observations but treats its length as a hyperparameter, setting it to zero on some benchmarks.

\chapter{Conclusion}

\section{Topics not discussed}

There are many other issues that are important to MARL but are not discussed in this text, including topics such as self-interested agents, ad hoc (or zero-shot) coordination, other agent (opponent) modeling, transfer and multi-task learning, and coordination with humans. These topics, while important, are excluded here to maintain brevity and focus. Some surveys of these areas include ad hoc teamwork \citep{mirsky2022survey}, and modeling other agents \citep{AlbrechtStone18}. \citet{Zhang21theory} review theoretical results for MARL, although these are mostly for the fully observable case of Markov games, since results for the Dec-POMDP learning case are limited (but could potentially be extended from the planning case \citep{Book16}).

\section{Evaluation domains}

While earlier work focused on domains such as multi-agent particle environments \citep{MADDPG} and the Starcraft Multi-Agent Challenge (SMAC) \citep{SMAC}, recently, many more domains have become available. Some examples include a more complex version of SMAC that is no longer deterministic and includes more partial observability \citep{SMACv2} as well as soccer-based environments  \citep{googlefootball}. 
Other widely used domains include Hanabi, a card game in which each agent can see its partners' cards but not their own, so agents have to reason about what their partners know \citep{Bard20}, and Overcooked, a cooking game requiring spatial and temporal coordination that has also been 
used for coordinating with humans \citep{Carroll19}. As with SMAC, a second version of Overcooked was developed because the original is fully observable and gives the agents no private information \citep{Gessler25}. OvercookedV2 restricts what each agent can see and adds stochasticity. 
Level-based foraging and multi-robot warehouse are other common gridworld domains \citep{SEAC}, and \citet{Papoudakis21} compare a range of cooperative MARL algorithms across several of these environments.

Continuous control domains are less common but also available. Multi-agent MuJoCo divides the joints of a single MuJoCo robot among separate agents, each observing only part of the body \citep{facmac}, and JaxMARL provides MABrax, a JAX-based reimplementation of it \citep{jaxmarl}. Other
continuous domains include VMAS, a vectorized simulator for a range of collective robotics tasks \citep{VMAS}, and Bi-DexHands, which poses bimanual manipulation problems in which each hand is controlled by a different agent \citep{BiDexHands}.

There are also beginning to be application-oriented domains such as power grid operation \cite{marchesini2026marlgridtr}, voltage control in distribution networks \cite{MAPDN}, and inventory management \cite{MABIM}.

Resources like pettingzoo\footnote{https://pettingzoo.farama.org/} provide environment code, while algorithm implementations are available through platforms such as PyMARL, which was released with SMAC \citep{SMAC}, EPyMARL, which extends it with actor-critic methods and additional environments \citep{Papoudakis21}, MARLlib, which is based on RLlib \citep{marllib}, and BenchMARL, which is based on TorchRL \citep{benchmarl}. JaxMARL provides both, reimplementing many of the environments above in JAX along with a set of algorithms, so that the environment and the learning can run together on a GPU and training runs can be executed in parallel \citep{jaxmarl}.

Evaluation practices have also been examined. For example, in a meta-analysis of 75 cooperative MARL papers, \citet{Gorsane22} find that most evaluate on a single environment, that a third report no measure of uncertainty at all, and that reported results for the same algorithm on the same SMAC maps differ substantially across papers. MARL is a fast-growing area but we have to do a better job ensuring the correctness of papers and evaluating the significance of the results. 

\section{Applications}

Cooperative MARL is increasingly applied across diverse domains, including gaming, robotics, and traffic systems. Some examples include AlphaStar for playing StarCraft \citep{AlphaStar}, multi-robot systems such as in warehouses \citep{krnjaic2022scalable,IJRR25}, military drones \citep{10713174}, and multi-robot pathfinding \citep{damani2021primal} as well as traffic signal control \citep{van2016coordinated,wei2021recent}, autonomous driving \citep{zhang2024multi}, and power systems \citep{MAPDN}. Many more examples exist and I expect many more will exist in the near future. 

\section{Future topics}
There are many different ways of performing cooperative MARL and current work has only scratched the surface. Studying the differences and similarities between DTE and CTDE variants of algorithms seems like a promising direction for understanding current methods and developing improved approaches for both cases. Determining what centralized information to use and how to best use it is another key question. 
Lastly, to the best of my knowledge, there are no globally optimal model-free MARL methods for Dec-POMDPs. This is surprising since there are optimal \emph{planning} methods where the model is assumed known \citep{Book16} and many globally optimal model-free methods for single-agent RL \citep{SuttonBarto18}. Developing such methods (even for the tabular case) would be interesting and may lead to new, better methods that could be approximated in the deep case.

\section*{Acknowledgements}

I thank Andrea Baisero, Shuo Liu, Roi Yehoshua, and the other members of my Lab for Learning and Planning in Robotics (LLRP) as well as my 7180 MARL seminar class (such as Asteria Kaeberlein) for reading my (very) rough drafts and providing comments that helped improve the document as well as Frans Oliehoek for the many discussions about these topics and the resulting text. I also thank NSF CAREER Award 2044993 for funding this effort. 

\bibliographystyle{abbrvnat}
\bibliography{theBib}

\end{document}